\documentclass[11pt]{article}
\usepackage{booktabs}
\usepackage{enumitem}
\usepackage[section]{placeins}
\usepackage{mathrsfs}
\usepackage{verbatim}
\usepackage{amsmath}
\usepackage{amsfonts}
\usepackage{amssymb}
\usepackage{latexsym}
\usepackage{graphicx}
\usepackage{hyperref}
\usepackage{xcolor}
\usepackage{comment}
\usepackage{subcaption}
\usepackage{diagbox}
\newtheorem{theorem}{\bf Theorem}
\newtheorem{lemma}[theorem]{\bf Lemma}
\newtheorem{proposition}[theorem]{\bf Proposition}
\newtheorem{corollary}[theorem]{\bf Corollary}

\newtheorem{example}[theorem]{\bf Example}
\newtheorem{defn}[theorem]{\bf Definition}
\newtheorem{definition}[theorem]{\bf Definition}
\newenvironment{proof}{\noindent{\em Proof:}}{\quad \hfill\(\Box\)\vspace{2ex}}

\newenvironment{remark}{\noindent{\bf Remark}}{\vspace{2ex}}

\def \naturals {{\mathbb N}}
\def \reals {{\mathbb R}}

\def \supp {\,{\rm supp}\,}

\def \qquad {\quad\quad}

\makeatletter
\newcommand{\rmnum}[1]{\romannumeral #1}
\newcommand{\Rmnum}[1]{\expandafter\@slowromancap\romannumeral #1@}
\g@addto@macro\appendix{%
  % ...

}
\makeatother

\graphicspath{{./}{../RKBS with the lp norm/}}

\usepackage{tcolorbox}
\tcbuselibrary{breakable}

\newif\ifrevnotes
\revnotestrue 

\NewDocumentEnvironment{revnote}{m +b}{%
  \ifrevnotes
    \begin{nolinenumbers}
    \begin{tcolorbox}[
      breakable,
      title={Revision note: #1},
      colback=yellow!5,
      colframe=orange!70!black,
      fonttitle=\bfseries,
      fontupper=\small
    ]
      #2
    \end{tcolorbox}
    \end{nolinenumbers}%
  \fi
}{}

\def \bN {\Bbb N}

\def \bR {\Bbb R}

\def \bx {{\bf x}}

\def \by {{\bf y}}

\def \and {\, \mbox{\rm and}\, }

\def \supp {\,{\rm supp}\,}
\def \spa {\,{\rm span}\,}

\begin{document}
\title{\bf Reproducing Kernel (Quasi-)Banach Spaces with the \(\ell^p\) Coefficient Norms}
\author
{
Tianfang Wang,\thanks{School of Mathematics, Sun Yat-sen University, Zhuhai, P.
R.
China.
E-mail address: {\it wangtf23@mail2.sysu.edu.cn}.
}\quad
Guohui Song\thanks{Department of Mathematics and Statistics, Old Dominion University, Norfolk,
VA 23529, USA.
E-mail address: {\it gsong@odu.edu}.
Supported in part by the National Science Foundation under grant DMS-2318781.
},
\quad
and \quad
    Haizhang Zhang\thanks{School of Mathematics, Sun Yat-sen University, Zhuhai, P.
    R.
    China.
    E-mail address: \textit{zhhaizh2@sysu.edu.cn}.
    Supported in part by the National Natural Science Foundation of China (Grant Nos.
    12671642 and 12371103) and the Guangdong Basic and Applied Basic Research Foundation (Grant No. 2024A1515011194).
    Corresponding author.
}
}
\date{}
\maketitle
\begin{abstract}
    This paper advances the theory of reproducing kernel Banach spaces (RKBSs) by investigating the general \(\ell^p\) regularization framework, with a particular focus on the non-convex regime \(0<p<1\).
    While our preceding work established the foundation for RKBSs with the \(\ell^1\) norm, the pursuit of stronger sparsity---closer to the \(\ell^0\) pseudo-norm---naturally motivates this exploration into \(\ell^p\) quasi-norms.
    For \(0<p<1\), we construct reproducing kernel quasi-Banach spaces (RKQBSs) with \(\ell^p\) quasi-norms and establish existence and sparsity results for interpolation and regularized learning under appropriate assumptions.
    We characterize the linear and relaxed representer theorems in terms of a generalized Lebesgue constant and derive bounds under kernel decay, sample spacing, and stability assumptions.
    For \(p>1\), we construct RKBSs using a fixed countable dictionary, establish minimizer existence, and identify the optimal relaxed representer factor through an off-sample coefficient operator.
    These results reveal fundamental differences between the regimes \(0<p<1\), \(p=1\), and \(p>1\).
    Numerical experiments compare the accuracy and sparsity of solutions for these three cases.
\end{abstract}

\section{Introduction}
Kernel methods constitute a vital component of machine learning theory, with wide applications in function approximation, regression analysis, classification, numerical solutions of partial differential equations, and scientific computing, \cite{Berlinet2004,Scholkopf2002,Shawe-Taylor2004,Steinwart2008,Vapnik1998}.
In recent years, with the rapid development of large-scale data analysis and high-dimensional learning tasks, reducing model complexity and improving representation efficiency while maintaining approximation capability has become a major research direction in kernel learning theory.
Among others, learning models that yield sparse representations are highly appealing as they not only exhibit lower computational complexity but also enhance interpretability and generalization ability.
Consequently, sparse models have attracted significant attention from fields such as machine learning, compressed sensing, and signal processing, \cite{Candes2006,Chen2001,Donoho2006,Tibshirani1996,Wang2026,Xu2012}.

Classical kernel learning theory is fundamentally built upon the framework of reproducing kernel Hilbert spaces (RKHSs) \cite{Aronszajn1950,Cucker2002,Cucker2007}.
Owing to the inner product structure of RKHSs, continuous pointwise evaluation functionals can be uniquely represented via the Riesz representation theorem, leading to elegant reproducing properties.
Building upon this foundation, Kimeldorf and Wahba \cite{Kimeldorf1971} introduced the renowned Representer Theorem, which was later generalized to a broader context \cite{Argyriou2009,Huang2021,Scholkopf2001,Wang2021}, thereby establishing the mathematical underpinnings of contemporary kernel learning theory.

Nonetheless, RKHSs employ the Hilbert norm as the regularization term, which essentially corresponds to quadratic regularization and typically leads to dense representations of the optimal solutions.
With the advances in compressed sensing, sparse representation, and high-dimensional statistics, researchers have paid increasing attention to learning models that can directly generate sparse solutions.
Compared with quadratic regularization, regularization based on the \(\ell^p\) norm (\(0 < p \le 1\)) induces sparsity much more effectively, where \(p=1\) corresponds to the classical convex sparse regularization widely studied in the literature \cite{Mallat1993,Song2013,Wang2024,Xu2023}.
Indeed, when \(0 < p < 1\), although the \(\ell^p\) space is no longer a Banach space, it can induce even stronger sparsity than \(\ell^1\), which is of great importance in compressed sensing, non-convex optimization, variable selection, and signal recovery, \cite{Blumensath2008,Chartrand2007,Chartrand2008,Foucart2009,Xu2012}.
Moreover, for the particular case when \(p=\frac12\) or \(p=\frac23\), the proximity operator for solving the \(\ell^p\)-regularized learning problems has an explicit form \cite{Cao2013}.
On the other hand, when \(p > 1\), the \(\ell^p\) space reverts to a Banach space, whose geometric structure differs fundamentally from that of the \(0 < p < 1\) regime, necessitating the redevelopment of its representer and approximation theories.
Therefore, studying reproducing kernel spaces associated with different parameters \(p\) under a unified framework is of importance to enriching kernel learning theory.

In recent years, the theory of reproducing kernel Banach spaces (RKBSs) has witnessed rapid development.
Existing works have constructed different types of RKBSs based on semi-inner products, bi-dual mappings, and bilinear forms, successfully extending representer theorems, interpolation theories, and regularized learning frameworks from RKHSs \cite{Fasshauer2015,Lin2022,Wang2024,Xu2019,Zhang2009}.
Furthermore, remarkable progress has recently been made in utilizing the RKBS framework to describe shallow and deep neural networks, introduce the hypothesis space, establish duality theory, analyze generalization bounds via Rademacher complexity, and extend these models to vector-valued settings for operators and hypernetworks \cite{Bartolucci2023,deLaHiguera2026,Dummer2025,Heeringa2025,Shilton2025,Spek2025,Wang2026}.
However, the current RKBS theory heavily relies on the geometric properties of Banach spaces, such as local convexity, the Hahn-Banach theorem, and uniform smoothness, which prevents its direct application to function spaces associated with \(0 < p < 1\) \cite{Megginson1998}.
To establish a learning theory in this regime, it is first necessary to understand the fundamental analytical properties of the \(\ell^p\) space (\(0 < p < 1\)), including characterization of continuous linear functionals, basic inequalities of the \(p\)-norm, and structural properties of the associated function spaces.

Building upon the aforementioned foundations, the primary purpose of the paper is to construct reproducing kernel (quasi-) Banach spaces with \(\ell^p\) coefficient (quasi-) norms and characterize their linear and relaxed representer theorems.
Specifically,  given a function \(K(x,y)\) (not necessarily positive-definite or symmetric), we seek a space \(B_p\) with the following properties:
\begin{enumerate}
    \item Point evaluations are continuous linear functionals on \(B_p\).
    \item The space \(B_p\) possesses the \(\ell^p\) norm (quasi-norm for \(0<p<1\)), that is,
          \[
              \|f\|_{B_p}=\|\mathbf{c}\|_{p}=\biggl( \sum_{m=1}^\infty |c_m|^p\biggr)^{1/p}
          \mbox{ for }
              f(x)=\sum_{m=1}^\infty c_mK(x_m,x).
          \]
    \item Under appropriate conditions, a \textbf{linear representer theorem} holds
          \[
              \min_{f=\sum_{j=1}^n c_jK(x_j, \cdot)}\sum_{j=1}^n L(f(x_j),y_j)+\lambda \|\mathbf{c}\|_{p}=\min_{f\in B_p}\sum_{j=1}^n L(f(x_j),y_j)+\lambda \|f\|_{B_p},
          \]
          where \(L\) stands for a loss function.
\end{enumerate}
The case when \(p=1\) was established in \cite{Song2013}.
What deters us from investigating the case when \(0<p<1\) is that \(\ell^p\) has a quasi-norm and that the dual space of \(L^p[0,1]\) when \(0<p<1\) is \(\{0\}\) \cite{Conway1990}, which means that there are no meaningful continuous linear functionals on \(L^p[0,1]\).
We now realize that the discrete space \(\ell^p\) is fundamentally different from the continuous space \(L^p[0,1]\).
Although \(\ell^p\) when \(0<p<1\) has a quasi-norm which does not satisfy the triangle inequality, it does satisfy a quasi-triangle inequality, which is sufficient for approximation in the corresponding RKBSs.
Also, the dual space of \(\ell^p\) when \(0<p<1\) is always \(\ell^\infty\), which results in bounded point evaluations when \(K(\cdot,x)\) is bounded for each \(x\). For \(0<p<1\), we are able to construct reproducing kernel quasi-Banach spaces (RKQBSs) with \(\ell^p\) quasi-norms to accommodate sparse learning models with continuous point evaluations.

This paper first investigates the minimum norm interpolation problem in the \(0 < p < 1\) regime.
Under the continuity and compactness or decay assumptions stated below, we prove existence of minimal norm interpolants for feasible targets and regularized minimizers with freely chosen centers. Nonsingularity of the sample matrix guarantees feasibility for every target.
Minimal norm interpolants use at most \(n\) centers in the real-valued setting; a regularized minimizer with the same bound exists.
These centers need not be sampling points, so their locations must also be determined, which complicates computation. This motivates the linear representer theorem, which guarantees the existence of a minimizer with the sampling points as centers. Compared with existing \(\ell^1\) regularization theories, this work extends representer theorems to non-locally convex function spaces, providing a unified functional analytic framework for non-convex kernel learning models. We emphasize that the linear representer theorem in the current study seeks minimal solutions expressed as a linear combination of the kernel functions at the sampling points, making the corresponding coefficient regularization learning algorithms computationally tractable, whereas existing studies \cite{Wang2021,Wang2024,Xu2023,Zhang2009} on representer theorems in Banach spaces express the minimizer in the dual space of RKBSs, which may cause computational difficulties.

The linear representer theorem is closely related to the generalized Lebesgue constant of the associated kernel.
Our main theoretical results distinguish \(0<p<1\) and \(p>1\), with the classical case \(p=1\) providing a comparison.
For \(p=1\), the linear representer theorem is characterized by the condition \(\Lambda_1(\mathbf x)=1\), where \(\Lambda_1\) is the Lebesgue constant associated with interpolation \cite{Song2013}. Kernels satisfying this condition include the Brownian bridge.
For \(0<p<1\), the necessary and sufficient condition is \(\Lambda_p(\mathbf x)=1\), where \(\Lambda_p\) is the generalized Lebesgue constant defined in \eqref{eq:generalized-lebesgue-constant}. This condition fails for two-point configurations of several classical kernels.
By allowing a relaxed representer theorem with a factor \(\beta_n>1\), we show that for \(0<p<1\), the relaxed theorem holds precisely when \(\Lambda_p(\mathbf x)\le\beta_n\). Decay assumptions, sample spacing, and uniform stability of the finite sample matrices give bounds on this factor as the sample size grows.
For \(p>1\), we show in Section~\ref{sec:p-greater-one} that the representer theorem fails in general. We shall turn to establish a sufficient condition for a relaxed representer theorem by requiring the \(\ell^p\)-Lebesgue constant to be bounded by \(\beta_n - 1\).

The remainder of this paper is organized as follows.
Section 2 introduces RKQBSs with \(\ell^p\) quasi-norms for \(0<p<1\), establishes their basic properties, and gives sufficient conditions for their construction.
Section 3 establishes existence and sparsity results for interpolation and regularized learning and characterizes the linear representer theorem.
Section 4 establishes the relaxed representer theorem in the \(0 < p < 1\) regime and investigates the corresponding Lebesgue constant estimates.
Section~\ref{sec:p-greater-one} develops the fixed-dictionary RKBS construction, spectral conditions, linear representer obstruction, and sharp relaxed estimates for \(p>1\).
Section 6 presents numerical experiments using three typical translation-invariant kernels: the Gaussian kernel, the Laplacian kernel, and the inverse multiquadric (IMQ) kernel.
The experiments are performed under various target functions, noise levels, regularization parameters, and values of \(p\).
The experimental results indicate that when \(0 < p < 1\), our proposed model achieves sparser representations while maintaining high approximation accuracy.

\section{RKQBSs with \(\ell^p\) Norms for \(0<p<1\)}

This section constructs RKQBSs with \(\ell^p\) quasi-norms for
\(0<p<1\). We begin with the sequence-space properties needed for
kernel synthesis and define the function space through a global
coefficient representation. We then verify the kernel assumptions
for a class of translation-invariant kernels, establish completeness
and bounded point evaluations, and construct a continuous bilinear
pairing that gives the reproducing identity. Throughout the paper, we shall work mainly on the field $\mathbb{R}$ of real numbers.

\subsection{\(\ell^p\) Spaces}

\begin{definition}
    For \(0<p<+\infty\), the sequence space \(\ell^p\) is defined as the set of all real-valued sequences \(\mathbf{c} = \{c_n\}_{n \in \naturals}\) such that \[ \sum_{n=1}^\infty |c_n|^p < \infty.
    \]
    It is equipped with the norm for $p\ge1$ or quasi-norm for $0<p<1$, denoted by \(\|\cdot\|_{p}\) and defined by
    \begin{equation}
        \|\mathbf{c}\|_{p} := \left( \sum_{n=1}^\infty |c_n|^p \right)^{1/p}.
    \end{equation}
    The spaces \(\ell^\infty\) and \(c_0\) consist of bounded real-valued sequences, and real sequences that converge to zero at infinity, respectively.
    The norm on \(\ell^\infty\) and \(c_0\) is \[ \|\mathbf{c}\|_\infty=\sup_{n\in\bN}|c_n|.
    \]
\end{definition}

Note that \(c_0\) is a subspace of \(\ell^\infty\).
We present below properties for \(0<p<1\) that will be needed in later sections.

When \(0 < p < 1\), the space \(\ell^p\) differs significantly from standard Banach spaces.
\begin{lemma}\label{lemma:norm_comparison}\cite{Kalton2003}
    Let \(0<p<1\).
    The following statements hold:
    \begin{enumerate}
        \item \(\|\cdot\|_p^p\) satisfies the triangle inequality
              \begin{equation}\label{triangleofmetric}
                  \|\mathbf{x}+\mathbf{y}\|_{p}^{p} \le \|\mathbf{x}\|_{p}^{p} + \|\mathbf{y}\|_{p}^{p},\ \ \mathbf{x},\mathbf{y}\in \ell^p.
              \end{equation}
        \item \(\ell^p\) is a complete metric space with the metric
              \[
                  d(\mathbf{x},\mathbf{y})=\|\mathbf{x}-\mathbf{y}\|_p^p,\ \ \mathbf{x},\mathbf{y}\in \ell^p.
              \]
        \item The quasi-norm \(\|\cdot\|_p\) satisfies a quasi-triangle inequality
              \[
                  \|\mathbf{x}+\mathbf{y}\|_{p} \le 2^{\frac1p-1}(\|\mathbf{x}\|_{p} + \|\mathbf{y}\|_{p}),\ \ \mathbf{x},\mathbf{y}\in \ell^p.
              \]
        \item Every continuous linear functional \(L\) on
              \(\ell^p\) has a unique representation
              \[
                  L(\mathbf a)=\sum_{n=1}^{\infty}a_nc_n,
                  \qquad \mathbf a\in\ell^p,
              \]
              with \(\mathbf c\in\ell^\infty\), and
              \[
                  \|L\|:=\sup_{\|\mathbf a\|_p\le1}|L(\mathbf a)|
                  =\|\mathbf c\|_\infty.
              \]
              Thus \((\ell^p)^*\) is isometrically identified with
              \(\ell^\infty\) through this bilinear pairing.
    \end{enumerate}

\end{lemma}

The following lemma extends finite subadditivity of $\|\cdot\|_p^p$ to infinite subadditivity.

\begin{lemma}\label{lemma:Infinite Subadditivity of Quasi-Norm}
    Let \(0<p<1\) and let
    \(\mathbf x^{(j)}=\{x_i^{(j)}\}_{i\ge1}\in\ell^p\) for \(j\ge0\).
    Suppose that, for every \(i\ge1\), the scalar series
    \[
        x_i:=\sum_{j=0}^{\infty}x_i^{(j)}
    \]
    converges and that \(\mathbf x=\{x_i\}_{i\ge1}\in\ell^p\). Then
    \[
        \|\mathbf x\|_p^p\le\sum_{j=0}^{\infty}
        \|\mathbf x^{(j)}\|_p^p.
    \]
\end{lemma}
\begin{proof}
    Finite subadditivity gives
    \begin{equation}\label{eq:finite_sub_lp}
        \biggl\|\sum_{j=0}^{N}\mathbf x^{(j)}\biggr\|_p^p
        \le\sum_{j=0}^{N}\|\mathbf x^{(j)}\|_p^p.
    \end{equation}
    By coordinatewise convergence and Fatou's lemma,
    \[
        \|\mathbf x\|_p^p
        \le\liminf_{N\to\infty}
        \left\|\sum_{j=0}^{N}\mathbf x^{(j)}\right\|_p^p
        \le\sum_{j=0}^{\infty}\|\mathbf x^{(j)}\|_p^p.
    \]
\end{proof}

The next lemma indicates that for $0<p<1$, $\ell^p\subseteq \ell^1$.
\begin{lemma}\label{lemma:lpinl1}
For $0<p<1$, \(\ell^p\subseteq\ell^1\) and it holds
\[
    \|\mathbf a\|_1\le\|\mathbf a\|_p,\qquad \mathbf a\in\ell^p.
\]
\end{lemma}
\begin{proof}
    As $0<p<1$, we have the finite subadditivity
\[
\biggl(\sum_{n=1}^N|a_n|\biggr)^p\le\sum_{n=1}^N|a_n|^p.
\]
Letting \(N\to\infty\) and then raising both sides to the power $\frac{1}{p}$ proves the claim.
\end{proof}

\subsection{Kernel-based Function Spaces with \(\ell^p\) Quasi-Norms}

Let \(X\) be a nonempty set and let \(K:X\times X\to\reals\).
We construct a function space from linear combinations of
\(K(t,\cdot)\), with coefficients in \(\ell^p\).
The function \(K\) is not required to be symmetric or positive semi-definite.

\begin{definition}\label{def: lpX}
    For \(0<p<1\), denote by \(\ell^p(X)\) the space of \(p\)-integrable functions on \(X\) with respect to the counting measure on \(X\). Namely,
    \[
        \ell^p(X):=\left\{a:X\to\reals:
        \sum_{t\in X}|a(t)|^p<+\infty\right\},
    \]
    equipped with the quasi-norm
    \[
        \|a\|_{\ell^p(X)}:=
        \left(\sum_{t\in X}|a(t)|^p\right)^{1/p}.
    \]
    \end{definition}
    
    Note that if \(a\in\ell^p(X)\) then for each
    \(m\ge1\), the set \(\{t\in X:|a(t)|\ge1/m\}\) is finite. Therefore every 
   \(a\in\ell^p(X)\) has at most countable support. The sums indexed by \(X\) in the above definition are taken over at most countable nonzero coefficients of $a\in\ell^p(X)$.

We next construct function spaces with the $\ell^p$ quasi-norm from $K$.

    \begin{definition}\label{def: representation space}
    Assume that the kernel \(K:X\times X\to\reals\) satisfies:
    \begin{description}
        \item[(K1)] For every \(x\in X\),
              \(M_x:=\sup_{t\in X}|K(t,x)|<+\infty\).
        \item[(K2)] Synthesis is injective on the entire coefficient space:
              if \(a\in\ell^p(X)\) and
              \(\sum_{t\in X}a(t)K(t,x)=0\) for every \(x\in X\), then \(a=0\).
    \end{description}
    Define
    $$
        (Ta)(x):=\sum_{t\in X}a(t)K(t,x),\ \ a\in\ell^p(X).
        $$
    Construct 
    \begin{equation}\label{eq:B_p}
        B_p:=T(\ell^p(X))=\left\{Ta=\sum_{t\in X}a(t)K(t,\cdot):\ a\in\ell^p(X)\right\}
    \end{equation}
    and equip $B_p$ with the quasi-norm $\|Ta\|_{B_p}:=\|a\|_{\ell^p(X)}$.
\end{definition}

For \(0<p<1\), by Lemma \ref{lemma:lpinl1},
\[
    \sum_{t\in X}|a(t)K(t,x)|
    \le M_x\|a\|_{\ell^1(X)}
    \le M_x\|a\|_{\ell^p(X)}.
\]
Thus the synthesis series is absolutely convergent and independent of enumeration.
Condition (K2) gives a unique coefficient function for every element of
\(B_p\), so its quasi-norm is well-defined.

A sufficient condition for (K2) is provided by \(c_0\)-universality.
Suppose \(X\) is a locally compact Hausdorff space and \(K\) is a bounded
continuous real-valued positive-definite kernel with
\(K(\cdot,x)\in C_0(X)\) for every \(x\in X\). Recall that $C_0(X)$ is the space of continuous functions $f$ on $X$ such that for every $\varepsilon>0$, $\{t\in X:|f(t)|\ge \varepsilon\}$ is compact. If the RKHS of $K$ is dense in \(C_0(X)\), then the embedding
of finite signed Borel measures induced by \(K\) is injective
\cite[Proposition 2]{Sriperumbudur2011}.
For \(a\in\ell^p(X)\subseteq\ell^1(X)\), the coefficient measure
\(\mu_a=\sum_{t\in X}a(t)\delta_t\) is finite. Applying injectivity shows that \(Ta=0\) implies
\(\mu_a=0\), and hence \(a=0\). Thus a \(c_0\)-universal kernel
satisfies (K2).

We next verify the global conditions for \(0<p<1\) for a class of translation-invariant kernels.
Recall that \((\ell^p)^*=\ell^\infty\) for \(0<p<1\).
Throughout the paper, \(\bR^d\) is equipped with the standard Euclidean metric \[ d(\mathbf{x},\mathbf{y})=\|\mathbf{x}-\mathbf{y}\|_2=\biggl(\sum_{k=1}^d |x_k-y_k|^2\biggr)^{1/2},\ \ \mathbf{x},\mathbf{y}\in\bR^d.
\]
The following theorem verifies (K1)--(K2) using the Fourier transform
of a finite signed Borel measure.

\begin{theorem}\label{K1K2pro}
    Let \(X=\bR^d\).
    For a given function \(\varphi\in L^1(\bR^d)\) , define \(K\) by
    \begin{equation}
        \label{eq:fourier_repr}
        K(\mathbf{x}, \mathbf{y}) = \int_{\reals^d} e^{-i(\mathbf{x}-\mathbf{y})\cdot\xi} \varphi(\xi) d\xi, \quad \mathbf{x}, \mathbf{y} \in \reals^d.
    \end{equation}
    For \(0<p<1\), (K1) holds globally, and \(K(\cdot,\bx)\in C_0(\bR^d)\) for every \(x\). If \(\varphi\) is nonzero almost everywhere, (K2) holds on \(\ell^p(\bR^d)\).
\end{theorem}

\begin{proof}
    Define the Fourier transform of \(\varphi\) by
    \[
        \widehat\varphi(\mathbf u):=\int_{\bR^d}e^{-i \mathbf u\cdot\xi}
        \varphi(\xi)\,d\xi,\qquad \mathbf u\in\bR^d.
    \]
    By the Riemann--Lebesgue lemma,
    \(\widehat\varphi\in C_0(\bR^d)\). Since
    \(K(\mathbf t,\mathbf x)=\widehat\varphi(\mathbf t-\mathbf x)\), the function
    \(K(\cdot,\mathbf x)\) belongs to \(C_0(\bR^d)\) for every \(\mathbf x\in\bR^d\), and
    \[
        \sup_{\mathbf t\in\bR^d}|K(\mathbf t,\mathbf x)|
        \le\|\varphi\|_{L^1(\bR^d)}.
    \]
    Thus (K1) holds.

    Suppose now that \(\varphi\) is nonzero almost everywhere.
    Fix \(0<p<1\), and let \(a\in\ell^p(\bR^d)\) satisfy
    \[
        (Ta)(x)=\sum_{\mathbf t\in\bR^d}a(\mathbf t)K(\mathbf t,\mathbf x)=0,
        \qquad \mathbf x\in\bR^d.
    \]
    As \(a\) has at most countable support and
    \(\|a\|_{\ell^1(\bR^d)}\le\|a\|_{\ell^p(\bR^d)}\), the series
    \[
        \mu:=\sum_{\mathbf t\in\bR^d}a(\mathbf t)\delta_{\mathbf t}
    \]
    converges in total variation and defines a finite signed Borel measure. Its Fourier transform is the bounded continuous function
    \[
        \widehat\mu(\xi):=\int_{\bR^d}e^{-i \mathbf t\cdot\xi}\,d\mu(\mathbf t)
        =\sum_{\mathbf t\in\bR^d}a(\mathbf t)e^{-i \mathbf t\cdot\xi}.
    \]
    Moreover,
    \[
        \int_{\bR^d}\sum_{\mathbf t\in\bR^d}
        |a(\mathbf t)\varphi(\xi)|\,d\xi
        =\|a\|_{\ell^1(\bR^d)}\|\varphi\|_{L^1(\bR^d)}<\infty.
    \]
    Fubini's theorem and \eqref{eq:fourier_repr} therefore give
    \[
        0=(Ta)(\mathbf x)=\int_{\bR^d}e^{i \mathbf x\cdot\xi}
        \varphi(\xi)\widehat\mu(\xi)\,d\xi,
        \qquad \mathbf x\in\bR^d.
    \]
    The product \(\varphi\widehat\mu\) belongs to \(L^1(\bR^d)\).
    Uniqueness of the Fourier transform on \(L^1(\bR^d)\) implies
    \(\varphi\widehat\mu=0\) almost everywhere. Since
    \(\varphi\ne0\) almost everywhere, \(\widehat\mu=0\) almost
    everywhere. The continuity of \(\widehat\mu\) then gives \(\widehat\mu=0\) on all of
    \(\bR^d\). Uniqueness of the Fourier transform for finite signed Borel measures yields \(\mu=0\). Finally,
    \(a(\mathbf t)=\mu(\{\mathbf t\})=0\) for every \(\mathbf t\in\bR^d\), proving (K2).
\end{proof}

We now give three translation-invariant kernels to which Theorem \ref{K1K2pro} applies.

\begin{example}[Gaussian Kernel]
    Let \(X = \reals^d\) and let \(K\) be the Gaussian kernel defined by
    \begin{equation}\label{eq:Gaussian kernel}
        K(\mathbf{x}, \mathbf{y}) = \exp \left( - \frac{\|\mathbf{x}-\mathbf{y}\|_2^2}{2\sigma^2} \right), \quad \sigma > 0.
    \end{equation}
    This is a translation-invariant kernel of the form \eqref{eq:fourier_repr} with the spectral density \[ \varphi(\xi) = \left(\frac{\sigma^2}{2\pi}\right)^{d/2} \exp\left(-\frac{\sigma^2 \|\xi\|_2^2}{2}\right).
    \]
\end{example}

\begin{example}[Laplacian Kernel]
    Let \(X = \reals^d\) and let \(K\) be the Laplacian kernel defined by
    \begin{equation}
        \label{eq:Laplacian Kernel}
        K(\mathbf{x}, \mathbf{y}) = \exp \left( -\gamma \|\mathbf{x}-\mathbf{y}\|_2 \right), \quad \gamma > 0.
    \end{equation}
    This is a translation-invariant kernel associated with the spectral density \[ \varphi(\xi) = \frac{\gamma \Gamma(\frac{d+1}{2})}{\pi^{\frac{d+1}{2}} (\gamma^2 + \|\xi\|_2^2)^{\frac{d+1}{2}}}, \] where \(\Gamma\) is the Gamma function.
\end{example}

\begin{example}[Inverse Multiquadric Kernel]
    Let \(X = \reals^d\) and let \(K\) be the inverse multiquadric kernel defined by
    \begin{equation}
        \label{eq:Inverse Multiquadric Kernel}
        K(\mathbf{x}, \mathbf{y}) = \left( c^2 + \|\mathbf{x}-\mathbf{y}\|_2^2 \right)^{-\beta}, \quad c > 0, \quad \beta > d/2.
    \end{equation}
    This kernel can be represented as \(K(\mathbf{x}, \mathbf{y}) = \Phi(\mathbf{x} - \mathbf{y})\), where \(\Phi(\mathbf{x}) = \left( c^2 + \|\mathbf{x}\|_2^2 \right)^{-\beta}\).
    Since \(\beta > d/2\), we have \(\Phi \in L^1(\reals^d)\), and by the Fourier inversion theorem, its associated spectral density \(\varphi\) is given by \[ \varphi(\xi) = \frac{1}{(2\pi)^{\frac{d}{2}}}\frac{2^{1-\beta}}{\Gamma(\beta)} \left( \frac{\|\xi\|_2}{c} \right)^{\beta - \frac{d}{2}} \mathcal{K}_{\frac{d}{2}-\beta}(c\|\xi\|_2),\ \ \xi\ne 0, \] where \(\mathcal{K}_{\nu}\) is the modified Bessel function of the second kind of order \(\nu\).
    The spectral density \(\varphi\) belongs to \(L^1(\bR^d)\) and is positive almost everywhere \cite{Hamm2017}.
\end{example}

\subsection{Reproducing Kernel Quasi-Banach Space Structure of \(B_p\)}
\label{subsec:RKBS}

We establish completeness and bounded point evaluations for the global
construction. We use the term reproducing kernel Quasi-Banach space
for the following adaptation of the Banach space framework in
\cite{Lin2022}.

\begin{definition}[Reproducing Kernel Quasi-Banach Space]\label{def:RKBS}
    Let \(X\) be a nonempty set. A quasi-Banach space \(B\) of
    real-valued functions on \(X\) is called a reproducing kernel
    Quasi-Banach space if, for every \(x\in X\), the point
    evaluation functional \(\delta_x:B\to\reals\), defined by
    \(\delta_x(f)=f(x)\), is continuous. Equivalently, for every
    \(x\in X\), there exists a constant \(C_x>0\) such that
    \begin{equation}\label{eq:continuous-eval}
        |f(x)|\le C_x\|f\|_B,\qquad f\in B.
    \end{equation}
\end{definition}

\begin{proposition}\label{prop:global-completeness}
    Under (K1)--(K2), \(B_p\) is isometrically isomorphic to
    \(\ell^p(X)\). In particular, it is a quasi-Banach space. Moreover,
    \[
        |f(x)|\le M_x\|f\|_{B_p},\qquad x\in X.
    \]
\end{proposition}
\begin{proof}
The evaluation bound follows from
    \[
        |(Ta)(x)|\le \sum_{t\in X}|a(t)||K(t,x)|\le M_x\|a\|_{\ell^1(X)}\le M_x\|a\|_{\ell^p(X)}.
    \]
    For completeness, \(\ell^p(X)\) is complete even if \(X\) is
    uncountable: a Cauchy sequence is supported in the countable union
    of its supports and hence lies in one complete sequence space.
    Conditions (K1)--(K2) make \(T\) a linear bijection from
    \(\ell^p(X)\) onto \(B_p\), and the quasi-norm definition makes it an
    isometry. Thus a Cauchy sequence \(Ta_m\) converges to \(Ta\),
    where \(a_m\to a\) in \(\ell^p(X)\).
    Completeness is with respect to the metric
    \(d(f,g)=\|f-g\|_{B_p}^p\).
\end{proof}

\begin{definition}[Reproducing kernel]\label{def:reproducing-kernel}
    Let \(\Omega_1\) and \(\Omega_2\) be nonempty sets, and let
    \(B_1\) be a reproducing kernel quasi-Banach space on \(\Omega_1\).
    A function \(K:\Omega_2\times\Omega_1\to\reals\) is called a
    \emph{reproducing kernel} for \(B_1\) if there exist a Banach space
    \(B_2\) of real-valued functions on \(\Omega_2\) and a continuous bilinear form \(\langle\cdot,\cdot\rangle_{B_1\times B_2}\) such that
    \(K(\cdot,x)\in B_2\) for every \(x\in\Omega_1\) and
    \begin{equation}\label{eq:reproducing-property}
        f(x)=\langle f,K(\cdot,x)\rangle_{B_1\times B_2},
        \qquad f\in B_1,\quad x\in\Omega_1.
    \end{equation}
\end{definition}

To construct the reproducing pairing, let
\[
    B_2:=\ell^\infty(X)
    =\left\{g:X\to\reals:
    \|g\|_\infty:=\sup_{t\in X}|g(t)|<\infty\right\},
\]
a Banach space with the supremum norm. Define the bilinear form on
\(B_p\times B_2\) by
\[
    \langle Ta,g\rangle_{B_p\times B_2}
    :=\sum_{t\in X}a(t)g(t),
    \qquad a\in\ell^p(X),\quad g\in B_2.
\]
The series converges absolutely, and (K2) makes the form well-defined.
Moreover,
\[
    |\langle Ta,g\rangle_{B_p\times B_2}|
    \le\|a\|_{\ell^1(X)}\|g\|_\infty
    \le\|Ta\|_{B_p}\|g\|_\infty,
\]
so the form is continuous. By (K1), \(K(\cdot,x)\in B_2\) for every
\(x\in X\), and the definition of synthesis gives
\[
    \langle f,K(\cdot,x)\rangle_{B_p\times B_2}=f(x),
    \qquad f\in B_p,\quad x\in X.
\]
Thus \(K\) is a reproducing kernel for \(B_p\) in the sense of
Definition~\ref{def:reproducing-kernel}.

If $X$ is a metric space and \(K(\cdot,x)\in C_0(X)\) for every \(x\in X\), the same construction
applies with \(B_2=C_0(X)\). Define \(\mu_a=\sum_{t\in X}a(t)\delta_t\). Then \(\mu_a\) is a finite signed Borel measure. The pairing then has the integral representation
\[
    \langle Ta,g\rangle_{B_p\times C_0(X)}
    =\int_Xg(t)\,d\mu_a(t),\qquad g\in C_0(X).
\]

\section{Linear Representer Theorems for \(0<p<1\)}

This section studies the existence and representation of minimizers
for interpolation and regularized learning in \(B_p\), with \(0<p<1\).
We establish the equivalence of the linear representer properties
for interpolation and regularized learning
and show by a counterexample why additional assumptions are needed
for general existence results. Under sufficient compactness and
continuity or decay conditions, we prove existence and sparsity of
minimizers. We then characterize the linear representer theorem by a
coefficient condition and illustrate its failure for several kernels.

Throughout this section, assume (K1)--(K2).
A regularized learning scheme in \(B_p\) is typically formulated as finding an optimal function \(f_0 \in B_p\) that solves the regularized minimization problem:
\begin{equation}
    \label{eq:reg_learning_Bp}
    \min_{f \in B_p} \left[ V(f(\bx)) + \mu \phi(\|f\|_{B_p}) \right].
\end{equation}
Here, \(\mathbf{x}:=(x_1,\ldots,x_n)\) is an ordered tuple of pairwise distinct sampling points in \(X\), \(f(\mathbf{x}) := (f(x_j) : j \in \mathbb{N}_n) \in \reals^n\), and \(\mathbb{N}_n:=\{1,2,\cdots, n\}\).
The function \(V : \reals^n \to \reals_{+}\) represents the loss function, \(\mu > 0\) is a positive regularization parameter, and \(\phi : \reals_{+} \to \reals_{+}\) is a nondecreasing regularization function, with \(\reals_{+} := [0, +\infty)\).
For a fixed choice of these data, write
\[
    J(f):=V(f(\mathbf x))+\mu\phi(\|f\|_{B_p}),\qquad f\in B_p.
\]

We use the following conditions on the loss and regularization functions \cite{Song2013}.

\begin{definition}\label{def:regularized learning scheme is acceptable}
    The regularized learning scheme \eqref{eq:reg_learning_Bp} is \emph{acceptable} if \(V\) and \(\phi\) are continuous, and
    \begin{equation}
        \lim_{t \to \infty} \phi(t) = +\infty.
    \end{equation}
\end{definition}

\begin{definition}\label{def:linear representer theorems}
    The space \(B_p\) is said to satisfy the {\bf linear representer theorem for regularized learning} if, for every \(n\ge1\) and every ordered tuple \(\mathbf x\) of pairwise distinct points in \(X\), every acceptable learning scheme (\ref{eq:reg_learning_Bp}) in \(B_p\) admits a minimizer in
    \begin{equation}\label{eq:span K}
        S^{\mathbf{x}} := \mathrm{span}\{K(x_j,\cdot) : j \in \mathbb{N}_n\},
    \end{equation}
    equivalently, there exists \(f_*\in S^{\mathbf x}\) such that
    \[
        J(f_*)=\inf_{f\in B_p}J(f).
    \]
\end{definition}

The linear representer theorem is central to kernel methods because it ensures that, for a broad class of regularized learning problems, at least one optimal solution lies in the finite-dimensional span of the kernel functions evaluated at the training points.
This reduces an apparently infinite-dimensional optimization problem in \(B_p\) to a finite-dimensional one in \(S^{\mathbf{x}}\) involving only the training data, making learning algorithms computationally tractable.
Moreover, the linear representer theorem provides a theoretical guarantee on the generalization ability of the minimizer from \[ \min_{f \in S^{\mathbf{x}}} \left[ V(f(\bx)) + \mu \phi(\|f\|_{B_p}) \right].
\]
A widely-used mathematical framework \cite{Cucker2002,Cucker2007} to analyze the generalization ability or learning rate of a regularized kernel method is to decompose the error into the sum of the sampling error, the hypothesis error, and the regularization error.
When there is a linear representer theorem, the hypothesis error will disappear automatically \cite{Song2011}.

It is well-known that regularized kernel methods are closely related to the minimal norm interpolation problems.
For any target vector \(\mathbf{y} = (y_j : j \in \mathbb{N}_n) \in \reals^n\), let \(I_{\mathbf{x}}(\mathbf{y})\) denote the subset of functions in \(B_p\) that interpolate the given data at the distinct sampling points \(\mathbf{x}:=(x_1,\ldots,x_n)\), defined by
\begin{equation}
    \label{eq:interp_subset}
    I_{\mathbf{x}}(\mathbf{y}) := \left\{ f \in B_p : f(\mathbf{x}) = \mathbf{y} \right\}.
\end{equation}
To ensure that \(I_{\mathbf{x}}(\mathbf{y}) \) is always nonempty, we impose another requirement on \(K\).
\begin{description}
    \item[(K3)] For every \(n\ge1\) and every ordered tuple \(\mathbf x=(x_1,\ldots,x_n)\) of pairwise distinct points in \(X\), the matrix
          \[
              \mathrm{K}[\bx] := [K(x_j, x_i)]_{1 \le i, j \le n}\]
          is nonsingular.
\end{description}
With this convention, if \(f=\sum_{j=1}^n c_jK(x_j,\cdot)\), then
\(f(\bx)=\mathrm K[\bx]\mathbf c\).
Condition (K3) holds for kernels of the form \eqref{eq:fourier_repr}
if \(\varphi\ge0\) almost everywhere and \(\{\xi:\varphi(\xi)>0\}\)
has positive Lebesgue measure. Indeed, for \(\mathbf c\in\reals^n\),
\[
    \mathbf c^{\mathsf{T}}\mathrm K[\bx]\mathbf c
    =\int_{\bR^d}\varphi(\xi)
    \left|\sum_{j=1}^n c_je^{-i x_j\cdot\xi}\right|^2\,d\xi.
\]
For \(\mathbf c\ne0\), the exponential polynomial in the integrand is not identically zero, since its frequencies are distinct. Its zero set has
Lebesgue measure zero, so the displayed integral is positive.
Thus \(\mathrm K[\bx]\) is strictly positive definite and hence nonsingular.

A minimal norm interpolation problem in \(B_p\) for given \(\mathbf{x}=(x_1,\ldots,x_n)\) and \(\by\in \bR^n\) is
\begin{equation}
    \label{eq:min_norm_interp}
    \min_{f \in I_{\mathbf{x}}(\mathbf{y})} \|f\|_{B_p}.
\end{equation}

\begin{definition}\label{def:linear representer theorems2}
    The space \(B_p\) is said to satisfy the {\bf linear representer theorem for minimal norm interpolation} if, for every \(n\ge1\), every ordered tuple \(\mathbf x\) of pairwise distinct points in \(X\), and every \(\by\in\reals^n\), (\ref{eq:min_norm_interp}) has a minimizer in \(S^{\bx}\).
\end{definition}

\begin{lemma}\label{equivalentRLMI}
    Let \(0<p<1\) and assume (K1)--(K3). Fix an ordered tuple
    \(\mathbf x=(x_1,\ldots,x_n)\) of pairwise distinct sampling points.
    Every acceptable regularized learning problem based on \(\mathbf x\)
    has a minimizer in \(S^{\mathbf x}\) if and only if, for every
    \(\mathbf y\in\reals^n\), the minimal norm interpolation problem
    has a minimizer in \(S^{\mathbf x}\).
    Consequently, the two linear representer properties are equivalent.
\end{lemma}
\begin{proof}
    Suppose first that every interpolation problem has a minimal norm
    solution in \(S^{\mathbf x}\). Given an acceptable objective
    \(J(f)=V(f(\bx))+\mu\phi(\|f\|_{B_p})\), replace any \(f\in B_p\)
    by such a solution \(g\) with target \(f(\bx)\). Then
    \(g(\bx)=f(\bx)\) and \(\|g\|_{B_p}\le\|f\|_{B_p}\), so
    \(J(g)\le J(f)\). Therefore the full-space infimum equals the
    infimum over \(S^{\mathbf x}\). On that span the coefficient
    objective is
    \[
        \mathbf c\longmapsto
        V(\mathrm K[\bx]\mathbf c)+\mu\phi(\|\mathbf c\|_p).
    \]
    It is continuous and coercive on \(\reals^n\), since \(V\ge0\)
    and \(\phi(r)\to+\infty\) as $r\to+\infty$. It attains its minimum, giving a
    full-space minimizer in \(S^{\mathbf x}\).

    Conversely, suppose that every acceptable learning problem based
    on \(\mathbf x\) has a minimizer in \(S^{\mathbf x}\).
    Fix \(\mathbf y\in\reals^n\), and let
    \(h=\sum_{j=1}^n[\mathrm K[\bx]^{-1}\mathbf y]_jK(x_j,\cdot)\).
    For each integer \(m\ge1\), choose a full-space minimizer
    \(f_m=\sum_{j=1}^n c_{m,j}K(x_j,\cdot)\) of the acceptable objective
    \[
        J_m(f):=m\|f(\bx)-\mathbf y\|_2^2+\|f\|_{B_p}^p.
    \]
    Comparison with \(h\) yields
    \[
        m\|\mathrm K[\bx]\mathbf c_m-\mathbf y\|_2^2
        +\|\mathbf c_m\|_p^p\le\|h\|_{B_p}^p.
    \]
    Thus \(\{\mathbf c_m\}\) is bounded. Passing to a subsequence gives
    \(\mathbf c_{m_k}\to\mathbf c\) with
    \(\mathrm K[\bx]\mathbf c=\mathbf y\).
    Set \(f=\sum_jc_jK(x_j,\cdot)\).
    For every \(g\in I_{\bx}(\mathbf y)\), minimality gives
    \(\|\mathbf c_{m_k}\|_p^p\le J_{m_k}(f_{m_k})\le J_{m_k}(g)=\|g\|_{B_p}^p\).
    Passing to the limit proves \(\|f\|_{B_p}\le\|g\|_{B_p}\).
    Hence \(f\) is a minimal norm interpolant in \(S^{\mathbf x}\).
\end{proof}

We next establish existence and sparsity for \(0<p<1\) before characterizing the linear representer theorem. Section~\ref{sec:relaxed-small-p} treats the relaxed representer theorem.

The following example shows that (K1)--(K3) alone do not guarantee
attainment of the minimal norm interpolation problem, even when the
kernel is bounded and the target is feasible.

\begin{example}[Nonattainment of the interpolation infimum]
    \label{ex:interpolation-nonattainment}
    Let \(X=\mathbb N\) with its discrete topology, fix \(0<p<1\),
    and define
    \[
        a_i:=2-\frac1i,\qquad
        K(i,j):=a_i a_j+\frac12\delta_{ij},\qquad i,j\in\mathbb N,
    \]
    where \(\delta_{ij}\) is the Kronecker delta.
    The kernel is bounded and hence satisfies (K1).
    To verify (K2), let \(c\in\ell^p\) satisfy \(Tc=0\), and put
    \(S:=\sum_{i\ge1}c_i a_i\). This series converges absolutely since
    \( c\in\ell^1\) and \(\{a_i\}\) is bounded. For every \(j\),
    \[
        0=(Tc)(j)=a_j S+\frac12c_j.
    \]
    Thus \(c_j=-2Sa_j\). Since \(a_j\to2\) and \(c_j\to0\),
    we obtain \(S=0\) and \(c=0\).
    For any finite sampling tuple of pairwise distinct points in \(\mathbb N\), its kernel matrix is
    \(\frac12 I+\mathbf a_{\mathbf x}\mathbf a_{\mathbf x}^{T}\),
    where \(\mathbf a_{\mathbf x}\) consists of the corresponding
    \(a_i\). This matrix is positive definite, so (K3) also holds.

    Consider the interpolation constraint \(f(1)=1\), and set
    \(r_i:=K(i,1)\). Then
    \[
        r_1=\frac32,\qquad r_i=2-\frac1i\quad(i\ge2).
    \]
    In particular, \(0<r_i<2\) for every \(i\), and \(r_i\to2\).
    If \(f=Tc\) satisfies \(f(1)=1\), then \(c\ne0\) and
    \[
        1=\bigl|\sum_i c_i r_i\bigr|
        \le\sum_i|c_i|r_i
        <2\sum_i|c_i|
        \le2\|c\|_p=2\|f\|_{B_p}.
    \]
    The strict inequality follows from
    \(\sum_i|c_i|(2-r_i)>0\).
    On the other hand, the functions
    \[
        f_i:=\frac{K(i,\cdot)}{r_i}
    \]
    satisfy \(f_i(1)=1\), and uniqueness of the coefficient
    representation gives \(\|f_i\|_{B_p}=r_i^{-1}\to\frac12\).
    Consequently,
    \[
        \inf\{\|f\|_{B_p}:f(1)=1\}=\frac12,
    \]
    but no feasible function attains this infimum.
\end{example}

Example~\ref{ex:interpolation-nonattainment} shows that additional
hypotheses beyond (K1)--(K3) are needed to guarantee existence for
all feasible interpolation problems. Such hypotheses may take different
forms. For the existence results below, we use the following sufficient
set of continuity and compactness or decay conditions:
\begin{description}
    \item[(K4)] Either \(X\) is a nonempty compact metric space and
          \(K(\cdot,x)\in C(X)\) for every \(x\in X\), or
          \(X=\bR^d\) and \(K(\cdot,x)\in C_0(\bR^d)\) for every \(x\in X\).
\end{description}
On compact \(X\), \(C_0(X)=C(X)\). On \(\bR^d\), the functions \(K(\cdot,x)\) in (K4) are continuous and vanish at infinity.
Theorem~\ref{K1K2pro} verifies (K4) for its translation-invariant
kernels. In particular, for each of the three displayed kernels,
\(K(\cdot,x)\) is continuous and vanishes at infinity for every \(x\).

\begin{remark}
    Assumption (K4) is one sufficient set of additional conditions for
    the existence results below. Its role is to provide the compactness
    and continuity needed to pass to a limit of a minimizing sequence.
    Other assumptions on the domain and the kernel may also ensure
    existence; (K4) is not intended to characterize all such cases.
\end{remark}

\begin{lemma}[Compactness for moving centers]\label{lem:moving-center-compactness}
    Let \(0<p<1\), and assume (K1), (K2), and (K4).
    Every sequence \(\{f_m\}\subset B_p\) bounded in the quasi-norm has a subsequence
    \(\{f_{m_k}\}\) and a function \(f\in B_p\) such that
    \[
        f_{m_k}(x)\longrightarrow f(x)\quad(x\in X),\qquad
        \|f\|_{B_p}^p\le\liminf_{m\to\infty}\|f_m\|_{B_p}^p.
    \]
\end{lemma}
\begin{proof}
    First pass to a subsequence along which $\|f_m\|_{B_p}^p$ converge
    to their original lower limit, and denote this subsequence again by \(\{f_m\}\). Write
    \[
        f_m=\sum_{j\ge1}c_{m,j}K(t_{m,j},\cdot),\qquad
        \sum_j|c_{m,j}|^p\le C,
    \]
    ordering the nonzero coefficients by decreasing magnitude. Pad finite
    expansions with zero coefficients and arbitrary centers; these formal
    zero entries are not required to be distinct. Then
    \begin{equation}\label{eq:sorted-coefficient-tail}
        |c_{m,j}|\le C^{1/p}j^{-1/p},\qquad
        \sup_m\sum_{j>N}|c_{m,j}|
        \le C^{1/p}\sum_{j>N}j^{-1/p}\longrightarrow0.
    \end{equation}
    The last limit uses \(p<1\).

    If \(X\) is compact, put \(\widehat X=X\). If \(X=\bR^d\), let
    \(\widehat X=\bR^d\cup\{\infty\}\) be its one-point
    compactification and set \(K(\infty,x)=0\). In both cases
    \(\widehat X\) is compact metrizable and, for every \(x\in X\),
    the extended function \(K(\cdot,x)\) is continuous. Successive subsequence extraction and a
    diagonal argument give, for every \(j\),
    \[
        c_{m_k,j}\to c_j,\qquad t_{m_k,j}\to t_j\in\widehat X.
    \]
    Fatou's lemma gives
    \[
        \sum_j|c_j|^p\le\liminf_k\sum_j|c_{m_k,j}|^p
        =\liminf_m\|f_m\|_{B_p}^p.
    \]
    Discard indices whose limiting center is \(\infty\), and combine
    coefficients whose limiting centers coincide:
    \[
        a(t):=\sum_{j:t_j=t}c_j,\qquad t\in X.
    \]
    These sums converge absolutely, and countable subadditivity yields
    \[
        \sum_{t\in X}|a(t)|^p\le\sum_{j:t_j\ne\infty}|c_j|^p
        \le\sum_j|c_j|^p.
    \]
    Thus \(a\in\ell^p(X)\) and \(f:=Ta\in B_p\).
    For each fixed \(x\), the first \(N\) summands converge by continuity
    on \(\widehat X\). Their remaining tails, and the limiting tail,
    are bounded in absolute value by
    \(M_x C^{1/p}\sum_{j>N}j^{-1/p}\).
    First passing to the subsequence limit and then letting \(N\to\infty\)
    proves \(f_{m_k}(x)\to f(x)\).
    Finally \(\|f\|_{B_p}^p=\|a\|_{\ell^p(X)}^p\) gives the claimed lower-semicontinuity inequality.
\end{proof}

\begin{theorem}\label{Theorem: minimal norm interpolation admits a minimizer}
    Let \(0<p<1\), and assume (K1), (K2), and (K4).
    Fix an ordered tuple \(\mathbf x=(x_1,\ldots,x_n)\) of pairwise
    distinct points in \(X\) and a target \(\mathbf y\in\reals^n\).
    If \(I_{\mathbf x}(\mathbf y)\ne\varnothing\), then the minimal norm
    interpolation problem \eqref{eq:min_norm_interp} admits a minimizer.
    If (K3) also holds, then \(I_{\mathbf x}(\mathbf y)\ne\varnothing\)
    for every \(\mathbf y\in\reals^n\), and hence a minimizer exists
    for every target.
\end{theorem}
\begin{proof}
    Assume that the set \(I_{\mathbf x}(\mathbf y)\) is not empty, and define
    \[
        M := \inf_{g \in I_{\mathbf x}(\mathbf y)} \|g\|_{B_p}^p < +\infty.
    \]
    Select a sequence \(\{f_m\}\) in \(I_{\mathbf x}(\mathbf y)\) such that
    \(\|f_m\|_{B_p}^p \to M\). According to Lemma~\ref{lem:moving-center-compactness}, there exists a subsequence that converges at each sample point to a function \(f \in B_p\) with
    \(\|f\|_{B_p}^p \leq M\). Since \(f(\mathbf x) = \mathbf y\), it follows that \(f \in I_{\mathbf x}(\mathbf y)\). By the definition of \(M\), we have \(\|f\|_{B_p}^p = M\), which establishes the existence of a minimizer.

    Moreover, if condition (K3) is satisfied, then for any \(\mathbf y \in \reals^n\), the function
    \[
        h := \sum_{j=1}^n [\mathrm{K}[\mathbf x]^{-1} \mathbf y]_j K(x_j, \cdot)
    \]
    lies in \(B_p\) and fulfills \(h(\mathbf x) = \mathbf y\).
    Therefore, every target vector \(\mathbf y\) is attainable, and the previous argument applies.
\end{proof}

\begin{theorem}[Existence for regularized learning]\label{thm:regularized-existence}
    Let \(0<p<1\), and assume (K1), (K2), and (K4).
    Every acceptable regularized learning scheme
    \eqref{eq:reg_learning_Bp} admits a minimizer in \(B_p\).
\end{theorem}
\begin{proof}
    Write \(J(g)=V(g(\bx))+\mu\phi(\|g\|_{B_p})\).
    The zero function has finite objective, so a minimizing sequence \(\{f_m\}\subset B_p\) may be chosen with uniformly bounded objectives. Since \(V\ge0\), \(\mu>0\), and \(\phi(r)\to+\infty\) as $r\to+\infty$, the quasi-norms of $\{f_m\}$ are bounded. Choose first a subsequence whose quasi-norms converge to their lower limit,
    and apply Lemma~\ref{lem:moving-center-compactness} to obtain
    some \(f\in B_p\) with convergent sample values and no larger limiting norm.
    Continuity of \(V\), and continuity and monotonicity of \(\phi\), give
    \[
        J(f)\le\liminf_m J(f_m)=\inf_{g\in B_p}J(g).
    \]
    Hence the infimum is attained.
\end{proof}

Next, we demonstrate that \( p \in (0,1) \) leads to sparsity in the solutions of the minimal norm interpolation problem within \( B_p \). For each integer \(r\ge1\), define
\[
    S_r:=\left\{\sum_{j=1}^{r}c_jK(t_j,\cdot):
    c_j\in\mathbb R,\ t_j\in X,\ 1\le j\le r\right\}.
\]
Zero coefficients are allowed and the centers need not be sample points.

\begin{lemma}[Sparsity after existence]\label{lemmasuppn}
    Let \(0<p<1\), and assume (K1)--(K2).
    Every minimizer of a feasible minimal norm interpolation problem
    belongs to \(S_{n}\).
\end{lemma}
\begin{proof}
Suppose, for contradiction, that \eqref{eq:min_norm_interp} admits a minimizer of the form 
    $$
    f_0(x)=\sum_{j=1}^\infty c_jK(t_j,x)
    $$
    with $|\supp(\mathbf{c})|>n$, where the centers $t_j$ corresponding to nonzero coefficients are pairwise distinct. As the vectors $(K(t_j,x_i):j\in\supp(\mathbf{c}))$, $1\le i\le n$, live in a space with dimension larger than $n$, and the dimension of
    $$
    \spa\{(K(t_j,x_i):j\in\supp(\mathbf{c})):\ 1\le i\le n\}
    $$
    is at most $n$, there exists a nontrivial vector $\mathbf{h}\in \ell^p$ with $\supp(\mathbf{h})\subseteq\supp(\mathbf{c})$ such that $|\supp(\mathbf{h})|<+\infty$ and
    $$
    \sum_{j=1}^\infty h_j K(t_j,x_i)=0,\ \forall 1\le i\le n.
    $$
    Now consider 
    $$
    f_\tau(x)=\sum_{j=1}^\infty (c_j+\tau h_j)K(t_j,x),\ \ \tau \in\bR.
    $$
    Then $f_\tau (\bx)=f_0(\bx)=\by$. Since $f_0$ is a minimizer of \eqref{eq:min_norm_interp}, we must have
    \begin{equation}\label{lemmasuppneq1}
        \|f_\tau\|_{B_p}^p\ge \|f_0\|_{B_p}^p\mbox{ for all }\tau \in\bR.
    \end{equation}
    Let 
    $$
    \Psi(\tau)=\|f_\tau\|_{B_p}^p=\sum_{j\in\supp(\mathbf{c})}|c_j+\tau h_j|^p,\ \ \tau \in\bR.
    $$
    By \eqref{lemmasuppneq1}, $\tau=0$ is a minimum point of $\Psi(\tau)$. 
    
    There exists a sufficiently small $\delta > 0$ such that $c_j + \tau h_j \neq 0$ for all $j\in\supp(\mathbf{c})\cap \supp(\mathbf{h})$ whenever $|\tau|<\delta$. For $0 < p < 1$, the function $x \mapsto |x|^p$ is twice continuously differentiable on $\reals \setminus \{0\}$ with a strictly negative second derivative
        $$
        \frac{d^2}{dx^2} |x|^p = p(p-1) |x|^{p-2} < 0,\ \ x\ne0.
        $$
        Therefore, for each $j\in\supp(\mathbf{c})\cap \supp(\mathbf{h})$, the function $\tau \mapsto |c_j + \tau h_j|^p$ is strictly concave on $\tau \in (-\delta, \delta)$, whereas terms with $h_j = 0$ remain constant. Since $\mathbf{h} \neq 0$, there exists at least one index $j\in\supp(\mathbf{c})\cap \supp(\mathbf{h})$. As a sum of concave functions containing at least one strictly concave term, $\Psi(\tau)$ is strictly concave on $(-\delta, \delta)$.       
        However, this implies that $\tau = 0$ cannot be a local minimizer of $\Psi(\tau)$, a contradiction.
    \end{proof}
    

\begin{corollary}[A sparse regularized minimizer]\label{cor:sparse-regularized}
    Under the assumptions of Theorem~\ref{thm:regularized-existence}, every acceptable regularized learning scheme \eqref{eq:reg_learning_Bp} in $B_p$ has a minimizer $f^*$ that belongs to the sparse set $S_n$.
\end{corollary}
\begin{proof}
    Let \(f^*\) be a regularized minimizer. The interpolation problem
    with target \(f^*(\bx)\) is feasible. Moreover, according to Theorem~\ref{Theorem: minimal norm interpolation admits a minimizer} and Lemma~\ref{lemmasuppn}, there exists a minimal norm interpolant \(g\) that satisfies the interpolation conditions, and \(g\) belongs to \(S_n\). Since \(g(\bx)=f^*(\bx)\), \(\|g\|_{B_p}\le\|f^*\|_{B_p}\),
    and \(\phi\) is nondecreasing, we have \(J(g)\le J(f^*)\). As \(f^*\) is a global minimizer, equality holds. Thus \(g\in S_n\) is also a global minimizer.
\end{proof}

The sparse representation above allows
centers anywhere in \(X\). We now characterize when a minimizer can
be chosen in the sample span \(S^{\mathbf x}\).

For an ordered tuple \(\mathbf x=(x_1,\ldots,x_n)\) of pairwise
distinct points, write
\[
    K_{\mathbf x}(t):=(K(t,x_1),\ldots,K(t,x_n))^{\intercal}.
\]

Assume (K1)--(K3).
The \emph{generalized Lebesgue constant} associated with the sampling
tuple \(\mathbf x\) is defined by
\begin{equation}\label{eq:generalized-lebesgue-constant}
    \Lambda_p(\mathbf x):=\sup_{t\in X}
    \|K[\bx]^{-1}K_{\mathbf x}(t)\|_p^p.
\end{equation}
This convention uses the \(p\)-th power of the coefficient quasi-norm.

For comparison with the endpoint \(p=1\), define
\begin{equation}\label{eq:classical-lebesgue-coefficients}
    \Lambda_1(\mathbf x):=\sup_{t\in X}
    \|\mathrm K[\bx]^{-1}K_{\mathbf x}(t)\|_1.
\end{equation}

To relate this definition to classical interpolation, set
\[
    \ell_j(t):=[K[\bx]^{-1}K_{\mathbf x}(t)]_j,
    \qquad 1\le j\le n.
\]
These functions satisfy \(\ell_j(x_i)=\delta_{ij}\). If \(K\) is
symmetric, they form the cardinal basis of \(S^{\mathbf x}\), and
\[
    (\mathcal I_{\mathbf x}g)(t):=\sum_{j=1}^n g(x_j)\ell_j(t)
\]
is the interpolation operator on bounded functions. Its operator norm
with respect to the supremum norm is the classical Lebesgue constant
\(\Lambda_1(\mathbf x)\) from \eqref{eq:classical-lebesgue-coefficients}:
\[
    \Lambda_1(\mathbf x)=\sup_{t\in X}\sum_{j=1}^n|\ell_j(t)|.
\]
Thus \(\Lambda_p(\mathbf x)=\sup_t\sum_j|\ell_j(t)|^p\) extends
this expression by using the \(p\)-th power of the \(\ell^p\)
quasi-norm. In particular, for \(0<p<1\),
\[
    \Lambda_1(\mathbf x)
    \le\Lambda_p(\mathbf x)^{1/p}
    \le n^{1/p-1}\Lambda_1(\mathbf x).
\]
For a nonsymmetric kernel, the same cardinal functions instead belong
to \(\operatorname{span}\{K(\cdot,x_i):1\le i\le n\}\), so the
classical interpretation applies to interpolation with the transposed
kernel.

\begin{theorem}\label{representerp<1}
    Let \(0<p<1\) and assume (K1)--(K3). Fix an ordered tuple
    \(\mathbf x=(x_1,\ldots,x_n)\) of pairwise distinct points in \(X\).
    Every acceptable learning problem based on \(\mathbf x\) has a
    minimizer in \(S^{\mathbf x}\) if and only if
    \begin{equation}\label{eq:coefficient_bound1}
        \Lambda_p(\mathbf x)=1.
    \end{equation}
    Consequently, \(B_p\) satisfies the linear representer theorem
    if and only if this condition holds for every sampling tuple.
\end{theorem}
\begin{proof}
According to Lemma \ref{equivalentRLMI}, it suffices to demonstrate that the equality
        \begin{equation}\label{eq: Interpolation equation}
            \min_{f \in I_{\mathbf{x}}(\mathbf{y}) \cap S^{\mathbf{x}}} \|f\|_{B_p}^p =  \min_{f \in I_{\mathbf{x}}(\mathbf{y})} \|f\|_{B_p}^p
        \end{equation}
        holds for all $\mathbf{y} \in \reals^n$ if and only if condition \eqref{eq:coefficient_bound1} is satisfied. For any $\mathbf{y} \in \reals^n$, the intersection $I_{\mathbf{x}}(\mathbf{y}) \cap S^{\mathbf{x}}$ contains a unique element $f_0 = K^{\mathbf{x}}(\cdot) \mathrm{K}[\bx]^{-1} \mathbf{y}$, where $K^{\mathbf{x}}(\cdot) := (K(x_j,\cdot), \ j \in \naturals_n)$. Letting $\mathbf{a} := \mathrm{K}[\bx]^{-1} \mathbf{y} \in \reals^n$, we have
        \[
        \min_{f \in I_{\mathbf{x}}(\mathbf{y}) \cap S^{\mathbf{x}}} \|f\|_{B_p}^p = \|f_0\|_{B_p}^p = \|\mathbf{a}\|_{p}^p.
        \]
        Note that $\mathrm{K}[\mathbf{x}]^{-1} K_{\mathbf{x}}(x_i) = \mathbf{e_i},$ where $ \mathbf{e_i} $ denotes the standard basis vector of $\reals^n$. Consequently, 
        \begin{equation}\label{lambdabxge1}
       \Lambda_p(\mathbf x)\ge 1.
        \end{equation}

        Suppose \eqref{eq: Interpolation equation} holds for any $\by\in\reals^n$. For any $t \in X \setminus \bx$ and any $\alpha > 0$, we construct a specific function $g \in I_{\mathrm {x}}(\mathbf{y}) \cap S^{\mathbf{x} \cup \{t\}}$ of the form
        \begin{equation*}
            g = \sum_{j=1}^n b_j K(x_j,\cdot) + \alpha K(t, \cdot),
        \end{equation*}
        where $\mathbf{b} = (b_j)_{j=1}^n \in \reals^n$. The interpolation condition $g(\bx) = \mathbf{y}$ implies $\mathrm{K}[\bx] \mathbf{b} + \alpha K_{\mathbf{x}}(t) = \mathbf{y}$, which yields $\mathbf{b} = \mathbf{a} - \alpha \mathbf{c}$ with $\mathbf{c} := \mathrm{K}[\bx]^{-1} K_{\mathbf{x}}(t) \in \reals^n$. Thus, the quasi-norm of $g$ is
        \[
        \|g\|_{B_p}^p = \|\mathbf{a} - \alpha \mathbf{c}\|_{p}^p + |\alpha|^p.
        \]
        Applying \eqref{eq: Interpolation equation} to $g \in I_{\mathbf{x}}(\mathbf{y})$, we obtain
        \[
        \|\mathbf{a}\|_{p}^p \le \|\mathbf{a} - \alpha \mathbf{c}\|_{p}^p + |\alpha|^p.
        \]
        Since $\mathrm{K}[\bx]$ is invertible, and the above inequality holds for any $\mathbf{y} \in \reals^n$ and $\alpha > 0$, by setting $\mathbf{y} = \alpha K_{\mathbf{x}}(t)$, we obtain $\mathbf{a} = \alpha \mathbf{c}$. Substituting this into the inequality yields
        \[
        |\alpha|^p \|\mathbf{c}\|_{p}^p \le  |\alpha|^p
        \]
        and thus
        \begin{equation*}
        \|\mathrm{K}[\bx]^{-1} K_{\mathbf{x}}(t)\|_{p}^p \le 1,
        \end{equation*}
        which together with \eqref{lambdabxge1} proves the necessity.
        
        Conversely, assume \eqref{eq:coefficient_bound1} holds. Let $\mathbf{y} \in \reals^n$ and $f_0 = K^{\mathbf{x}}(\cdot) \mathrm{K}[\bx]^{-1} \mathbf{y} \in I_{\mathbf{x}}(\mathbf{y}) \cap S^{\mathbf{x}}$ with $\mathbf{a} := \mathrm{K}[\bx]^{-1} \mathbf{y}$. Any function $g \in I_{\mathbf{x}}(\mathbf{y})$ can be represented as
        \begin{equation*}
        g = \sum_{j=1}^n b_j K(x_j,\cdot) + \sum_{j=1}^\infty \alpha_j K(t_j, \cdot),
        \end{equation*}
        where $\{t_j\}_{j=1}^\infty \subset X \setminus \bx$. The interpolation condition $g(\bx) = \mathbf{y}$ implies 
        $$
        \mathrm{K}[\bx] \mathbf{b} + \sum_{j=1}^\infty \alpha_j K_{\mathbf{x}}(t_j) = \mathbf{y},
        $$
        which gives $\mathbf{b} = \mathbf{a} - \sum_{j=1}^\infty \alpha_j \mathbf{c^{(j)}}$ with $\mathbf{c^{(j)}} := \mathrm{K}[\bx]^{-1} K_{\mathbf{x}}(t_j)$. The norm of $g$ is 
        \[
        \|g\|_{B_p}^p = \|\mathbf{b}\|_{p}^p + \sum_{j=1}^\infty |\alpha_j|^p.
        \]
        
        For $i > n$, we assign $a_i = b_i = c^{(j)}_i = 0$. Combining with $\mathbf{a} = \mathbf{b} + \sum_{j=1}^\infty \alpha_j \mathbf{c}^{(j)}$, it follows that
        \begin{equation*}
        a_i = b_i + \sum_{j=1}^\infty \alpha_j c^{(j)}_i, \quad \forall i \ge 1.
        \end{equation*}
        By Lemma \ref{lemma:Infinite Subadditivity of Quasi-Norm}, 
        \begin{align*}
        \|f_0\|_{B_p}^p &= \|\mathbf{a}\|_{p}^p = \biggl\| \mathbf{b} + \sum_{j=1}^\infty \alpha_j \mathbf{c}^{(j)} \biggr\|_{p}^p \nonumber \\
        &\le \|\mathbf{b}\|_{p}^p + \sum_{j=1}^\infty \|\alpha_j \mathbf{c}^{(j)}\|_{p}^p \nonumber \\
        &= \|\mathbf{b}\|_{p}^p + \sum_{j=1}^\infty |\alpha_j|^p \|\mathbf{c}^{(j)}\|_{p}^p.
        \end{align*}
        Since $\|\mathbf{c}^{(j)}\|_{p}^p \le 1$ for all $j \ge 1$ by assumption, we obtain
        \begin{align}
        \label{eq:ineq_chain for bound1}
        \|f_0\|_{B_p}^p &\le \|\mathbf{b}\|_{p}^p +  \sum_{j=1}^\infty |\alpha_j|^p 
        =\|g\|_{B_p}^p.
        \end{align}
        Taking the infimum of the right-hand side of \eqref{eq:ineq_chain for bound1} over all $g \in I_{\mathbf{x}}(\mathbf{y})$, we conclude that
        \begin{equation*}
        \min_{f \in I_{\mathbf{x}}(\mathbf{y}) \cap S^{\mathbf{x}}} \|f\|_{B_p}^p \le \min_{f \in I_{\mathbf{x}}(\mathbf{y})} \|f\|_{B_p}^p.
        \end{equation*}
        The reverse direction of the inequality is clear as $I_{\mathbf{x}}(\mathbf{y}) \cap S^{\mathbf{x}} \subseteq I_{\mathbf{x}}(\mathbf{y}).$
        \end{proof}

Introduce an interpolation operator on $B_p$ defined by
    \begin{equation}\label{interpolationQ}
        Q_{\mathbf x}f:=\sum_{j=1}^n[K[\bx]^{-1}f(\bx)]_jK(x_j,\cdot).
    \end{equation}
It follows from infinite
subadditivity and testing the functions \(K(t,\cdot)\) that
\[
    \Lambda_p(\mathbf x)
    =\left(\sup_{\|f\|_{B_p}\le1}
    \|Q_{\mathbf x}f\|_{B_p}\right)^p.
\]
This identity explains
the role of the generalized Lebesgue constant in controlling the
coefficient quasi-norm under interpolation and in the relaxed
representer theorem of Section~\ref{sec:relaxed-small-p}.

We have observed in the proof of Theorem \ref{representerp<1} that  \(\Lambda_1(\mathbf x)\) is at least one. Kernels for which this constant exactly equals
one have been studied in \cite{Lin2026,Lin2021,Song2013}.
For \(0<p<1\), \eqref{eq:coefficient_bound1} fails to be true for these kernels, as shown in the following calculations.

First, let \(X=(0,1)\), and take either the generalized Brownian bridge
kernel
\[
    K(s,x)=\min\{s,x\}-\tau sx,\qquad 0\le\tau\le1,
\]
or the triangular kernel \(K(s,x)=1-|s-x|\).

Both kernels are bounded, so (K1) holds. To verify (K2), let
\(c\in\ell^p((0,1))\), put
\(\mu=\sum_t c(t)\delta_t\), and suppose \(Tc=0\).
Since \(c\in\ell^1((0,1))\) and the kernels are bounded, the
synthesis series converges uniformly. Its second derivative in the
sense of distributions on \((0,1)\) is therefore obtained termwise:
\[
    -(Tc)''=\mu\quad\text{for the Brownian bridge kernel},\qquad
    -(Tc)''=2\mu\quad\text{for the triangular kernel}.
\]
Thus \(\mu=0\), and \(c(t)=\mu(\{t\})=0\) for every \(t\), proving (K2).

For (K3), the Brownian bridge kernel has the representation
\[
    K(s,x)=\int_0^1
    (\mathbf 1_{[0,s]}(u)-s)(\mathbf 1_{[0,x]}(u)-x)\,du
    +(1-\tau)sx.
\]
For distinct points \(x_j\in(0,1)\), the functions
\(\mathbf 1_{[0,x_j]}-x_j\) are linearly independent in \(L^2(0,1)\):
a linear combination that vanishes almost everywhere has zero jump
at each \(x_j\), forcing every coefficient to vanish.
The integral term consequently gives a strictly positive quadratic
form, and the remaining term is nonnegative. For the triangular kernel,
\[
    K(s,x)=\int_{\bR}
    \mathbf 1_{[s-1/2,s+1/2]}(u)
    \mathbf 1_{[x-1/2,x+1/2]}(u)\,du,
    \qquad s,x\in(0,1).
\]
The interval indicators for distinct centers are linearly independent:
order their left endpoints and examine a short interval immediately
to the right of the first one; its coefficient must be zero.
Repeating this argument proves independence. Hence the triangular
kernel also has strictly positive definite finite matrices.
This verifies (K3) for both kernels.

Choose \(0<a<b<1\) and \(a<t<b\), and set \(\mathbf x=(a,b)\).
For either kernel, each component of \(K_{\mathbf x}(t)\) is affine
in \(t\) on \([a,b]\). The sample matrix is nonsingular: its
determinant is \(a(b-a)(1-\tau b)>0\) for the Brownian bridge kernel,
and \(1-(1-b+a)^2>0\) for the triangular kernel. Since the coefficient
vector equals \(\mathbf e_1\) at \(a\) and \(\mathbf e_2\) at \(b\),
\[
    \mathrm K[\bx]^{-1}K_{\mathbf x}(t)
    =\left(\frac{b-t}{b-a},\frac{t-a}{b-a}\right)^{\intercal}.
\]
Both entries belong to \((0,1)\) and sum to one. However,
\[
    \|\mathrm K[\bx]^{-1}K_{\mathbf x}(t)\|_p^p
    =\left(\frac{b-t}{b-a}\right)^p
    +\left(\frac{t-a}{b-a}\right)^p>1,\ \ t\in(a,b).
\]

Next, let \(X=\bR\) and \(K(s,x)=e^{-|s-x|}\).
Its spectral density in \eqref{eq:fourier_repr} is
\(\varphi(\xi)=[\pi(1+\xi^2)]^{-1}\), which is integrable and
strictly positive. Theorem~\ref{K1K2pro} and the quadratic-form
argument following (K3) verify (K1)--(K3).
Choose samples
\(\mathbf x=(a,b)\) with \(a<b\), and put \(h=b-a\).
For \(t=a+\varepsilon\), where \(0<\varepsilon<h\), direct inversion gives
\[
    \mathrm K[\bx]^{-1}K_{\mathbf x}(a+\varepsilon)
    =\left(\frac{\sinh(h-\varepsilon)}{\sinh h},
    \frac{\sinh\varepsilon}{\sinh h}\right)^{\intercal}.
\]
As \(\varepsilon\downarrow0\), its coefficient cost satisfies
\[
    \|\mathrm K[\bx]^{-1}K_{\mathbf x}(a+\varepsilon)\|_p^p
    =1+\frac{\varepsilon^p}{(\sinh h)^p}+o(\varepsilon^p)>1
\]
for all sufficiently small positive \(\varepsilon\); here
\(\varepsilon=o(\varepsilon^p)\) because \(p<1\).
Thus \eqref{eq:coefficient_bound1} fails for these two-point configurations.

The preceding discussion naturally motivates the development of a relaxed representer theorem that does not affect the generalization ability of acceptable learning schemes.

\section{Relaxed Representer Theorems for \(0<p<1\)}
\label{sec:relaxed-small-p}

This section develops relaxed representer theorems for \(0<p<1\),
comparing learning over \(B_p\) with learning over the sample span
through an adjustment of the regularization parameter. We identify
the smallest factor as a generalized Lebesgue constant and estimate
its growth with the sample size. Under uniform stability of the
sample matrices, algebraic kernel decay yields three growth regimes,
while exponential decay gives a uniform upper bound independently of the
sample size. These results extend the relaxed representer theorem
considered for \(p=1\) in \cite{Song2013}. In the subsequent analysis, \(\|\cdot\|\) denotes the matrix norm induced by the Euclidean norm. 

Throughout this section, assume (K1)--(K2). Fix an ordered tuple
\(\mathbf x=(x_1,\ldots,x_n)\) of pairwise distinct points for which
\(\mathrm K[\mathbf x]\) is nonsingular.

In the relaxed representer theorem below, the penalty is \(\|f\|_{B_p}^p\).
The general nondecreasing penalties of Section 3 are retained for the
existence results and the linear representer theorem.

\begin{defn}\label{def:relaxed-representer}
    The relaxed representer theorem holds at \(\mathbf x\) with
    factor \(\beta_n\ge1\) if, for every continuous function
    \(V:\reals^n\to[0,\infty)\) and every \(\mu>0\),
    \begin{equation}\label{relaxedbetan}
        \min_{f\in S^{\mathbf x}}
        \{V(f(\mathbf x))+\mu\|f\|_{B_p}^p\}
        \le\inf_{f\in B_p}
        \{V(f(\mathbf x))+\mu\beta_n\|f\|_{B_p}^p\}.
    \end{equation}
    For a prescribed family of sampling tuples, the theorem holds
    uniformly if the same \(\beta_n\) applies to every tuple of size \(n\)
    in that family.
\end{defn}

By (K2), the quasi-norm of \(\sum_{j=1}^n c_jK(x_j,\cdot)\)
is \(\|\mathbf c\|_p\). The coefficient objective on the left of
\eqref{relaxedbetan} is therefore continuous and coercive on
\(\reals^n\), so its minimum is attained.
The infimum on the right need not be attained. If it is attained,
for example under (K4), it may be written as a minimum.
When \(\beta_n=1\), the inclusion \(S^{\mathbf x}\subseteq B_p\)
forces equality and hence gives an exact sample-span minimizer.
A factor greater than one compares the two problems at different
regularization parameters.

The following discussion of minimizers presupposes their existence.
If the quantity \(\beta_n\) in (\ref{relaxedbetan}) does not grow too fast as \(n\) tends to infinity, for example, \[ \beta_n\mbox{ is bounded,}\ \beta_n=O(\log n),\mbox{ or }\beta_n=O(n^\gamma) \] for some small $\gamma > 0$, then the minimizer in the whole space \(B_p\) and the one in its finite-dimensional subspace \(S^{\mathbf{x}}\) \[ \arg\min_{f \in S^{\mathbf{x}}}\left\{ V(f(\bx)) + \mu \|f\|_{B_p}^p \right\},\ \ \arg\min_{f \in B_p} \left\{ V(f(\bx)) + \mu \beta_n \|f\|_{B_p}^p \right\} \] will have comparable generalization ability \cite{Song2011}.

\subsection{The coefficient criterion}

\begin{lemma}\label{pro:Sufficient condition for relaxed linear representer theorem}
    For \(\beta_n\ge1\), the relaxed representer theorem
    \eqref{relaxedbetan} holds at \(\mathbf x\) if and only if
    \begin{equation}\label{eq:Condition for the Relaxed Representer Theorem}
        \min_{f\in I_{\mathbf x}(\mathbf y)\cap S^{\mathbf x}}
        \|f\|_{B_p}^p
        \le\beta_n\inf_{f\in I_{\mathbf x}(\mathbf y)}\|f\|_{B_p}^p,
        \qquad \mathbf y\in\reals^n.
    \end{equation}
\end{lemma}
\begin{proof}
    Suppose \eqref{eq:Condition for the Relaxed Representer Theorem}
    holds. For each \(f\in B_p\), let \(g\in S^{\mathbf x}\) be the
    unique interpolant of \(f(\mathbf x)\). Then
    \(\|g\|_{B_p}^p\le\beta_n\|f\|_{B_p}^p\), so
    \[
        V(g(\mathbf x))+\mu\|g\|_{B_p}^p
        \le V(f(\mathbf x))+\mu\beta_n\|f\|_{B_p}^p.
    \]
    Taking the full-space infimum proves \eqref{relaxedbetan}.

    Conversely, suppose \eqref{relaxedbetan} holds. Fix
    \(\mathbf y\in\reals^n\) and \(h\in I_{\mathbf x}(\mathbf y)\).
    For each integer \(m\ge1\), let \(\mathbf c_m\) minimize
    \[
        m\|\mathrm{K}[\bx]\mathbf c-\mathbf y\|_2^2+\|\mathbf c\|_p^p
    \]
    over \(\reals^n\). By \eqref{relaxedbetan},
    \[
        m\|\mathrm{K}[\bx]\mathbf c_m-\mathbf y\|_2^2
        +\|\mathbf c_m\|_p^p\le\beta_n\|h\|_{B_p}^p.
    \]
    Hence \(\mathrm{K}[\bx]\mathbf c_m\to\mathbf y\) as $m\to\infty$, and
    nonsingularity gives \(\mathbf c_m\to \mathrm{K}[\bx]^{-1}\mathbf y\).
    Passing to the limit and taking the infimum over \(h\) proves
    \eqref{eq:Condition for the Relaxed Representer Theorem}.
\end{proof}

\begin{theorem}\label{thm:relaxed-coefficient-criterion}
    The generalized Lebesgue constant defined in
    \eqref{eq:generalized-lebesgue-constant} satisfies
    \(1\le\Lambda_p(\mathbf x)<\infty\). The relaxed representer
    theorem holds at \(\mathbf x\) with factor \(\beta_n\ge1\) if and
    only if
    \begin{equation}\label{eq:coefficient_bound}
        \Lambda_p(\mathbf x)\le\beta_n.
    \end{equation}
    Thus \(\Lambda_p(\mathbf x)\) is the smallest factor in the relaxed representer theorem.
\end{theorem}
\begin{proof}
    By (K1), the components of \(K_{\mathbf x}(t)\) are bounded
    uniformly in \(t\), so \(\Lambda_p(\mathbf x)<\infty\).
    Since \(\mathrm{K}[\bx]^{-1}K_{\mathbf x}(x_i)=\mathbf e_i\),
    we also have \(\Lambda_p(\mathbf x)\ge1\).
    If the relaxed representer theorem holds, then
    Lemma~\ref{pro:Sufficient condition for relaxed linear representer theorem}
    gives \eqref{eq:Condition for the Relaxed Representer Theorem}.
    Apply this inequality to \(\mathbf y=K_{\mathbf x}(t)\). The function \(K(t,\cdot)\)
    is feasible and has quasi-norm one by (K2), giving
    \eqref{eq:coefficient_bound}.

    Conversely, assume \eqref{eq:coefficient_bound}, let
    \(f=Ta\in B_p\), and let \(Q_{\mathbf x}\) be
    the interpolation operator \eqref{interpolationQ}. The absolutely convergent coefficient
    identity and infinite subadditivity give
    \[
        \|Q_{\mathbf x}f\|_{B_p}^p
        \le\sum_{t\in X}|a(t)|^p
        \|\mathrm{K}[\bx]^{-1}K_{\mathbf x}(t)\|_p^p
        \le\beta_n\|f\|_{B_p}^p.
    \]
    Taking the infimum over all interpolants of a given target proves
    \eqref{eq:Condition for the Relaxed Representer Theorem}.
    Lemma~\ref{pro:Sufficient condition for relaxed linear representer theorem}
    then yields the relaxed representer theorem.
\end{proof}

The factor for a fixed sampling tuple is therefore finite under (K1).
Uniform estimates over growing sample sets require quantitative control
of the sample matrices. We obtain such estimates below under a uniform
stability assumption.

\subsection{Algebraic decay and uniform stability}

\begin{lemma}[Uniform algebraic inverse decay]
    \label{lemma:Jaffard's Theorem}
    Let \(\alpha>1\). An infinite matrix \(\mathrm A\), indexed by \(\mathbb N\),
    with \(|\mathrm A_{ij}|\le M(1+|i-j|)^{-\alpha}\), which is boundedly
    invertible on \(\ell^2\), has an inverse with the same algebraic
    decay. Moreover, if a family of nonsingular finite matrices
    \(\mathrm A_n\in\reals^{n\times n}\) satisfies
    \[
        |(\mathrm A_n)_{ij}|\le M(1+|i-j|)^{-\alpha},\qquad
        \sup_{n\ge1}\|\mathrm A_n^{-1}\|\le L,
    \]
    then
    \[
        |(\mathrm A_n^{-1})_{ij}|\le C(1+|i-j|)^{-\alpha},
    \]
    where \(C\) depends only on \(\alpha,M,L\).
\end{lemma}
\begin{proof}
    The quantitative form of Jaffard's theorem controls the algebraic
    decay bound of an inverse in terms of the original decay bound and
    its inverse operator norm; see \cite{Grochenig2014}.
    For finite \(\mathrm A_n\), extend it by the identity on the complement of
    \(\{1,\ldots,n\}\) in \(\mathbb Z\). The resulting block diagonal
    operator has decay bound at most \(\max\{M,1\}\) and inverse norm
    at most \(\max\{L,1\}\). Applying that theorem and restricting
    to the finite block gives the uniform conclusion. Extending an
    operator on \(\ell^2(\mathbb N)\) by the identity on the remaining
    indices gives the infinite version in the same way.
\end{proof}

In the next two theorems, the center domain is \(X=\mathbb R\).
The sampling points form an increasing sequence \(\{x_j\}_{j\ge1}\)
with \(x_{j+1}-x_j\ge h>0\), and
\[
    \mathbf x_n=(x_1,\ldots,x_n),\qquad
    \mathrm A_n:=\mathrm K[\mathbf x_n].
\]
We impose uniform stability of the finite sample matrices:
\begin{equation}\label{eq:uniform-sample-stability}
    \mathrm A_n\text{ is nonsingular for every }n,\qquad
    \sup_{n\ge1}\|\mathrm A_n^{-1}\|\le L<\infty.
\end{equation}
This condition holds, for example, if each \(\mathrm A_n\) is real symmetric and
\(\mathbf {c}^{\mathsf{T}}\mathrm A_n\mathbf c\ge m_0\|\mathbf c\|_2^2\) with a constant
\(m_0 >0\) independent of \(n\). One may then take \(L=m_0^{-1}\).

\begin{theorem}\label{Thm :Lebesgue constant satisfying algebraic decay conditions}
    Let \(X=\mathbb R\), and let \(\{x_j\}_{j\ge1}\) be an increasing
    sampling sequence with \(x_{j+1}-x_j\ge h>0\).
    Put \(\mathbf x_n=(x_1,\ldots,x_n)\) and
    \(\mathrm A_n=\mathrm K[\mathbf x_n]\).
    Assume \eqref{eq:uniform-sample-stability} and
    \[
        |K(u,v)|\le C_0(1+|u-v|)^{-\alpha},
        \qquad u,v\in\mathbb R,
    \]
    where \(\alpha>1\). There is a constant \(C\), depending only on
    \(p,\alpha,h,C_0,L\), such that
    \[
        \Lambda_p(\mathbf x_n)\le C
        \begin{cases}
            1,                  & \alpha p>1,   \\
            1+\log n,          & \alpha p=1,   \\
            n^{1-\alpha p}, & 0<\alpha p<1.
        \end{cases}
    \]
\end{theorem}
\begin{proof}
    The spacing bound gives \(|x_i-x_j|\ge h|i-j|\), so
    \(|(\mathrm A_n)_{ij}|\le M(1+|i-j|)^{-\alpha}\), with
    \(M=C_0\max\{1,h^{-\alpha}\}\).
    Lemma~\ref{lemma:Jaffard's Theorem} therefore gives
    \(|(\mathrm A_n^{-1})_{ij}|\le C_1(1+|i-j|)^{-\alpha}\), uniformly in \(n\).

    Fix \(t\in\mathbb R\), and choose \(k\in\{1,\ldots,n\}\) such that $|t-x_k|=\min_{1\le j\le n}|t-x_j|$.
    Then \(h|j-k|\le|x_j-x_k|\le2|t-x_j|\). Consequently,
    \begin{equation}\label{eq:t_xj_decay}
        (1+|t-x_j|)^{-\alpha}
        \le D_h(1+|j-k|)^{-\alpha},\qquad
        D_h:=\max\{1,(2/h)^\alpha\}.
    \end{equation}
    Writing \(\mathbf c(t)=\mathrm A_n^{-1}K_{\mathbf x_n}(t)\), we obtain
    \begin{equation}\label{eq:ci_est_1}
        |c_i(t)|\le C_0C_1D_h
        \sum_{j=1}^n(1+|i-j|)^{-\alpha}(1+|j-k|)^{-\alpha}.
    \end{equation}
    For each \(j\), at least one of \(1+|i-j|\) and \(1+|j-k|\)
    is not less than \((1+|i-k|)/2\). Since \(\alpha>1\), summation gives
    \[
        \sum_{j=1}^n(1+|i-j|)^{-\alpha}(1+|j-k|)^{-\alpha}
        \le 2^{\alpha+1}
        \left(\sum_{r\in\mathbb Z}(1+|r|)^{-\alpha}\right)
        (1+|i-k|)^{-\alpha}.
    \]
    Hence \(|c_i(t)|\le C_2(1+|i-k|)^{-\alpha}\), with \(C_2\)
    independent of \(n,t\). It follows that
    \begin{equation}\label{eq:sum_estim_lp}
        \|\mathbf c(t)\|_p^p
        \le C_2^p\sum_{i=1}^n(1+|i-k|)^{-\alpha p}
        \le2C_2^p\sum_{r=1}^n r^{-\alpha p}.
    \end{equation}
    The final sum is bounded uniformly if \(\alpha p>1\), is at most
    \(1+\log n\) if \(\alpha p=1\), and satisfies
    \begin{equation}\label{eq:algebraic_regime_final}
        \sum_{r=1}^n r^{-\alpha p}
        \le1+\int_1^n u^{-\alpha p}\,du
        \le\frac{n^{1-\alpha p}}{1-\alpha p}
        \quad\text{if }0<\alpha p<1.
    \end{equation}
    Taking the supremum over \(t\) proves the asserted bounds.
\end{proof}

\subsection{Exponential decay}

\begin{lemma}[Uniform exponential inverse decay]
    \label{lemma:Exponential Off-Diagonal Decay of Matrix Inverses}
    Let \(\mathrm A_n\in\reals^{n\times n}\) be nonsingular matrices with
    \[
        |(\mathrm A_n)_{ij}|\le M e^{-\alpha|i-j|},\qquad
        \sup_n\|\mathrm A_n^{-1}\|\le L,
    \]
    where \(M,\alpha,L>0\). There exists \(\theta\in(0,\alpha/2)\),
    depending only on \(M,\alpha,L\), such that
    \begin{equation}\label{eq:exp_decay_inv}
        |(\mathrm A_n^{-1})_{ij}|\le2L e^{-\theta|i-j|}
    \end{equation}
    for every \(n,i,j\).
\end{lemma}
\begin{proof}
    Fix a column \(j\), and let
    \(\mathrm W=\operatorname{diag}(e^{\theta|i-j|})_{i=1}^n\).
    The entries of \(\mathrm B_n:=\mathrm W\mathrm A_n \mathrm W^{-1}\) satisfy
    \[
        |(\mathrm B_n-\mathrm A_n)_{ik}|
        \le M e^{-\alpha|i-k|}(e^{\theta|i-k|}-1).
    \]
    Both absolute row sums and absolute column sums are bounded by
    \[
        \eta(\theta):=2M\sum_{r=1}^{\infty}
        e^{-\alpha r}(e^{\theta r}-1).
    \]
    Schur's test gives \(\|\mathrm B_n-\mathrm A_n\|\le\eta(\theta)\).
    Since \(\eta(\theta)\to0\) as \(\theta\downarrow0\), we may choose
    \(0<\theta<\alpha/2\) so that \(L\eta(\theta)\le1/2\).
    The Neumann series \cite{Golub:2013} then yields \(\|\mathrm B_n^{-1}\|\le2L\).
    Because \((\mathrm B_n^{-1})_{ij}=e^{\theta|i-j|}(\mathrm A_n^{-1})_{ij}\),
    this proves \eqref{eq:exp_decay_inv}.
\end{proof}

\begin{theorem}\label{thm:exponential-lebesgue-bound}
    Let \(X=\mathbb R\), and let \(\{x_j\}_{j\ge1}\) be an increasing
    sampling sequence with \(x_{j+1}-x_j\ge h>0\).
    Put \(\mathbf x_n=(x_1,\ldots,x_n)\) and
    \(\mathrm A_n=\mathrm K[\mathbf x_n]\).
    Assume \eqref{eq:uniform-sample-stability} and
    \[
        |K(u,v)|\le C_0 e^{-\gamma|u-v|},
        \qquad u,v\in\mathbb R,
    \]
    where \(\gamma>0\). There is a constant \(C\), depending only on
    \(p,h,C_0,\gamma,L\), such that
    \(\Lambda_p(\mathbf x_n)\le C\) for every \(n\).
\end{theorem}
\begin{proof}
    The entries of \(\mathrm A_n\) are bounded by
    \(C_0e^{-\gamma h|i-j|}\). The preceding lemma gives
    \(|(\mathrm A_n^{-1})_{ij}|\le2Le^{-\theta|i-j|}\), with
    \(0<\theta<\gamma h/2\) independent of \(n\).
    Fix \(t\in\mathbb R\), and choose \(k\in\{1,\ldots,n\}\) such that $|t-x_k|=\min_{1\le j\le n}|t-x_j|$.
    As in \eqref{eq:t_xj_decay}, \(|t-x_j|\ge h|j-k|/2\), so
    \begin{equation}\label{eq:ci_est_sum}
        |[\mathrm A_n^{-1}K_{\mathbf x_n}(t)]_i|
        \le2LC_0\sum_{j=1}^n
        e^{-\theta|i-j|}e^{-\gamma h|j-k|/2}.
    \end{equation}
    Put \(\delta=\gamma h/2-\theta>0\). The triangle inequality gives
    \[
        e^{-\theta|i-j|}e^{-\gamma h|j-k|/2}
        \le e^{-\theta|i-k|}e^{-\delta|j-k|}.
    \]
    Hence the left side of \eqref{eq:ci_est_sum} is at most
    \(C_*e^{-\theta|i-k|}\), where
    \(C_*=2LC_0\sum_{r\in\mathbb Z}e^{-\delta|r|}\).
    Therefore
    \[
        \|\mathrm A_n^{-1}K_{\mathbf x_n}(t)\|_p^p
        \le C_*^p\sum_{r\in\mathbb Z}e^{-p\theta|r|}
        =C_*^p\frac{1+e^{-p\theta}}{1-e^{-p\theta}}.
    \]
    This bound is independent of \(n,t\), proving the claim.
\end{proof}

By Theorem~\ref{thm:relaxed-coefficient-criterion}, the preceding
estimates give factors in the relaxed representer theorem bounded uniformly in \(n\) when
\(\alpha p>1\) in the algebraic case, and for every \(0<p<1\) in the
exponential case. In the remaining algebraic regimes, they give
logarithmic or polynomial growth. The constants are uniform over
families sharing the same decay, spacing, and stability bounds.

\section{RKBSs with \(\ell^p\) norms for \(p>1\)}
\label{sec:p-greater-one}

This section studies kernel-based learning in RKBSs with \(\ell^p\)
coefficient norms for \(1<p<\infty\). We first motivate the use of a
fixed countable dictionary and construct the associated RKBS, including
its reproducing pairing and minimizer-existence results. We then give
sufficient conditions for translation-invariant kernels to satisfy the
construction assumptions. Finally, we characterize the optimal factor
in the relaxed representer theorem, observe an obstruction to the
exact representer theorem, and derive a uniform relaxed factor for a
separated discrete dictionary under decay and stability assumptions.

For \(p>1\), the freedom to choose arbitrary centers is incompatible,
in general, with bounded point evaluations under the \(\ell^p\)
coefficient norm. The following example uses finite expansions to motivate
the prescribed dictionary in this section.

\begin{example}[Unrestricted centers and unbounded evaluation]
    \label{ex:p-greater-one-clustered-centers}
    Let \(K(t,x)=e^{-(t-x)^2}\) on \(\mathbb R\), and fix \(p>1\).
    Suppose a function space contains every finite expansion with
    distinct centers and assigns it the \(\ell^p\) norm of its
    coefficients. Fix \(\varepsilon>0\). For each \(m\ge1\), choose
    distinct points \(t_{m,1},\ldots,t_{m,m}\in(-\varepsilon,\varepsilon)\)
    and set
    \[
        f_m:=m^{-1/p}\sum_{j=1}^m K(t_{m,j},\cdot).
    \]
    Each coefficient vector has \(\ell^p\) norm one, whereas
    \[
        f_m(0)=m^{-1/p}\sum_{j=1}^m e^{-t_{m,j}^2}
        \ge e^{-\varepsilon^2}m^{1-1/p}\longrightarrow+\infty,\ \ m\to\infty.
    \]
    Thus evaluation at zero is unbounded, and such a space cannot be
    an RKBS with the prescribed coefficient norm. The obstruction
    already involves finite expansions with pairwise distinct centers.
\end{example}

To obtain an RKBS while retaining the \(\ell^p\) coefficient norm,
we restrict the available centers and impose the dual summability
condition below. A fixed countable dictionary is one such choice;
countability alone does not ensure bounded evaluations. The nonzero
coefficients, and hence the active centers within this dictionary,
may vary from one function to another.

\subsection{A fixed countable dictionary and the reproducing pairing}

Fix \(1<p<\infty\), put \(q=p/(p-1)\), and let \(X\) be a nonempty
set. Let \(D=\{\eta_j:j\ge1\}\subseteq X\) be a fixed countably
infinite set of pairwise distinct centers. Write
\[
    \ell^p(D):=\left\{c:D\to\reals:
    \sum_{t\in D}|c(t)|^p<\infty\right\},\qquad
    \|c\|_{\ell^p(D)}:=\left(\sum_{t\in D}|c(t)|^p\right)^{1/p}.
\]
For \(K:X\times X\to\reals\), assume:
\begin{description}
    \item[(D1)] For every \(x\in X\), the function
          \(k_x:D\to\reals\), \(k_x(t)=K(t,x)\), belongs to \(\ell^q(D)\).
    \item[(D2)] The synthesis map
          \[
              (T_Dc)(x):=\sum_{t\in D}c(t)K(t,x),\qquad x\in X,
          \]
          is injective on \(\ell^p(D)\).
\end{description}
Define
\[
    B_p^D:=T_D(\ell^p(D)),\qquad
    \|T_Dc\|_{B_p^D}:=\|c\|_{\ell^p(D)}.
\]
By (D1) and H\"older's inequality, the synthesis series converges
absolutely at every \(x\), and
\[
    |(T_Dc)(x)|\le\|k_x\|_{\ell^q(D)}\|c\|_{\ell^p(D)}.
\]
Condition (D2) makes the norm well-defined. The map \(T_D\) is a
linear isometry onto \(B_p^D\), so \(B_p^D\) is complete. The
preceding bound shows that every evaluation is continuous. Therefore \(B_p^D\) is an RKBS on \(X\).

The partner space is the Banach space \(\ell^q(D)\). Define
\[
    \langle T_Dc,g\rangle_{B_p^D\times\ell^q(D)}
    :=\sum_{t\in D}c(t)g(t),
    \qquad c\in\ell^p(D),\quad g\in\ell^q(D).
\]
This is well-defined by (D2), and H\"older's inequality gives
\[
    |\langle T_Dc,g\rangle|
    \le\|T_Dc\|_{B_p^D}\|g\|_{\ell^q(D)}.
\]
Moreover,
\[
    \langle f,k_x\rangle=f(x),\qquad f\in B_p^D,\quad x\in X.
\]
Hence the restriction \(K|_{D\times X}\) is a reproducing kernel
for \(B_p^D\) with this pairing.

Conditions (D1)--(D2) replace (K1)--(K2) for this fixed-dictionary
construction.
For the remainder of this section, write \(B_p=B_p^D\), and take
all sampling points and all expansion centers in \(D\). The learning
and interpolation definitions from Section 3 are used with these
restrictions, and the relaxed representer theorem is understood in
the sense of \eqref{relaxedbetan}, with the present value of \(p\).
Whenever (K3) is required below, it is imposed on tuples of distinct
points in \(D\). The dictionary remains fixed throughout each problem.

\begin{proposition}[Existence]
    \label{prop:fixed-dictionary-existence}
    Under (D1)--(D2), every feasible minimal norm interpolation problem
    in \(B_p^D\) has a unique minimizer. Every acceptable regularized
    learning problem has a minimizer. If the sample matrix is
    nonsingular, every target is feasible.
\end{proposition}
\begin{proof}
    The isometric isomorphism with \(\ell^p(D)\) makes \(B_p^D\) reflexive and
    strictly convex. A norm-bounded sequence has a weakly convergent
    subsequence, and bounded point evaluations preserve the sample
    values in this limit. The norm is weakly lower semicontinuous,
    so a minimizing sequence for a feasible interpolation problem
    has a minimal norm limit. Strict convexity gives uniqueness.

    For an acceptable objective
    \(J(f)=V(f(\mathbf x))+\mu\phi(\|f\|_{B_p^D})\), its minimizing
    sequences are bounded because \(V\ge0\) and \(\phi(r)\to+\infty\) as $r\to+\infty$.
    Along a weakly convergent subsequence, the sample vectors converge
    in \(\reals^n\), so continuity of \(V\) and continuity and
    monotonicity of \(\phi\) give lower semicontinuity of \(J\).
    Thus \(J\) attains its infimum. Finally, an invertible sample
    matrix supplies a sample-centered interpolant for every target.
\end{proof}

In particular, the full-space infimum in \eqref{relaxedbetan} is
attained in the present fixed-dictionary setting.

\subsection{Translation-invariant kernels on a separated dictionary}

\begin{theorem}\label{thm:spectral-p-greater-one}
    Let \(X=\mathbb R^d\), let \(D=\{\eta_j:j\ge1\}\) have a positive
    separation distance, and let \(1<p<\infty\), with \(q=p/(p-1)\).
    Suppose \(\varphi\in C^d(\mathbb R^d)\) satisfies
    \begin{equation}\label{eq:partials}
        \partial^\nu\varphi\in L^1(\mathbb R^d),
        \qquad \nu\in\{0,1\}^d,
    \end{equation}
    and define \(K\) by \eqref{eq:fourier_repr}. Then
    \[
        \sup_{\mathbf x\in\mathbb R^d}
        \sum_{j\ge1}|K(\eta_j,\mathbf x)|^q<\infty.
    \]
    In particular, (D1) holds. If \(\varphi\) is nowhere zero, then
    synthesis on \(\ell^p(D)\) is injective, which implies (D2).
\end{theorem}
\begin{proof}    Write $$
\widehat g(\mathbf z)=\int_{\reals^d} e^{-i \mathbf z\cdot\xi}g(\xi)\,d\xi.
$$
    If \(g\) and its mixed derivatives \(\partial^\nu g\),
    \(\nu\in\{0,1\}^d\), belong to \(L^1\), the Fourier derivative
    identity gives
    \[
        \widehat g(\mathbf z)
        =\left(\prod_{r\in E}\frac1{iz_r}\right)
        \widehat{\partial_E g}(\mathbf z),
        \qquad E=\{r:|z_r|>1\}.
    \]
    Here \(\partial_E=\prod_{r\in E}\partial_{\xi_r}\), with the
    empty product interpreted as the identity. Thus
    \begin{equation}\label{eq:kernel_anisotropic_decay}
        |\widehat g(\mathbf z)|\le M_g
        \prod_{r=1}^d(1+|z_r|)^{-1},\qquad
        M_g:=2^d\max_{\nu\in\{0,1\}^d}\|\partial^\nu g\|_{L^1}.
    \end{equation}
    These identities hold for integrable weak derivatives as well,
    so no boundary terms at infinity are required.

    Separation gives a uniform bound \(N_D\) on the number of centers
    in any translate of a unit cube. Apply this bound to the cubes
    \(\mathbf x+\mathbf m+[0,1)^d\), \(\mathbf m\in\mathbb Z^d\). For centers in such a
    cube, \((1+|\eta_{j,r}-x_r|)^{-1}\le2(1+|m_r|)^{-1}\).
    Since \(q>1\), it follows that
    \begin{equation}\label{eq:separated-product-sum}
        \sup_\mathbf x\sum_j\prod_{r=1}^d
        (1+|\eta_{j,r}-x_r|)^{-q}
        \le N_D2^{dq}
        \left(\sum_{l\in\mathbb Z}(1+|l|)^{-q}\right)^d<\infty.
    \end{equation}
    Taking \(g=\varphi\) proves the asserted uniform bound on the
    kernel values and hence (D1).

    Suppose now that \(\varphi\) is nowhere zero and
    \(F(\mathbf x):=\sum_jc_jK(\eta_j,\mathbf x)=0\), with \(c\in\ell^p(D)\).
    By the uniform bound just proved and H\"older's inequality, the
    partial sums \(F_N=\sum_{j=1}^N c_jK(\eta_j,\cdot)\) converge
    uniformly to \(F\). For the locally finite signed measure
    \(\mu=\sum_jc_j\delta_{\eta_j}\), separation gives
    \[
        |\mu|(B(0,R))\le C\|c\|_\infty(1+R)^d,\qquad R\ge0.
    \]
    This polynomial growth bound shows that \(\mu\) is a tempered
    distribution.
    Let \(T=\widehat\mu\). For smooth compactly supported \(g\),
    \[
        \langle T,g\rangle=\sum_jc_j\widehat g(\eta_j).
    \]
    Equations \eqref{eq:kernel_anisotropic_decay} and
    \eqref{eq:separated-product-sum} show more generally that this
    series defines a functional \(\widetilde T\) on \(C_c^d(\mathbb R^d)\),
    satisfying
    \begin{equation}\label{eq:distribution-cd-bound}
        |\widetilde T(g)|
        \le C_D\|c\|_p
        \max_{\nu\in\{0,1\}^d}\|\partial^\nu g\|_{L^1}.
    \end{equation}
    It is continuous in the \(C^d\) norm on functions with support
    in a fixed compact set, and agrees with \(T\) on smooth tests.

    Put \(T_N(\xi)=\sum_{j=1}^Nc_je^{-i\eta_j\cdot\xi}\), and
    regard it as a distribution through
    \[
        \langle T_N,g\rangle:=\int_{\mathbb R^d}T_N(\xi)g(\xi)\,d\xi,
        \qquad g\in C_c^d(\mathbb R^d).
    \]
    For \(\psi\in C_c^\infty(\mathbb R^d)\), a finite-sum Fourier
    calculation gives
    \[
        \langle\widehat F_N,\psi\rangle
        =(2\pi)^d\langle T_N,\varphi\psi\rangle.
    \]
    Uniform convergence of \(F_N\) permits passage to the limit on
    the left, since \(\widehat\psi\in L^1\). On the right,
    \(\varphi\psi\in C_c^d\), and \eqref{eq:distribution-cd-bound}
    applied to coefficient tails permits passage to \(\widetilde T\).
    As \(F=0\), we obtain \(\widetilde T(\varphi\psi)=0\).
    Approximation by smooth functions with support in a fixed compact
    set, together with \eqref{eq:distribution-cd-bound}, extends this to
    \begin{equation}\label{eq:phi_T_zero}
        \widetilde T(\varphi\psi)=0,
        \qquad \psi\in C_c^d(\mathbb R^d).
    \end{equation}
    Indeed, multiplication by \(\varphi\in C^d\) is continuous in
    this local \(C^d\) topology. For \(\chi\in C_c^\infty\), the
    nowhere-zero assumption gives \(\chi/\varphi\in C_c^d\).
    Substitution into \eqref{eq:phi_T_zero} yields
    \(\langle T,\chi\rangle=0\). Thus \(T=0\), and injectivity
    of the Fourier transform on tempered distributions gives \(\mu=0\).
    A smooth test function supported in a ball containing only
    \(\eta_j\), with value one there, then gives \(c_j=0\).
    This proves (D2).
\end{proof}

\subsection{Exact and relaxed representer theorems}

Fix a sampling tuple \(\mathbf x_n=(x_1,\ldots,x_n)\) in \(D\) with
nonsingular \(\mathrm A_n=\mathrm K[\mathbf x_n]\), and put
\(E=D\setminus\{x_1,\ldots,x_n\}\). Define the bounded operator
\[
    C_{\mathbf x_n}:\ell^p(E)\to\reals^n,\qquad
    C_{\mathbf x_n}b:=\mathrm A_n^{-1}\sum_{t\in E}b(t)K_{\mathbf x_n}(t).
\]
The series converges absolutely in each coordinate by (D1).
Under the isometry with
\(\ell^p(D)=\reals^n\oplus_p\ell^p(E)\), the interpolation
operator \(Q_{\mathbf x_n}\) has coefficient map
\[
    P_{\mathbf x_n}(\mathbf a,b)=\mathbf a+C_{\mathbf x_n}b.
\]
For operators between \(\ell^p\) spaces, \(\|\cdot\|_p\) denotes
the induced operator norm. For ordered tuples \(\mathbf a=(a_i)_{i=1}^r\) and
\(\mathbf b=(b_j)_{j=1}^s\) in \(D\), write
\(\mathrm K[\mathbf a,\mathbf b]=[K(b_j,a_i)]_{i,j}\).

As in Section~\ref{sec:relaxed-small-p}, the optimal factor is the
\(p\)-th power of the interpolation operator norm. In the present
Banach-space setting, the block map \(P_{\mathbf x_n}\) gives its
explicit value.

\begin{theorem}\label{Theorem: the upper and lower bounds of the Lebesgue constant}
    The smallest factor in the relaxed representer theorem at
    \(\mathbf x_n\) is
    \begin{equation}\label{eq:optimal-p-greater-factor}
        \beta_{\mathbf x_n}^*
        =\|P_{\mathbf x_n}\|_p^p
        =\bigl(1+\|C_{\mathbf x_n}\|_p^q\bigr)^{p/q}.
    \end{equation}
    Equivalently, the relaxed representer theorem holds with a factor
    \(\beta_n\ge1\) if and only if
    \begin{equation}\label{eq: coefficient inequality}
        \|Q_{\mathbf x_n}f\|_{B_p}^p
        \le\beta_n\|f\|_{B_p}^p,\qquad f\in B_p.
    \end{equation}
\end{theorem}
\begin{proof}
    First, \(\|Q_{\mathbf x_n}\|=\|P_{\mathbf x_n}\|_p\) by the
    coefficient isometry. The Banach adjoint of \(P_{\mathbf x_n}\)
    maps \(z\in\ell^q_n\) to \((z,C_{\mathbf x_n}'z)\). Hence
    \[
        \|P_{\mathbf x_n}\|_p
        =\sup_{\|z\|_q=1}
        \bigl(1+\|C_{\mathbf x_n}'z\|_q^q\bigr)^{1/q}
        =\bigl(1+\|C_{\mathbf x_n}\|_p^q\bigr)^{1/q},
    \]
    using equality of the norms of an operator and its Banach adjoint.

    The interpolation inequality
    \eqref{eq: coefficient inequality} implies the relaxed representer
    theorem by replacing each \(f\) with \(Q_{\mathbf x_n}f\).
    Conversely, suppose the relaxed representer theorem holds with
    factor \(\beta_n\). Fix \(f\in B_p\), put
    \(\mathbf y=f(\mathbf x)\), and minimize
    \[
        m\|\mathrm A_n \mathbf a-\mathbf y\|_2^2+\|\mathbf a\|_p^p
    \]
    over \(\mathbf a \in\reals^n\). This objective is continuous and
    coercive, so it has a minimizer \(\mathbf {a_m}\). The relaxed inequality
    and comparison with \(f\) give
    \[
        m\|\mathrm A_n \mathbf {a_m}-\mathbf y\|_2^2+\|\mathbf {a_m}\|_p^p
        \le\beta_n\|f\|_{B_p}^p.
    \]
    Thus \(\mathbf {a_m}\to \mathrm A_n^{-1}\mathbf y\); passing to the limit proves
    \eqref{eq: coefficient inequality}. It follows that a factor
    works precisely when \(\beta_n\ge\|Q_{\mathbf x_n}\|^p\), proving
    \eqref{eq:optimal-p-greater-factor}.
\end{proof}

\begin{corollary}\label{theorem: relaxed linear representer theorem iff}
    The relaxed representer theorem holds at \(\mathbf x_n\) with
    factor \(\beta_n\ge1\) if and only if, for every finite tuple
    \(\mathbf t\) of distinct points in \(E\),
    \begin{equation}\label{eq：p>1,upper bound of the Lebesgue constant}
        \|\mathrm A_n^{-1}\mathrm K[\mathbf x_n,\mathbf t]\|_p
        \le\bigl(\beta_n^{q/p}-1\bigr)^{1/q}.
    \end{equation}
\end{corollary}
\begin{proof}
    The norm of \(C_{\mathbf x_n}\) equals the supremum of the norms
    of its restrictions to finitely supported coefficient vectors:
    \[
        \|C_{\mathbf x_n}\|_p
        =\sup_{\mathbf t\subset E\text{ finite}}
        \|\mathrm A_n^{-1}\mathrm K[\mathbf x_n,\mathbf t]\|_p.
    \]
    This follows from density in \(\ell^p(E)\) and boundedness of
    \(C_{\mathbf x_n}\). The preceding theorem gives the claim.
\end{proof}

\begin{theorem}\label{nonexistenceforp>1}
    Assume (K3) on \(D\). If there are distinct \(u,v\in D\)
    with \(K(u,v)\ne0\), then \(B_p^D\) does not satisfy the exact
    linear representer theorem for all sampling tuples.
\end{theorem}
\begin{proof}
    Take the single sample \(v\), set
    \(b=K(u,v)/K(v,v)\ne0\), and prescribe the target \(y=K(u,v)\). The unique sample-span interpolant is
    \(f_0=bK(v,\cdot)\). For \(0\le t\le1\), the function
    \(g_t=(1-t)bK(v,\cdot)+tK(u,\cdot)\) has the same sample value.
    By (D2), its coefficient cost is
    \begin{equation}\label{eq:Norm inequality}
        \|g_t\|_{B_p}^p=(1-t)^p B+t^p,\qquad B:=|b|^p>0.
    \end{equation}
    Define
    \begin{equation}\label{eq: contradiction function}
        H(t):=(1-t)^pB+t^p-B.
    \end{equation}
    Its minimum on \([0,1]\) occurs at
    \begin{equation}\label{eq:t0_def}
        t_0=\frac{B^{1/(p-1)}}{1+B^{1/(p-1)}},
    \end{equation}
    and
    \[
        H(t_0)=\frac{B}{(1+B^{1/(p-1)})^{p-1}}-B<0.
    \]
    Hence \(f_0\) is not a minimal norm interpolant. Equivalently,
    \(Q_{(v)}\) strictly increases the norm of \(g_{t_0}\). Theorem~\ref{Theorem: the upper and lower bounds of the Lebesgue constant}, with \(\beta_n=1\), shows that the exact regularized
    representer property also fails for this sample.
\end{proof}

The optimal factor equals one exactly when \(C_{\mathbf x_n}=0\).
In this case \(Q_{\mathbf x_n}\) is norm-nonincreasing and preserves the
sample values. Applying it to a regularized minimizer also gives a
sample-span minimizer for every acceptable nondecreasing penalty.
Thus, under (K3), the exact property for every sampling tuple is
possible only if \(K(u,v)=0\) for all distinct dictionary points;
this condition is also sufficient.

\subsection{Entry-wise matrix bounds and a discrete-dictionary estimate}

For a real matrix \(\mathrm W\in\reals^{r\times s}\), define
\[
    \|\mathrm W\|_{ab}^E
    :=\left\|\bigl(\|\mathrm W_{\cdot j}\|_a\bigr)_{j=1}^s\right\|_b,
    \qquad 1\le a,b\le\infty.
\]
In particular,
\[
    \|\mathrm W\|_p\le\|\mathrm W\|_{pq}^E.
\]
Indeed, the triangle inequality followed by H\"older's inequality gives
\(\|\mathrm W \mathbf z\|_p\le\sum_j|z_j|\|\mathrm W_{\cdot j}\|_p
\le\|\mathbf z\|_p\|\mathrm W\|_{pq}^E\).

\begin{theorem}\label{thm:discrete-dictionary-uniform-factor}
    Fix \(1<p<\infty\) and let \(q=p/(p-1)\). Let \(X=\mathbb R\), and let the fixed dictionary have the form
    \(D=\{x_j:j\ge1\}\), where \(x_1<x_2<\cdots\) and
    \(x_{j+1}-x_j\ge h>0\). Assume (D1)--(D2), and put
    \(\mathbf x_n=(x_1,\ldots,x_n)\), \(\mathrm A_n=\mathrm K[\mathbf x_n]\).
    Suppose the finite matrices are uniformly stable:
    \[
        \mathrm A_n\text{ is nonsingular for every }n,\qquad
        \sup_n\|\mathrm A_n^{-1}\|\le L<\infty.
    \]
    If
    \[
        |K(u,v)|\le C_0(1+|u-v|)^{-\alpha},
        \qquad u,v\in\mathbb R,\qquad \alpha>1,
    \]
    then there exists a constant \(M>0\), depending only on \(p,\alpha,h,C_0,L\),
    such that, for every \(n\ge 1\) and every finite tuple \(\mathbf t\) of pairwise
    distinct points in \(D\setminus\{x_1,\ldots,x_n\}\),
    \[
        \|\mathrm A_n^{-1}\mathrm K[\mathbf x_n,\mathbf t]\|_{pq}^E\le M.
    \]
    Consequently, for every initial-segment sampling tuple \(\mathbf x_n\), the relaxed representer theorem holds with a factor
    \(\beta=(1+M^q)^{p/q}\) independent of \(n\).
\end{theorem}
\begin{proof}
    The spacing and decay bounds give
    \(|(\mathrm A_n)_{ij}|\le C_1(1+|i-j|)^{-\alpha}\).
    The uniform finite-matrix conclusion of
    Lemma~\ref{lemma:Jaffard's Theorem} yields
    \(|(\mathrm A_n^{-1})_{ij}|\le C_2(1+|i-j|)^{-\alpha}\), with constants
    independent of \(n\).

    Permuting columns preserves the mixed norm, so list the distinct
    off-sample centers as \(t_1<\cdots<t_m\). Writing
    \(t_k=x_{l_k}\), we have \(l_k\ge n+k\), and therefore
    \[
        |t_k-x_j|\ge h(n+k-j),\qquad 1\le j\le n.
    \]
    Put \(R=\mathrm A_n^{-1}\mathrm K[\mathbf x_n,\mathbf t]\).
    The algebraic convolution estimate in the proof of
    Theorem~\ref{Thm :Lebesgue constant satisfying algebraic decay conditions} gives
    \[
        |R_{ik}|\le C_3\sum_{j=1}^n
        (1+|i-j|)^{-\alpha}(1+n+k-j)^{-\alpha}
        \le C_4(1+n+k-i)^{-\alpha}.
    \]
    Since \(\alpha p>1\),
    \[
        \sum_{i=1}^n|R_{ik}|^p
        \le C_4^p\sum_{r=k}^{\infty}(1+r)^{-\alpha p}
        \le\frac{C_4^p}{\alpha p-1}k^{1-\alpha p}.
    \]
    Consequently,
    \[
        \bigl(\|R\|_{pq}^E\bigr)^q
        \le\frac{C_4^q}{(\alpha p-1)^{q/p}}
        \sum_{k=1}^{\infty}k^{-(\alpha p-1)q/p}<\infty,
    \]
    because \((\alpha p-1)q/p=(\alpha p-1)/(p-1)>1\).
    The bound is independent of \(n,m,\mathbf t\).
    Thus \(\|C_{\mathbf x_n}\|_p\le M\) by the finite-restriction
    identity in
    Corollary~\ref{theorem: relaxed linear representer theorem iff}, and
    \eqref{eq:optimal-p-greater-factor} gives the asserted factor.
\end{proof}

The initial-segment assumption ensures that every off-sample index
exceeds the size of the sampled block, which gives the distance bound
used in the proof.

\section{Numerical Experiments}
\subsection{Settings}
To validate the theoretical results established in this paper and further investigate the impact of different \(\ell^p\) regularization parameters on learning performance and sparsity, we conduct a series of numerical experiments.
Specifically, we compare the performance of the regularized learning models under various values of \(p\) with the classical Reproducing Kernel Hilbert Space (RKHS) models.
Our experiments primarily evaluate the function approximation accuracy and the sparse representation capability of each model under different noise conditions.

We employ three types of kernels: the Gaussian kernel\eqref{eq:Gaussian kernel}, the Laplacian kernel\eqref{eq:Laplacian Kernel}, and the inverse multiquadric (IMQ) kernel\eqref{eq:Inverse Multiquadric Kernel}.
We consider five different values for the parameter \(p\), namely \(p \in \{\frac{1}{2},\frac{2}{3}, 1, \frac{3}{2}, 2\}\).
Let \(B_p\) denote the \(p\)-normed space equipped with the \(\ell^p\) (quasi-)norm, and let \(\mathcal{H}_K\) be the RKHS associated with the kernel \(K\).
We formulate the regularized learning model in the space \(B_p\) as follows \[ \min_{f \in B_p} \left\{ \sum_{i=1}^n (f(x_i) - y_i)^2 + \mu \|f\|_{B_p}^p \right\}, \] where \(\mu > 0\) is the regularization parameter.
For comparison, the corresponding regularized model in the RKHS is formulated as \[ \min_{g \in \mathcal{H}_K} \left\{ \sum_{i=1}^n (g(x_i) - y_i)^2 + \mu \|g\|_{\mathcal{H}_K}^2 \right\}.
\]

By the representer theorems in previous sections, we search for candidate solutions within the finite-dimensional subspace \(S^{\mathbf{x}} := \mathrm{span}\{K(x_j,\cdot) : j \in \mathbb{N}_n\}\).
Consequently, the optimal solutions \(f_0\) and \(g_0\) of the two models can be represented as \[ f_0 = \mathrm{K}^{\mathbf{x}}(\cdot) \mathbf{a}, \quad g_0 = \mathrm{K}^{\mathbf{x}}(\cdot) \mathbf{b}, \] where the coefficient vectors \(\mathbf{a}\) and \(\mathbf{b}\) are given by \[ \mathbf{a} := \arg\min_{\mathbf{c} \in \reals^n} \left\{ \|\mathrm{K}[\bx]\mathbf{c} - \mathbf{y}\|_2^2 + \mu \|\mathbf{c}\|_{p}^p \right\}, \] \[ \mathbf{b} := \arg\min_{\mathbf{c} \in \reals^n} \left\{ \|\mathrm{K}[\bx]\mathbf{c} - \mathbf{y}\|_2^2 + \mu \mathbf{c}^T \mathrm{K}[\bx]\mathbf{c} \right\}.
\]

To solve the optimization problem for the coefficient vector \(\mathbf{a}\), we employ the proximal gradient method proposed in the literature.
For the coefficient vector \(\mathbf{b}\), the optimal solution has a well-known closed-form expression given by \(\mathbf{b} = (\mathrm{K}[\bx] + \mu \mathrm{I}_n)^{-1} \mathbf{y}\), where \(\mathrm{I}_n\) denotes the \(n \times n\) identity matrix.
Below we introduce the proximal operator and algorithm.

\subsection{Proximal gradient methods}

\begin{definition}\cite{Beck2009}
    Let \(\mu > 0\) be a regularization parameter.
    For any given vector \(\mathbf{v} \in \mathbb{R}^n\), the proxivmal mapping associated with the scaled function \(\mu \|\cdot\|_p^p\) is defined as \[ \operatorname{prox}_{\mu\|\cdot\|_p^p}(\mathbf{v}) := \arg\min_{\mathbf{u}\in \mathbb{R}^n} \left\{ \frac{1}{2}\|\mathbf{u} - \mathbf{v}\|_2^2 + \mu\|\mathbf{u}\|_p^p \right\}.
    \]
    The proximal gradient method to solve
    \[
        \min_{\mathbf{c}\in\bR^n}\|\mathrm{A}\mathbf{c}-\mathbf{y}\|_2^2+\mu\|\mathbf{c}\|_p^p
    \]
    starts with an initial \(\mathbf{c}^0\) and iterates by
    \[
        \mathbf{c}^{k+1}
        \in
        \operatorname{prox}_{\tau \mu \|\cdot\|_p^p}
        \left(
        \mathbf{c}^k-2\tau \mathrm{A}^T(\mathrm{A}\mathbf{c}^k-\mathbf{y})
        \right)
    \]
    with step size

    \[
        0<\tau<\frac{1}{2\|\mathrm{A}\|_2^2}.
    \]
\end{definition}

In our numerical experiments, we mainly employ the Fast Iterative Shrinkage-Thresholding Algorithm (FISTA), which incorporates an extrapolation step into the proximal gradient iteration described above.
Let \(L:=2\|\mathrm{A}\|_2^2\) be the Lipschitz constant of the gradient of the smooth loss term \(\|\mathrm{A}\mathbf{c}-\mathbf{y}\|_2^2\).
We choose the fixed step size \(\tau=1/(1.01L)\) and initialize \(\mathbf{c}^0=\mathbf{z}^0=0\) and \(\theta_0=1\).
At iteration \(k\), the coefficient vector is updated by \[ \mathbf{c}^{k+1}\in \operatorname{prox}_{\tau\mu\|\cdot\|_p^p} \left(\mathbf{z}^k-2\tau \mathrm{A}^T(\mathrm{A}\mathbf{z}^k-\mathbf{y})\right).
\]
If the stopping criterion is not met then the extrapolation parameters are updated as
\[
    \theta_{k+1}
    =\frac{1+\sqrt{1+4\theta_k^2}}{2},
    \qquad
    \mathbf{z}^{k+1}
    =\mathbf{c}^{k+1}
    +\frac{\theta_k-1}{\theta_{k+1}}
    \left(\mathbf{c}^{k+1}-\mathbf{c}^k\right).
\]
The experiments use a default stopping criterion
\[
    \frac{\|\mathbf{c}^{k+1}-\mathbf{c}^k\|_2}
    {\max\{\|\mathbf{c}^k\|_2,10^{-15}\}}
    <10^{-6},
\]
with a maximum of \(50000\) iterations. 

By the definition of the proximity operator, we have \[ \operatorname{prox}_{\mu\|\cdot\|_p^p}(\mathbf{v}) = (\operatorname{prox}_{\mu|\cdot|^p}(v_1), \operatorname{prox}_{\mu|\cdot|^p}(v_2),\cdots,\operatorname{prox}_{\mu|\cdot|^p}(v_n)),\ \ \mathbf{v} = (v_1,\cdots , v_n) \in \mathbb{R}^n.
\]

For those \(p\) used in our experiments, the proximal operator happens to have a closed-form.

\begin{lemma}\label{lemma: p=1/2,2/3,3/2,2 closed-form analytical solutions}
    Let \(\mu > 0\) and \(x \in \reals\).
    \begin{enumerate}
        \item[\rmnum{1})] For \(p = \frac{1}{2}\), the proximal operator \(\operatorname{prox}_{\mu|\cdot|^{\frac{1}{2}}}(x)\) has the closed-form analytical solution:
              \[
                  \operatorname{prox}_{\mu|\cdot|^{\frac{1}{2}}}(x) =
                  \begin{cases}
                      \{0\},                                                                                                                                                 & \text{if } |x| < \frac{3}{2}\mu^{\frac{2}{3}}; \\[2ex]
                      \{0, \operatorname{sign}(x) \cdot \mu^{\frac{2}{3}}\},                                                                                                 & \text{if } |x| = \frac{3}{2}\mu^{\frac{2}{3}}; \\[2ex]
                      \left\{ \frac{2}{3}x \left( 1 + \cos \left( \frac{2}{3}\arccos\left( -\frac{3^{\frac{3}{2}}}{4}\mu|x|^{-\frac{3}{2}} \right) \right) \right) \right\}, & \text{otherwise}.
                  \end{cases}
              \]
        \item[\rmnum{2})] For \(p = \frac{2}{3}\),
              \[
                  \operatorname{prox}_{\mu|\cdot|^{\frac{2}{3}}}(x) =
                  \begin{cases}
                      \{0\},                                                                                                                      & \text{if } |x| < 2 \left( \frac{2}{3}\mu \right)^{\frac{3}{4}}; \\[2ex]
                      \left\{ 0, \operatorname{sign}(x) \cdot \left( \frac{2}{3}\mu \right)^{\frac{3}{4}} \right\},                               & \text{if } |x| = 2 \left( \frac{2}{3}\mu \right)^{\frac{3}{4}}; \\[2ex]
                      \left\{\operatorname{sign}(x) \cdot \frac{1}{8} \left( \sqrt{2z} + \sqrt{ \frac{2|x|}{\sqrt{2z}} - 2z } \right)^3 \right\}, & \text{otherwise},
                  \end{cases}
              \]
              where the auxiliary variable \(z\) is defined as
              \[
                  z = \left( \frac{1}{16}x^2 + \sqrt{ \frac{x^4}{256} - \frac{8\mu^3}{729} } \right)^{1/3} + \left( \frac{1}{16}x^2 - \sqrt{ \frac{x^4}{256} - \frac{8\mu^3}{729} } \right)^{1/3}.
              \]
        \item[\rmnum{3})] For \(p=1\), \(\operatorname{prox}_{\mu|\cdot|}(x)\) becomes the soft-thresholding
              \[
                  \operatorname{prox}_{\mu|\cdot|}(x)=\operatorname{sign}(x)\max(|x|-\mu,0).
              \]

        \item[\rmnum{4})] For \(p = 2\),
              \[
                  \operatorname{prox}_{\mu|\cdot|^2}(x) = \frac{x}{2\mu + 1}.
              \]

        \item[\rmnum{5})] For \(p = \frac{3}{2}\),
              \begin{equation}\label{eq:p=3/2 prox}
                  \operatorname{prox}_{\mu|\cdot|^{\frac{3}{2} }}(x)=\operatorname{sign}(x)\left(\frac{-\frac{3}{2} \mu+\sqrt{\frac{9}{4} \mu^{2}+4|x|}}{2}\right)^{2} .
              \end{equation}
    \end{enumerate}
\end{lemma}

\begin{proof}
    Assertions (\romannumeral 1)-(\romannumeral 4) are well-known in the literature \cite{Chen2016,Parikh2014}, and (\romannumeral 5) follows from elementary calculations.
\end{proof}

\subsection{Results}
All experiments are conducted on the interval \([-1, 1]\).
We select an equispaced grid of \(200\) points on \([-1,1]\) as candidate kernel centers, given by \(x_i = -1+\frac{2(i-1)}{199}\) for \(i=1,\cdots,200\).
The true outputs are generated by the target function \(y_i = f(x_i)\).
To simulate the perturbations in real-world observations, we corrupt these clean outputs by adding random noises, including: Gaussian noise with variance \(0.0025\), uniform noise defined on the interval \([-0.1, 0.1]\), and random pepper sauce noise within the same interval.

For the Gaussian, Laplacian, and IMQ kernels, we set the parameters to \(\sigma = 0.1\), \(\gamma = 1\), and \(c = 0.3\), \(\beta = 1\), respectively.
To ensure that our findings are representative and not biased toward a specific class of functions, we choose four target functions with distinct structural characteristics as our test subjects.
These functions cover typical features such as local variations, global trends, periodic oscillations, and hybrid structures, thereby providing a comprehensive evaluation of the approximation accuracy, sparse representation capability, and algorithmic stability of different kernels and \(\ell^p\) regularization models.

The four target functions are explicitly defined as follows:
\begin{enumerate}
    \item[\rmnum{1})] \textbf{Gaussian Mixture with Oscillation (GMO)}:\(f_1(x) = e^{-8(x-0.4)^2} + 0.6 e^{-5(x+0.5)^2} + 0.3 \sin(4\pi x)\),
    \item[\rmnum{2})] \textbf{Hybrid Rational–Gaussian Function (HRG)}:\(f_2(x) = 0.45 e^{-6x^2} + \frac{0.25}{1 + 20(x+0.4)^2} + 0.15 \sin(2\pi x) + 0.10 x\),
    \item[\rmnum{3})] \textbf{Sigmoid–Polynomial Function (SPF)}:\(f_3(x) = \frac{0.8}{1 + e^{-4x}} + 0.15 x^3 - 0.4\),
    \item[\rmnum{4})] \textbf{Mixed Trigonometric–Hyperbolic Function (MTH)}:\(f_4(x) = 0.5 \sin(2\pi x) + 0.3 \cos(3\pi x) + 0.2 \tanh(2x)\).
\end{enumerate}

The regularization parameter \(\mu\) is chosen from the range \(\mu \in \{10^j : j = -7, -6, \dots, 1\}\), where the optimal value is selected to minimize the \(L^2([-1, 1])\) distance between the learned function and the target function.
For each parameter setting, the algorithms are initialized with the zero vector and iteratively solved.
The \(\ell^p\)-regularized models are computed using the FISTA method, whereas the RKHS is obtained by solving its associated linear system.
All iterations conclude when the relative difference between successive coefficient vectors falls below \(10^{-6}\), or upon reaching a maximum of \(50000\) iterations.

To comprehensively assess the performance of the models, we record the following evaluation metrics:
\begin{enumerate}
    \item[\rmnum{1})] \textbf{Approximation Error (\(\ell^p\) Model)}: The \(L^2\) approximation error between the learned function and the true target function;
    \item[\rmnum{2})] \textbf{RKHS Error}: The \(L^2\) approximation error of the corresponding RKHS model;
    \item[\rmnum{3})] \textbf{Sparsity}: The number of non-zero elements in the coefficient vector, which measures the sparsity of the learned model.
\end{enumerate}
To mitigate the effect of randomness, each experiment is independently repeated \(50\) times.
We report the average approximation errors, the average sparsity, and the maximum sparsity over the \(50\) runs as the final experimental data, which are summarized in the tables below.
The sparsity is also illustrated in Figures (a)-(l).
We conclude from the results that

\begin{enumerate}
    \item All models have comparable accuracy in terms of approximation errors.
    \item \(\ell^p\)-coefficient regularizations with \(p>1\) and the classical RKHS regularization method do not enjoy any sparsity.
    \item When \(p\le1\), sparse representation of the minimizers arises.
          Compared with \(p=1\), employing \(p<1\) such as \(p=\frac12\) and \(p=\frac23\) can usually enhance the sparsity dramatically.
          The solution becomes more sparse when \(p\) is closer to \(0\).
\end{enumerate}

\begin{table}[htbp]
    \centering
    \scalebox{0.9}{
        \begin{tabular}{lcccccc}
            \toprule
                                                     & \multicolumn{2}{l}{Gaussian noise} & \multicolumn{2}{l}{Uniform noise} & \multicolumn{2}{l}{Pepper sauce noise}                                             \\
            \cmidrule(r){2-3} \cmidrule(lr){4-5} \cmidrule(l){6-7}
                                                     & Error                              & Sparsity (Max)                    & Error                                  & Sparsity (Max) & Error   & Sparsity (Max) \\
            \midrule
            \(\ell^{\frac{1}{2}}\)-Regularized Model & 7.38E-4                            & 13 (16)                           & 8.18E-4                                & 13.08 (15)     & 9.00E-4 & 13.1 (16)      \\
            \(\ell^{\frac{2}{3}}\)-Regularized Model & 6.90E-4                            & 13.66 (16)                        & 7.77E-4                                & 13.66 (16)     & 8.63E-4 & 13.52 (16)     \\
            \(\ell^{1}\)-Regularized Model           & 6.32E-4                            & 23.04 (33)                        & 7.19E-4                                & 23.14(40)      & 8.20E-4 & 21.92(28)      \\
            \(\ell^{\frac{3}{2}}\)-Regularized Model & 6.21E-4                            & 199.98(200)                       & 7.11E-4                                & 200 (200)      & 8.23E-4 & 200 (200)      \\
            \(\ell^{2}\)-Regularized Model           & 6.52E-4                            & 200 (200)                         & 7.43E-4                                & 200 (200)      & 8.64E-4 & 200 (200)      \\
            RKHS                                     & 7.87E-4                            & 200 (200)                         & 8.87E-4                                & 200 (200)      & 1.05E-3 & 200 (200)      \\
            \bottomrule
        \end{tabular}
    }
    \caption{Comparison of Least-Squares Regularization in RKHS and with \(\ell^{p}\)-Regularized Models Using the \textbf{Gaussian}
        Kernel for the \textbf{GMO} function.
    }
\end{table}

\begin{table}[htbp]
    \centering
    \scalebox{0.9}{
        \begin{tabular}{lcccccc}
            \toprule
                                                     & \multicolumn{2}{l}{Gaussian noise} & \multicolumn{2}{l}{Uniform noise} & \multicolumn{2}{l}{Pepper sauce noise}                                              \\
            \cmidrule(r){2-3} \cmidrule(lr){4-5} \cmidrule(l){6-7}
                                                     & Error                              & Sparsity (Max)                    & Error                                  & Sparsity (Max) & Error    & Sparsity (Max) \\
            \midrule
            \(\ell^{\frac{1}{2}}\)-Regularized Model & 6.88E-04                           & 11.3 (14)                         & 7.52E-04                               & 11.5 (14)      & 8.27E-04 & 11.38 (14)     \\
            \(\ell^{\frac{2}{3}}\)-Regularized Model & 5.73E-04                           & 12.32 (15)                        & 6.37E-04                               & 12.04 (14)     & 7.12E-04 & 12.26 (15)     \\
            \(\ell^{1}\)-Regularized Model           & 5.51E-04                           & 20.42 (25)                        & 6.20E-04                               & 20.04 (24)     & 7.10E-04 & 20.18 (24)     \\
            \(\ell^{\frac{3}{2}}\)-Regularized Model & 5.86E-04                           & 200 (200)                         & 6.64E-04                               & 200 (200)      & 7.72E-04 & 200 (200)      \\
            \(\ell^{2}\)-Regularized Model           & 6.37E-04                           & 200 (200)                         & 7.20E-04                               & 200 (200)      & 8.39E-04 & 200 (200)      \\
            RKHS                                     & 7.83E-04                           & 200 (200)                         & 8.81E-04                               & 200 (200)      & 1.04E-03 & 200 (200)      \\
            \bottomrule
        \end{tabular}
    }
    \caption{Comparison of Least-Squares Regularization in RKHS and with \(\ell^{p}\)-Regularized Models Using the \textbf{Gaussian}
        Kernel for the \textbf{HRG} function.
    }
\end{table}

\begin{table}[htbp]
    \centering
    \scalebox{0.9}{
        \begin{tabular}{lcccccc}
            \toprule
                                                     & \multicolumn{2}{l}{Gaussian noise} & \multicolumn{2}{l}{Uniform noise} & \multicolumn{2}{l}{Pepper sauce noise}                                              \\
            \cmidrule(r){2-3} \cmidrule(lr){4-5} \cmidrule(l){6-7}
                                                     & Error                              & Sparsity (Max)                    & Error                                  & Sparsity (Max) & Error    & Sparsity (Max) \\
            \midrule
            \(\ell^{\frac{1}{2}}\)-Regularized Model & 6.71E-04                           & 11.84 (14)                        & 7.40E-04                               & 12.32 (14)     & 8.58E-04 & 11.86 (14)     \\
            \(\ell^{\frac{2}{3}}\)-Regularized Model & 6.35E-04                           & 12.36 (15)                        & 7.30E-04                               & 12.4 (16)      & 8.12E-04 & 12.44 (14)     \\
            \(\ell^{1}\)-Regularized Model           & 6.08E-04                           & 20.52 (25)                        & 6.97E-04                               & 20.4 (30)      & 8.06E-04 & 20.5 (24)      \\
            \(\ell^{\frac{3}{2}}\)-Regularized Model & 6.49E-04                           & 199.98 (200)                      & 7.46E-04                               & 200 (200)      & 8.60E-04 & 200 (200)      \\
            \(\ell^{2}\)-Regularized Model           & 6.84E-04                           & 200 (200)                         & 7.81E-04                               & 200 (200)      & 9.05E-04 & 200 (200)      \\
            RKHS                                     & 8.00E-04                           & 200 (200)                         & 8.99E-04                               & 200 (200)      & 1.06E-03 & 200 (200)      \\
            \bottomrule
        \end{tabular}
    }
    \caption{Comparison of Least-Squares Regularization in RKHS and with \(\ell^{p}\)-Regularized Models Using the \textbf{Gaussian}
        Kernel for the \textbf{SPF} function.
    }
\end{table}

\begin{table}[htbp]
    \centering
    \scalebox{0.9}{
        \begin{tabular}{lcccccc}
            \toprule
                                                     & \multicolumn{2}{l}{Gaussian noise} & \multicolumn{2}{l}{Uniform noise} & \multicolumn{2}{l}{Pepper sauce noise}                                              \\
            \cmidrule(r){2-3} \cmidrule(lr){4-5} \cmidrule(l){6-7}
                                                     & Error                              & Sparsity (Max)                    & Error                                  & Sparsity (Max) & Error    & Sparsity (Max) \\
            \midrule
            \(\ell^{\frac{1}{2}}\)-Regularized Model & 6.99E-04                           & 12.02 (15)                        & 7.33E-04                               & 12 (14)        & 8.52E-04 & 12.16 (14)     \\
            \(\ell^{\frac{2}{3}}\)-Regularized Model & 6.66E-04                           & 12.38 (15)                        & 7.24E-04                               & 12.44 (15)     & 8.50E-04 & 12.66 (15)     \\
            \(\ell^{1}\)-Regularized Model           & 6.38E-04                           & 21.46 (29)                        & 7.18E-04                               & 20.96 (26)     & 8.29E-04 & 20.9 (27)      \\
            \(\ell^{\frac{3}{2}}\)-Regularized Model & 6.59E-04                           & 200 (200)                         & 7.47E-04                               & 199.98 (200)   & 8.67E-04 & 199.98 (200)   \\
            \(\ell^{2}\)-Regularized Model           & 6.92E-04                           & 200 (200)                         & 7.80E-04                               & 200 (200)      & 9.09E-04 & 200 (200)      \\
            RKHS                                     & 8.04E-04                           & 200 (200)                         & 8.99E-04                               & 200 (200)      & 1.06E-03 & 200 (200)      \\
            \bottomrule
        \end{tabular}
    }
    \caption{Comparison of Least-Squares Regularization in RKHS and with \(\ell^{p}\)-Regularized Models Using the \textbf{Gaussian}
        Kernel for the \textbf{MTH} function.
    }
\end{table}

\begin{table}[htbp]
    \centering
    \scalebox{0.9}{
        \begin{tabular}{lcccccc}
            \toprule
                                                     & \multicolumn{2}{l}{Gaussian noise} & \multicolumn{2}{l}{Uniform noise} & \multicolumn{2}{l}{Pepper sauce noise}                                              \\
            \cmidrule(r){2-3} \cmidrule(lr){4-5} \cmidrule(l){6-7}
                                                     & Error                              & Sparsity (Max)                    & Error                                  & Sparsity (Max) & Error    & Sparsity (Max) \\
            \midrule
            \(\ell^{\frac{1}{2}}\)-Regularized Model & 1.38E-03                           & 18.9 (22)                         & 1.49E-03                               & 18.94 (23)     & 1.71E-03 & 18.94 (24)     \\
            \(\ell^{\frac{2}{3}}\)-Regularized Model & 1.32E-03                           & 19.98 (23)                        & 1.46E-03                               & 19.92 (22)     & 1.66E-03 & 19.62 (23)     \\
            \(\ell^{1}\)-Regularized Model           & 1.12E-03                           & 62.36 (93)                        & 1.25E-03                               & 58.16 (81)     & 1.43E-03 & 59.36 (78)     \\
            \(\ell^{\frac{3}{2}}\)-Regularized Model & 9.92E-04                           & 200 (200)                         & 1.11E-03                               & 200 (200)      & 1.26E-03 & 200 (200)      \\
            \(\ell^{2}\)-Regularized Model           & 9.92E-04                           & 200 (200)                         & 1.10E-03                               & 200 (200)      & 1.26E-03 & 200 (200)      \\
            RKHS                                     & 1.74E-03                           & 200 (200)                         & 2.12E-03                               & 200 (200)      & 2.54E-03 & 200 (200)      \\
            \bottomrule
        \end{tabular}
    }
    \caption{Comparison of Least-Squares Regularization in RKHS and with \(\ell^{p}\)-Regularized Models Using the \textbf{Laplacian}
        Kernel for the \textbf{GMO} function.
    }
\end{table}

\begin{table}[htbp]
    \centering
    \scalebox{0.9}{
        \begin{tabular}{lcccccc}
            \toprule
                                                     & \multicolumn{2}{l}{Gaussian noise} & \multicolumn{2}{l}{Uniform noise} & \multicolumn{2}{l}{Pepper sauce noise}                                              \\
            \cmidrule(r){2-3} \cmidrule(lr){4-5} \cmidrule(l){6-7}
                                                     & Error                              & Sparsity (Max)                    & Error                                  & Sparsity (Max) & Error    & Sparsity (Max) \\
            \midrule
            \(\ell^{\frac{1}{2}}\)-Regularized Model & 5.33E-04                           & 9.24 (11)                         & 5.78E-04                               & 9.64 (12)      & 6.61E-04 & 9.52 (12)      \\
            \(\ell^{\frac{2}{3}}\)-Regularized Model & 4.99E-04                           & 10.24 (12)                        & 5.57E-04                               & 10.36 (14)     & 6.45E-04 & 10.24 (13)     \\
            \(\ell^{1}\)-Regularized Model           & 4.75E-04                           & 25.78 (46)                        & 5.42E-04                               & 25.6 (37)      & 6.56E-04 & 24.9 (38)      \\
            \(\ell^{\frac{3}{2}}\)-Regularized Model & 5.62E-04                           & 200 (200)                         & 6.40E-04                               & 200 (200)      & 7.66E-04 & 200 (200)      \\
            \(\ell^{2}\)-Regularized Model           & 6.74E-04                           & 200 (200)                         & 7.68E-04                               & 200 (200)      & 9.05E-04 & 200 (200)      \\
            RKHS                                     & 1.61E-03                           & 200 (200)                         & 1.98E-03                               & 200 (200)      & 2.39E-03 & 200 (200)      \\
            \bottomrule
        \end{tabular}
    }
    \caption{Comparison of Least-Squares Regularization in RKHS and with \(\ell^{p}\)-Regularized Models Using the \textbf{Laplacian}
        Kernel for the \textbf{HRG} function.
    }
\end{table}

\begin{table}[htbp]
    \centering
    \scalebox{0.9}{
        \begin{tabular}{lcccccc}
            \toprule
                                                     & \multicolumn{2}{l}{Gaussian noise} & \multicolumn{2}{l}{Uniform noise} & \multicolumn{2}{l}{Pepper sauce noise}                                              \\
            \cmidrule(r){2-3} \cmidrule(lr){4-5} \cmidrule(l){6-7}
                                                     & Error                              & Sparsity (Max)                    & Error                                  & Sparsity (Max) & Error    & Sparsity (Max) \\
            \midrule
            \(\ell^{\frac{1}{2}}\)-Regularized Model & 3.43E-04                           & 4.02 (5)                          & 4.32E-04                               & 4.04 (5)       & 4.78E-04 & 4.04 (5)       \\
            \(\ell^{\frac{2}{3}}\)-Regularized Model & 3.43E-04                           & 4.24 (6)                          & 4.08E-04                               & 4.14 (5)       & 4.82E-04 & 4.18 (6)       \\
            \(\ell^{1}\)-Regularized Model           & 2.69E-04                           & 11.42 (19)                        & 3.05E-04                               & 11.16 (20)     & 3.61E-04 & 11.02 (17)     \\
            \(\ell^{\frac{3}{2}}\)-Regularized Model & 3.30E-04                           & 200 (200)                         & 3.63E-04                               & 200 (200)      & 4.33E-04 & 200 (200)      \\
            \(\ell^{2}\)-Regularized Model           & 4.15E-04                           & 200 (200)                         & 4.67E-04                               & 200 (200)      & 5.48E-04 & 200 (200)      \\
            RKHS                                     & 6.32E-04                           & 200 (200)                         & 7.32E-04                               & 200 (200)      & 8.68E-04 & 200 (200)      \\
            \bottomrule
        \end{tabular}
    }
    \caption{Comparison of Least-Squares Regularization in RKHS and with \(\ell^{p}\)-Regularized Models Using the \textbf{Laplacian}
        Kernel for the \textbf{SPF} function.
    }
\end{table}

\begin{table}[htbp]
    \centering
    \scalebox{0.9}{
        \begin{tabular}{lcccccc}
            \toprule
                                                     & \multicolumn{2}{l}{Gaussian noise} & \multicolumn{2}{l}{Uniform noise} & \multicolumn{2}{l}{Pepper sauce noise}                                              \\
            \cmidrule(r){2-3} \cmidrule(lr){4-5} \cmidrule(l){6-7}
                                                     & Error                              & Sparsity (Max)                    & Error                                  & Sparsity (Max) & Error    & Sparsity (Max) \\
            \midrule
            \(\ell^{\frac{1}{2}}\)-Regularized Model & 1.09E-03                           & 15.36 (18)                        & 1.20E-03                               & 15.54 (21)     & 1.37E-03 & 15.5 (19)      \\
            \(\ell^{\frac{2}{3}}\)-Regularized Model & 1.04E-03                           & 16.16 (19)                        & 1.14E-03                               & 16.42 (21)     & 1.28E-03 & 16.32 (20)     \\
            \(\ell^{1}\)-Regularized Model           & 8.69E-04                           & 50.7 (85)                         & 9.63E-04                               & 48.36 (71)     & 1.11E-03 & 47.6 (75)      \\
            \(\ell^{\frac{3}{2}}\)-Regularized Model & 8.40E-04                           & 200 (200)                         & 9.15E-04                               & 200 (200)      & 1.06E-03 & 200 (200)      \\
            \(\ell^{2}\)-Regularized Model           & 9.15E-04                           & 200 (200)                         & 9.98E-04                               & 200 (200)      & 1.15E-03 & 200 (200)      \\
            RKHS                                     & 1.72E-03                           & 200 (200)                         & 2.09E-03                               & 200 (200)      & 2.51E-03 & 200 (200)      \\
            \bottomrule
        \end{tabular}
    }
    \caption{Comparison of Least-Squares Regularization in RKHS and with \(\ell^{p}\)-Regularized Models Using the \textbf{Laplacian}
        Kernel for the \textbf{MTH} function.
    }
\end{table}

\begin{table}[htbp]
    \centering
    \scalebox{0.9}{
        \begin{tabular}{lcccccc}
            \toprule
                                                     & \multicolumn{2}{l}{Gaussian noise} & \multicolumn{2}{l}{Uniform noise} & \multicolumn{2}{l}{Pepper sauce noise}                                              \\
            \cmidrule(r){2-3} \cmidrule(lr){4-5} \cmidrule(l){6-7}
                                                     & Error                              & Sparsity (Max)                    & Error                                  & Sparsity (Max) & Error    & Sparsity (Max) \\
            \midrule
            \(\ell^{\frac{1}{2}}\)-Regularized Model & 5.32E-04                           & 13.04 (17)                        & 5.88E-04                               & 13.14 (16)     & 6.94E-04 & 13.24 (17)     \\
            \(\ell^{\frac{2}{3}}\)-Regularized Model & 5.15E-04                           & 18.62 (25)                        & 5.82E-04                               & 18.94 (23)     & 6.76E-04 & 18.46 (23)     \\
            \(\ell^{1}\)-Regularized Model           & 5.16E-04                           & 129.48 (156)                      & 5.84E-04                               & 126.34 (150)   & 6.81E-04 & 125.78 (149)   \\
            \(\ell^{\frac{3}{2}}\)-Regularized Model & 5.57E-04                           & 200 (200)                         & 6.34E-04                               & 200 (200)      & 7.37E-04 & 200 (200)      \\
            \(\ell^{2}\)-Regularized Model           & 6.00E-04                           & 200 (200)                         & 6.87E-04                               & 200 (200)      & 7.98E-04 & 200 (200)      \\
            RKHS                                     & 7.57E-04                           & 200 (200)                         & 8.54E-04                               & 200 (200)      & 1.00E-03 & 200 (200)      \\
            \bottomrule
        \end{tabular}
    }
    \caption{Comparison of Least-Squares Regularization in RKHS and with \(\ell^{p}\)-Regularized Models Using the \textbf{IMQ}
        Kernel for the \textbf{GMO} function.
    }
\end{table}

\begin{table}[htbp]
    \centering
    \scalebox{0.9}{
        \begin{tabular}{lcccccc}
            \toprule
                                                     & \multicolumn{2}{l}{Gaussian noise} & \multicolumn{2}{l}{Uniform noise} & \multicolumn{2}{l}{Pepper sauce noise}                                              \\
            \cmidrule(r){2-3} \cmidrule(lr){4-5} \cmidrule(l){6-7}
                                                     & Error                              & Sparsity (Max)                    & Error                                  & Sparsity (Max) & Error    & Sparsity (Max) \\
            \midrule
            \(\ell^{\frac{1}{2}}\)-Regularized Model & 3.80E-04                           & 9.22 (12)                         & 4.19E-04                               & 9.36 (11)      & 4.95E-04 & 9.54 (12)      \\
            \(\ell^{\frac{2}{3}}\)-Regularized Model & 3.90E-04                           & 10.28 (12)                        & 4.40E-04                               & 10.46 (13)     & 5.21E-04 & 10.62 (13)     \\
            \(\ell^{1}\)-Regularized Model           & 4.53E-04                           & 66.54 (109)                       & 5.02E-04                               & 70.14 (118)    & 5.97E-04 & 75.22 (112)    \\
            \(\ell^{\frac{3}{2}}\)-Regularized Model & 5.34E-04                           & 200 (200)                         & 6.00E-04                               & 200 (200)      & 7.00E-04 & 200 (200)      \\
            \(\ell^{2}\)-Regularized Model           & 5.89E-04                           & 200 (200)                         & 6.68E-04                               & 200 (200)      & 7.77E-04 & 200 (200)      \\
            RKHS                                     & 7.52E-04                           & 200 (200)                         & 8.48E-04                               & 200 (200)      & 1.00E-03 & 200 (200)      \\
            \bottomrule
        \end{tabular}
    }
    \caption{Comparison of Least-Squares Regularization in RKHS and with \(\ell^{p}\)-Regularized Models Using the \textbf{IMQ}
        Kernel for the \textbf{HRG} function.
    }
\end{table}

\begin{table}[htbp]
    \centering
    \scalebox{0.9}{
        \begin{tabular}{lcccccc}
            \toprule
                                                     & \multicolumn{2}{l}{Gaussian noise} & \multicolumn{2}{l}{Uniform noise} & \multicolumn{2}{l}{Pepper sauce noise}                                              \\
            \cmidrule(r){2-3} \cmidrule(lr){4-5} \cmidrule(l){6-7}
                                                     & Error                              & Sparsity (Max)                    & Error                                  & Sparsity (Max) & Error    & Sparsity (Max) \\
            \midrule
            \(\ell^{\frac{1}{2}}\)-Regularized Model & 4.01E-04                           & 6.34 (8)                          & 4.19E-04                               & 6.22 (8)       & 5.15E-04 & 6.16 (8)       \\
            \(\ell^{\frac{2}{3}}\)-Regularized Model & 3.83E-04                           & 6.22 (8)                          & 3.94E-04                               & 6.26 (7)       & 4.65E-04 & 6.22 (8)       \\
            \(\ell^{1}\)-Regularized Model           & 3.50E-04                           & 14.02 (23)                        & 3.85E-04                               & 14.38 (22)     & 4.59E-04 & 14.92 (29)     \\
            \(\ell^{\frac{3}{2}}\)-Regularized Model & 4.27E-04                           & 200 (200)                         & 4.82E-04                               & 200 (200)      & 5.71E-04 & 200 (200)      \\
            \(\ell^{2}\)-Regularized Model           & 5.03E-04                           & 200 (200)                         & 5.62E-04                               & 200 (200)      & 6.56E-04 & 200 (200)      \\
            RKHS                                     & 5.54E-04                           & 200 (200)                         & 6.26E-04                               & 200 (200)      & 7.31E-04 & 200 (200)      \\
            \bottomrule
        \end{tabular}
    }
    \caption{Comparison of Least-Squares Regularization in RKHS and with \(\ell^{p}\)-Regularized Models Using the \textbf{IMQ}
        Kernel for the \textbf{SPF} function.
    }
\end{table}

\begin{table}[htbp]
    \centering
    \scalebox{0.9}{
        \begin{tabular}{lcccccc}
            \toprule
                                                     & \multicolumn{2}{l}{Gaussian noise} & \multicolumn{2}{l}{Uniform noise} & \multicolumn{2}{l}{Pepper sauce noise}                                              \\
            \cmidrule(r){2-3} \cmidrule(lr){4-5} \cmidrule(l){6-7}
                                                     & Error                              & Sparsity (Max)                    & Error                                  & Sparsity (Max) & Error    & Sparsity (Max) \\
            \midrule
            \(\ell^{\frac{1}{2}}\)-Regularized Model & 4.93E-04                           & 11.04 (14)                        & 5.58E-04                               & 10.82 (13)     & 6.53E-04 & 10.7 (13)      \\
            \(\ell^{\frac{2}{3}}\)-Regularized Model & 4.86E-04                           & 12.92 (15)                        & 5.57E-04                               & 12.46 (15)     & 6.47E-04 & 12.42 (14)     \\
            \(\ell^{1}\)-Regularized Model           & 4.91E-04                           & 104.22 (140)                      & 5.57E-04                               & 102.1 (131)    & 6.51E-04 & 101.7 (131)    \\
            \(\ell^{\frac{3}{2}}\)-Regularized Model & 5.49E-04                           & 200 (200)                         & 6.24E-04                               & 200 (200)      & 7.25E-04 & 200 (200)      \\
            \(\ell^{2}\)-Regularized Model           & 5.96E-04                           & 200 (200)                         & 6.83E-04                               & 200 (200)      & 7.93E-04 & 200 (200)      \\
            RKHS                                     & 7.56E-04                           & 200 (200)                         & 8.53E-04                               & 200 (200)      & 1.00E-03 & 200 (200)      \\
            \bottomrule
        \end{tabular}
    }
    \caption{Comparison of Least-Squares Regularization in RKHS and with \(\ell^{p}\)-Regularized Models Using the \textbf{IMQ}
        Kernel for the \textbf{MTH} function.
    }
\end{table}

\begin{figure}[htbp]
    \centering
    \begin{subfigure}[b]{0.48\textwidth}
        \centering
        \includegraphics[width=\textwidth]{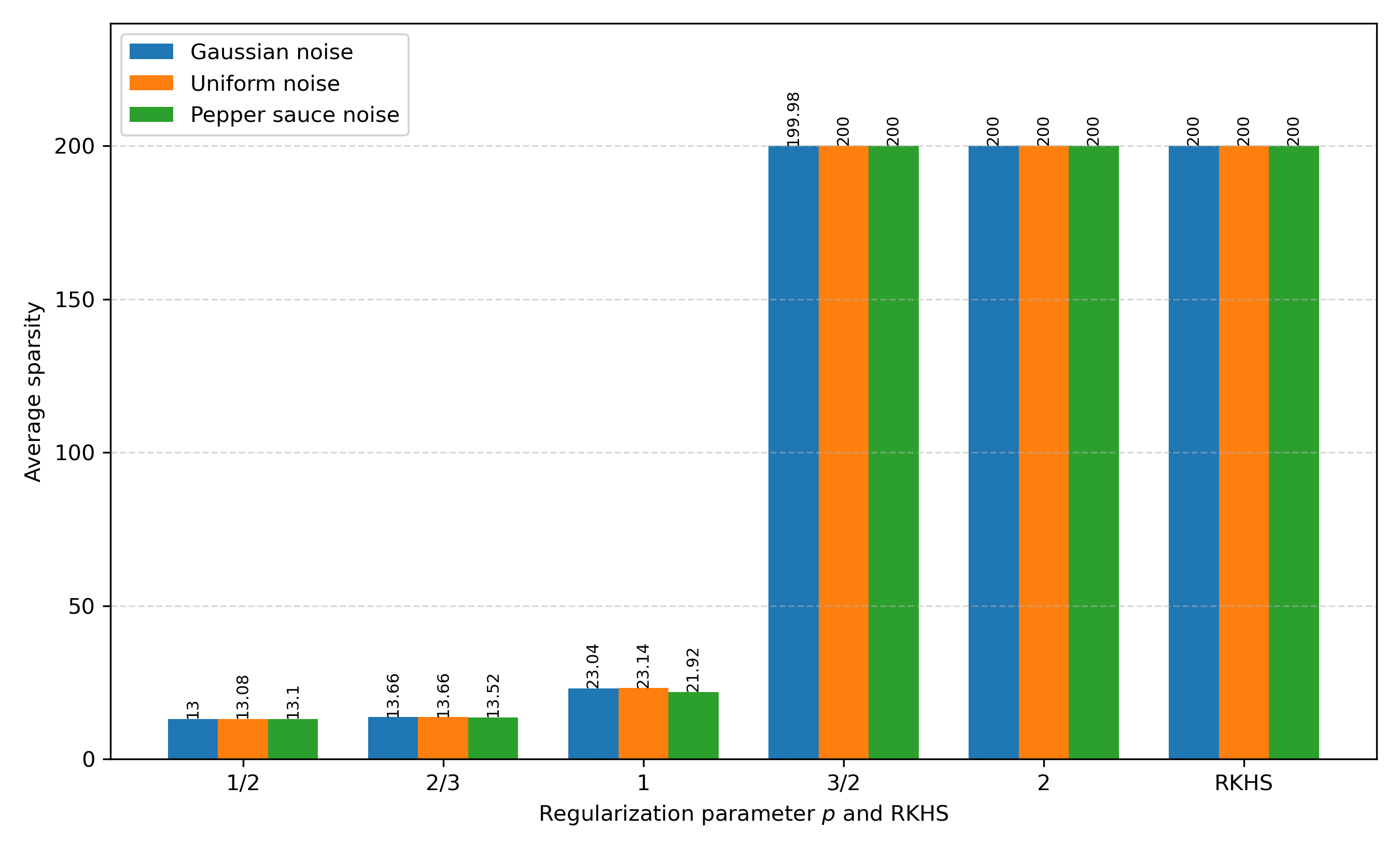}
        \caption{Average Sparsity of RKHS and \(\ell^p\)-Regularized Models under Different Noise Types Using the Gaussian Kernel for the GMO function.}
        \label{fig:1-1}
    \end{subfigure}
    \hfill
    \begin{subfigure}[b]{0.48\textwidth}
        \centering
        \includegraphics[width=\textwidth]{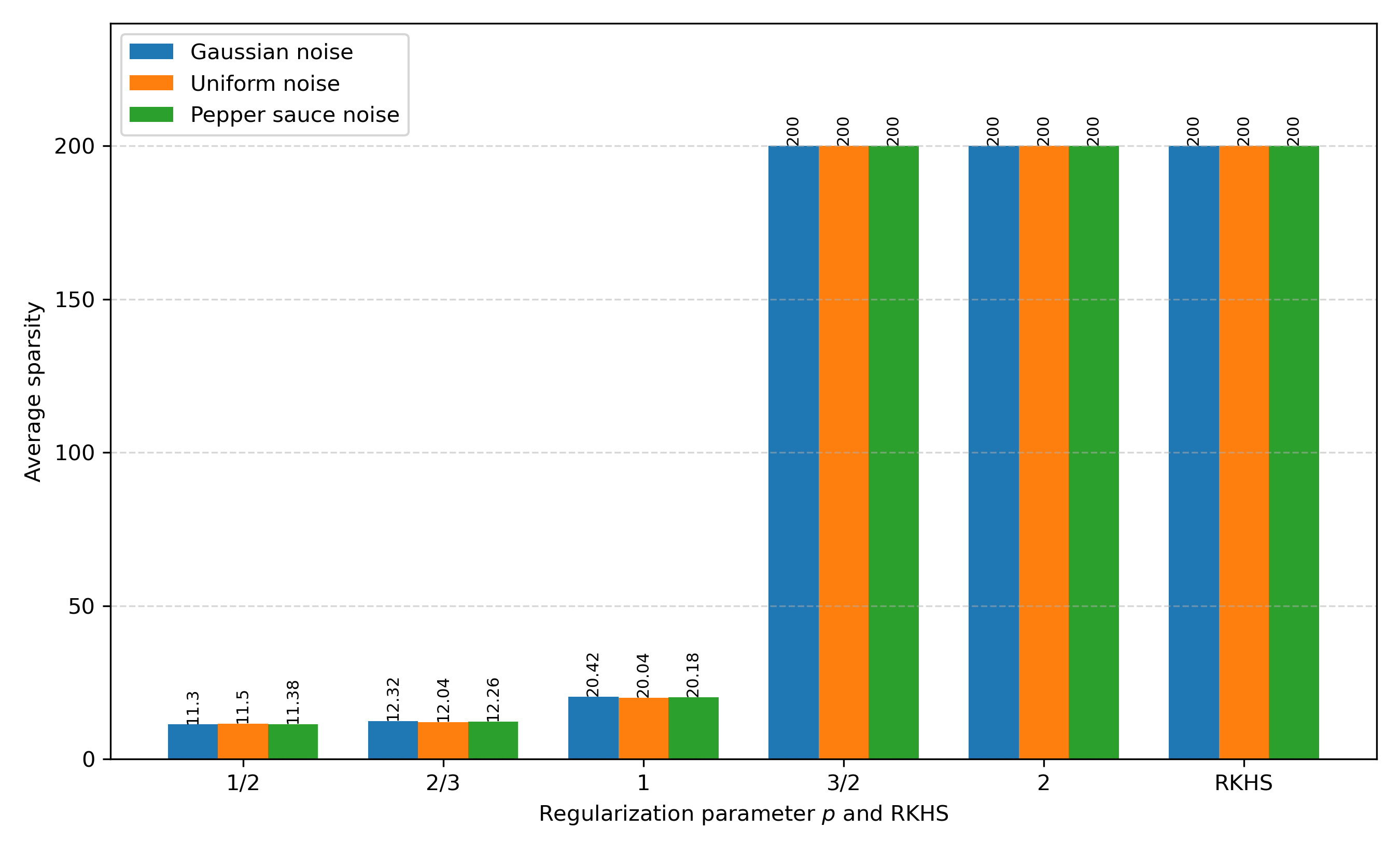}
        \caption{Average Sparsity of RKHS and \(\ell^p\)-Regularized Models under Different Noise Types Using the Gaussian Kernel for the HRG function.}
        \label{fig:1-2}
    \end{subfigure}

\end{figure}

\begin{figure}[htbp]
    \ContinuedFloat
    \centering
    \begin{subfigure}[b]{0.48\textwidth}
        \centering
        \includegraphics[width=\textwidth]{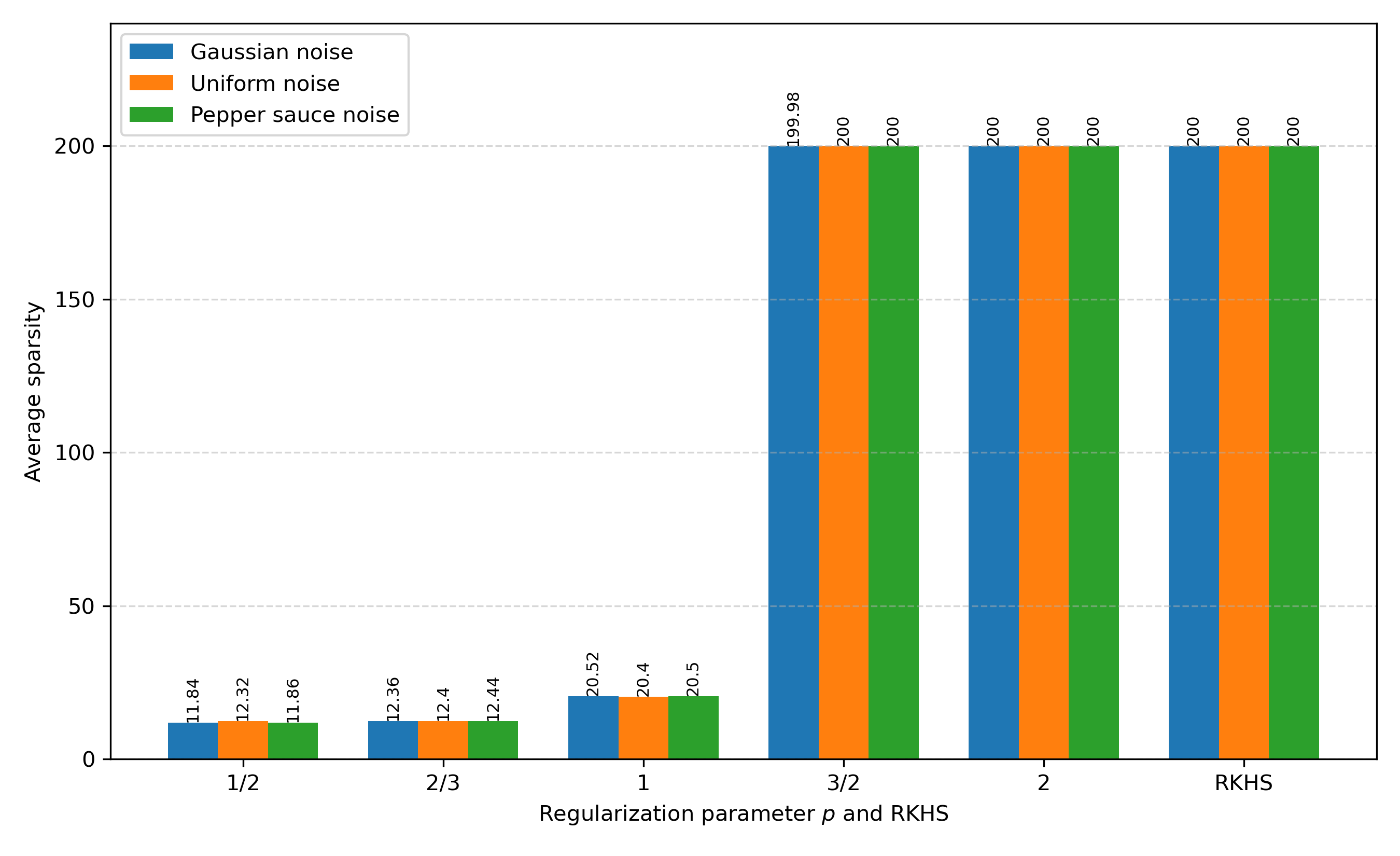}
        \caption{Average Sparsity of RKHS and \(\ell^p\)-Regularized Models under Different Noise Types Using the Gaussian Kernel for the SPF function.}
        \label{fig:1-3}
    \end{subfigure}
    \hfill
    \begin{subfigure}[b]{0.48\textwidth}
        \centering
        \includegraphics[width=\textwidth]{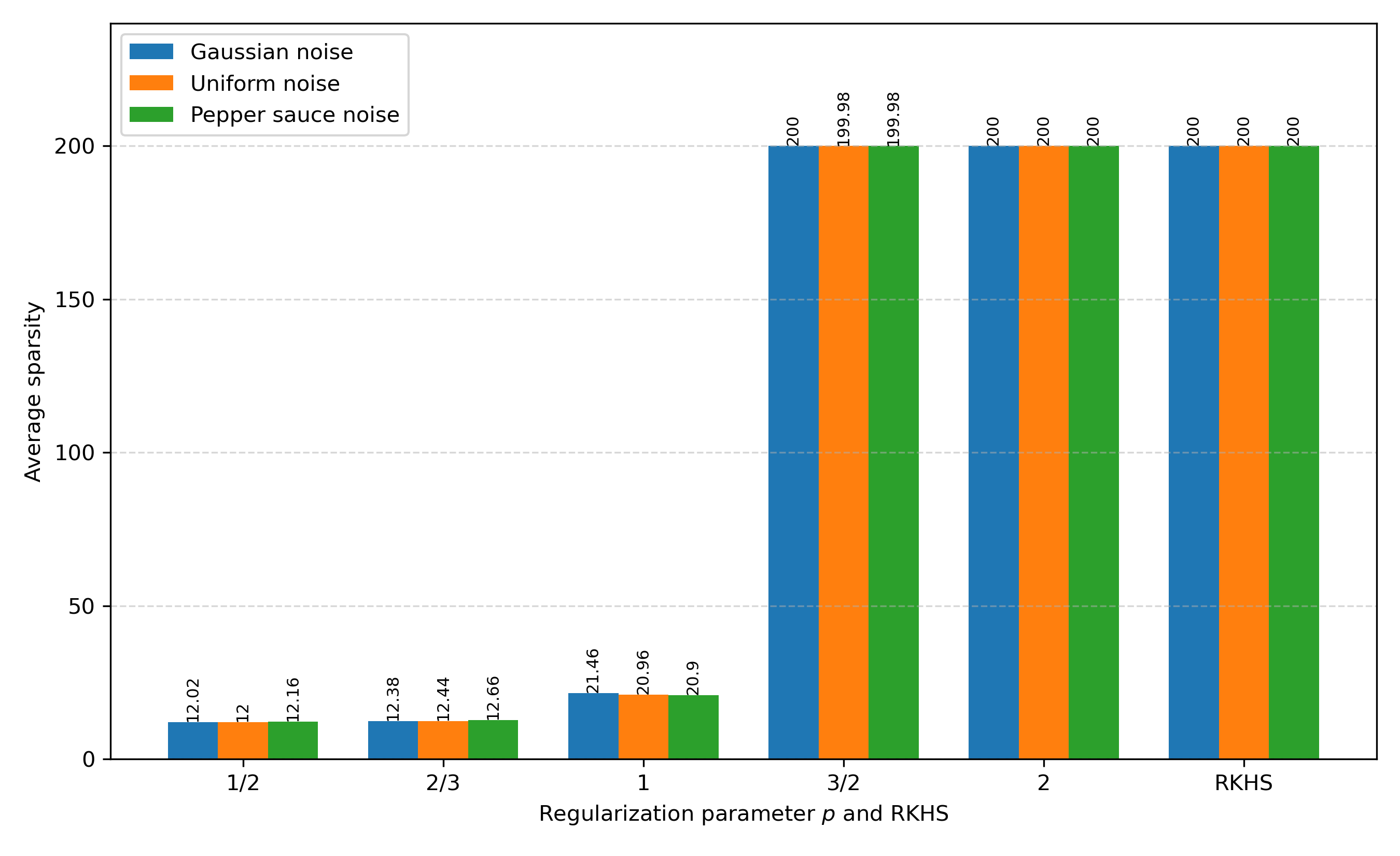}
        \caption{Average Sparsity of RKHS and \(\ell^p\)-Regularized Models under Different Noise Types Using the Gaussian Kernel for the MTH function.}
        \label{fig:1-4}
    \end{subfigure}
\end{figure}

\begin{figure}[htbp]
    \ContinuedFloat
    \centering
    \begin{subfigure}[b]{0.48\textwidth}
        \centering
        \includegraphics[width=\textwidth]{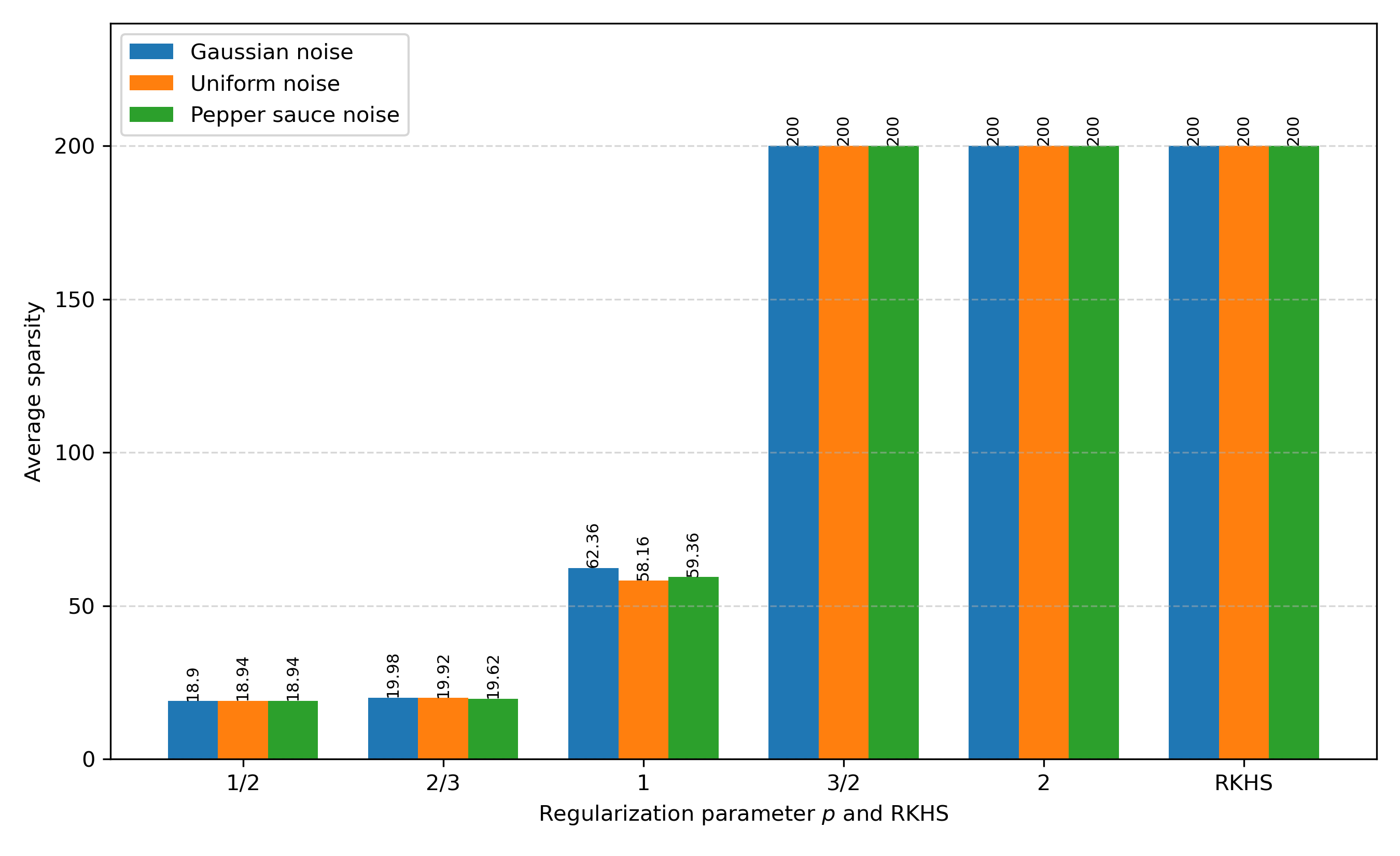}
        \caption{Average Sparsity of RKHS and \(\ell^p\)-Regularized Models under Different Noise Types Using the Laplacian Kernel for the GMO function.}
        \label{fig:2-1}
    \end{subfigure}
    \hfill
    \begin{subfigure}[b]{0.48\textwidth}
        \centering
        \includegraphics[width=\textwidth]{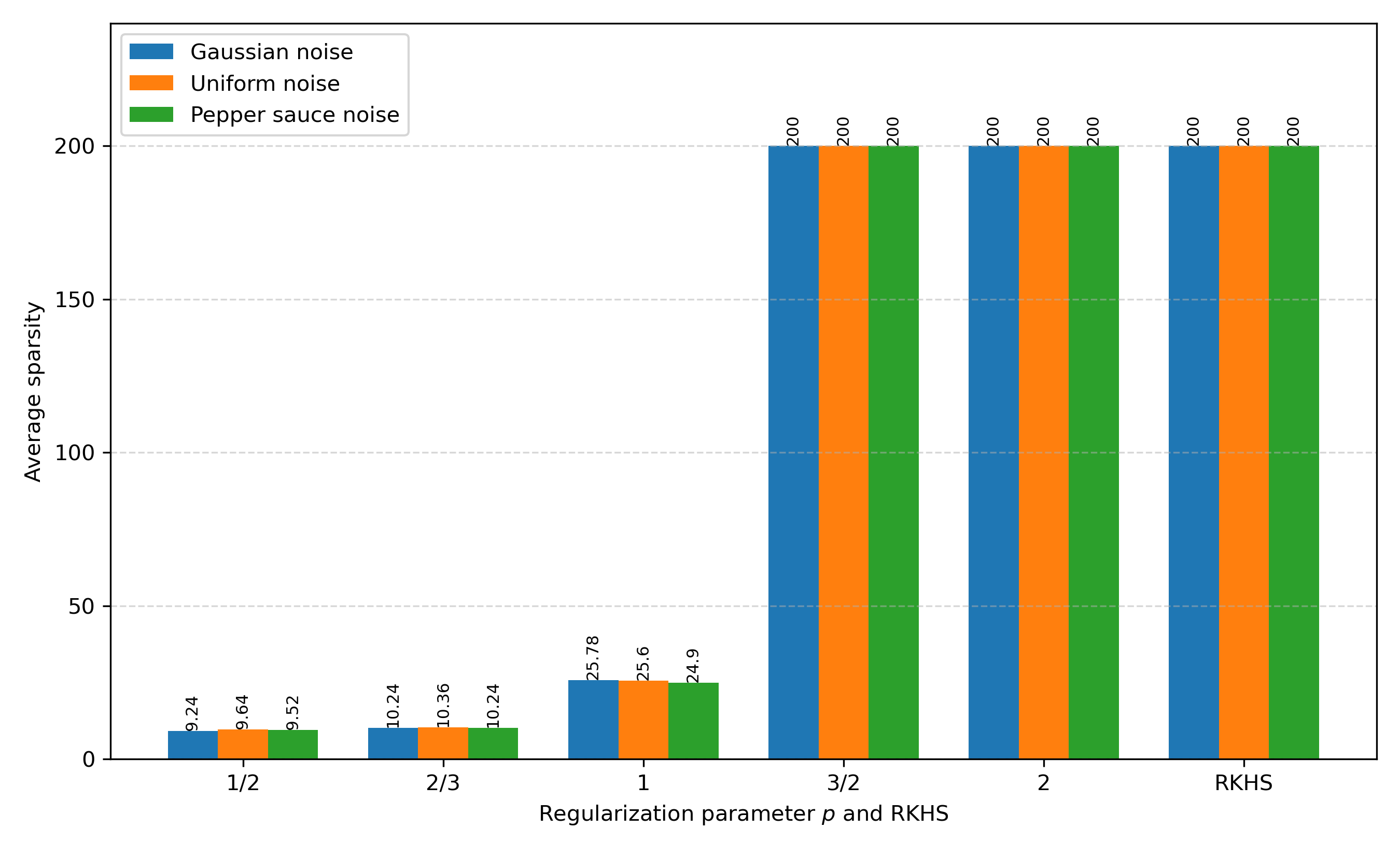}
        \caption{Average Sparsity of RKHS and \(\ell^p\)-Regularized Models under Different Noise Types Using the Laplacian Kernel for the HRG function.}
        \label{fig:2-2}
    \end{subfigure}
\end{figure}

\begin{figure}[htbp]
    \ContinuedFloat
    \centering
    \begin{subfigure}[b]{0.48\textwidth}
        \centering
        \includegraphics[width=\textwidth]{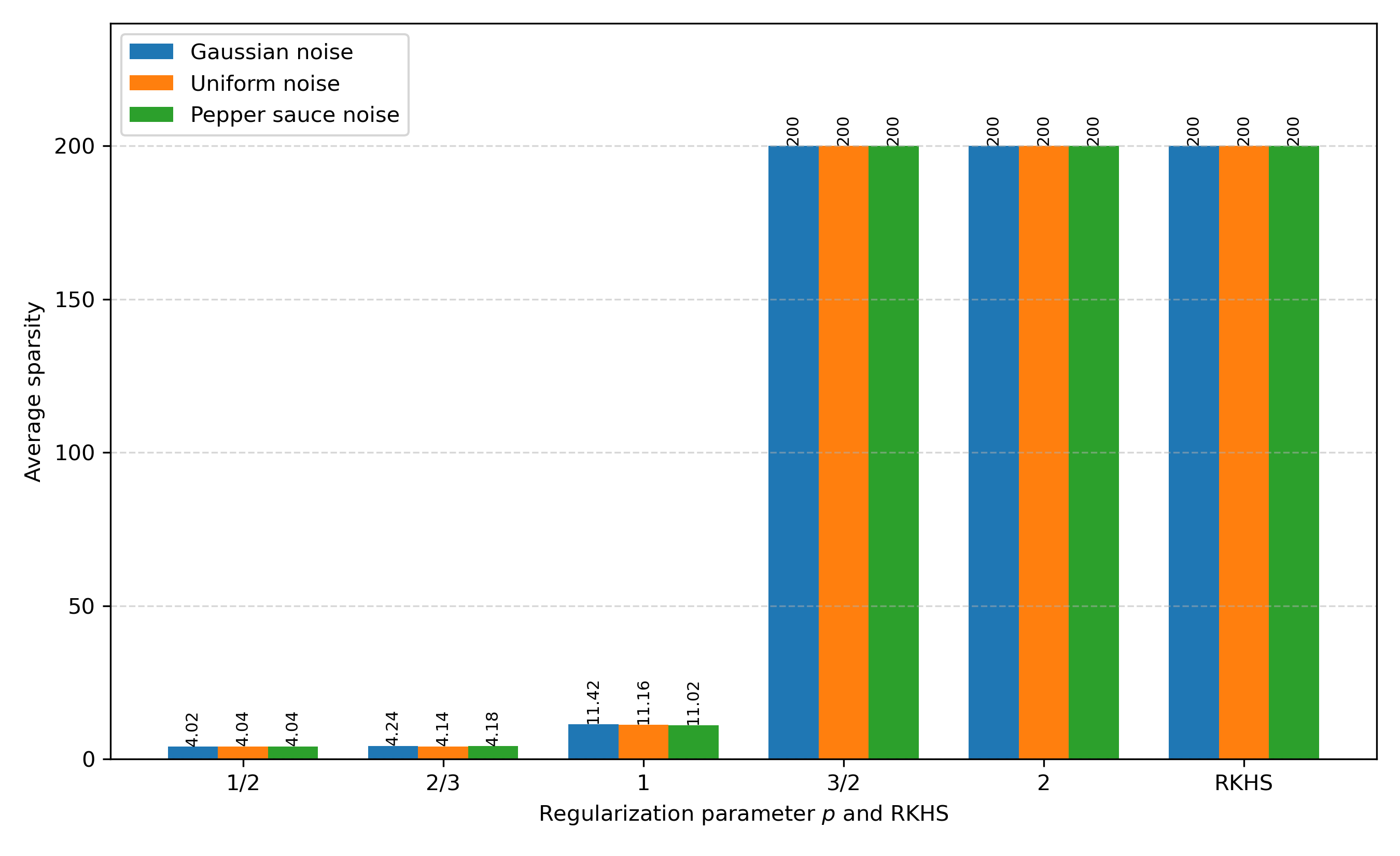}
        \caption{Average Sparsity of RKHS and \(\ell^p\)-Regularized Models under Different Noise Types Using the Laplacian Kernel for the SPF function.}
        \label{fig:2-3}
    \end{subfigure}
    \hfill
    \begin{subfigure}[b]{0.48\textwidth}
        \centering
        \includegraphics[width=\textwidth]{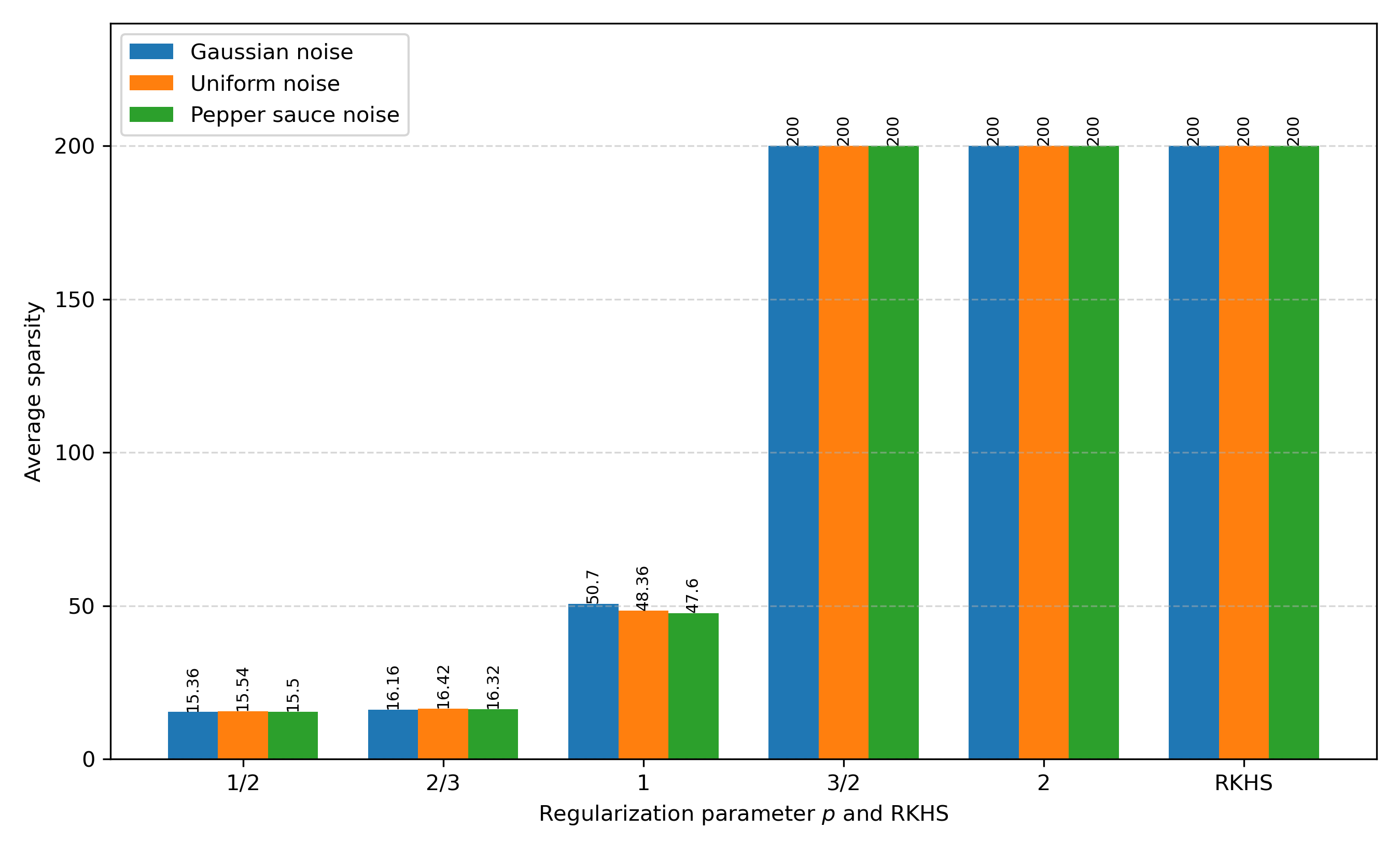}
        \caption{Average Sparsity of RKHS and \(\ell^p\)-Regularized Models under Different Noise Types Using the Laplacian Kernel for the MTH function.}
        \label{fig:2-4}
    \end{subfigure}
\end{figure}

\begin{figure}[htbp]
    \ContinuedFloat
    \centering
    \begin{subfigure}[b]{0.48\textwidth}
        \centering
        \includegraphics[width=\textwidth]{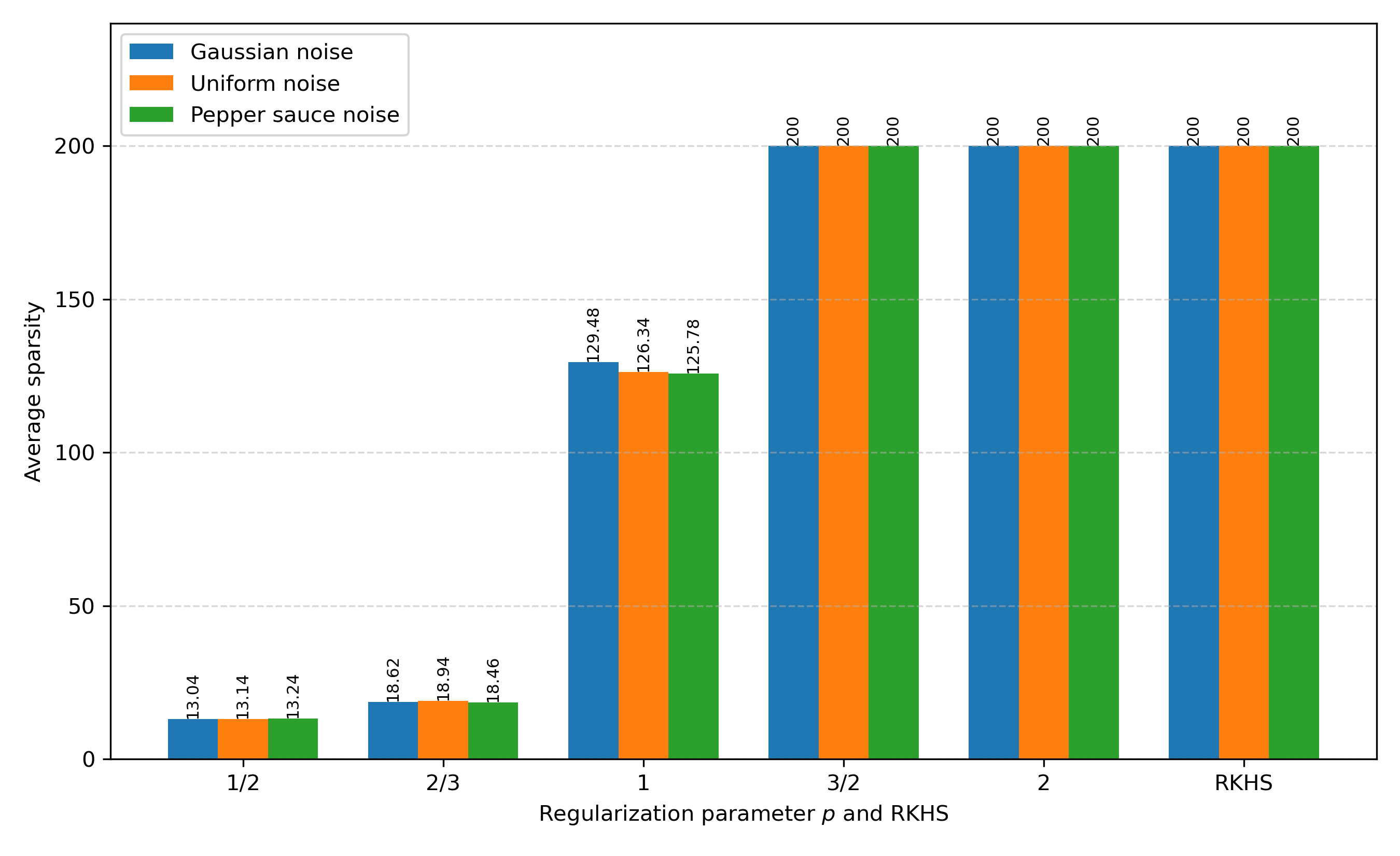}
        \caption{Average Sparsity of RKHS and \(\ell^p\)-Regularized Models under Different Noise Types Using the IMQ Kernel for the GMO function.}
        \label{fig:3-1}
    \end{subfigure}
    \hfill
    \begin{subfigure}[b]{0.48\textwidth}
        \centering
        \includegraphics[width=\textwidth]{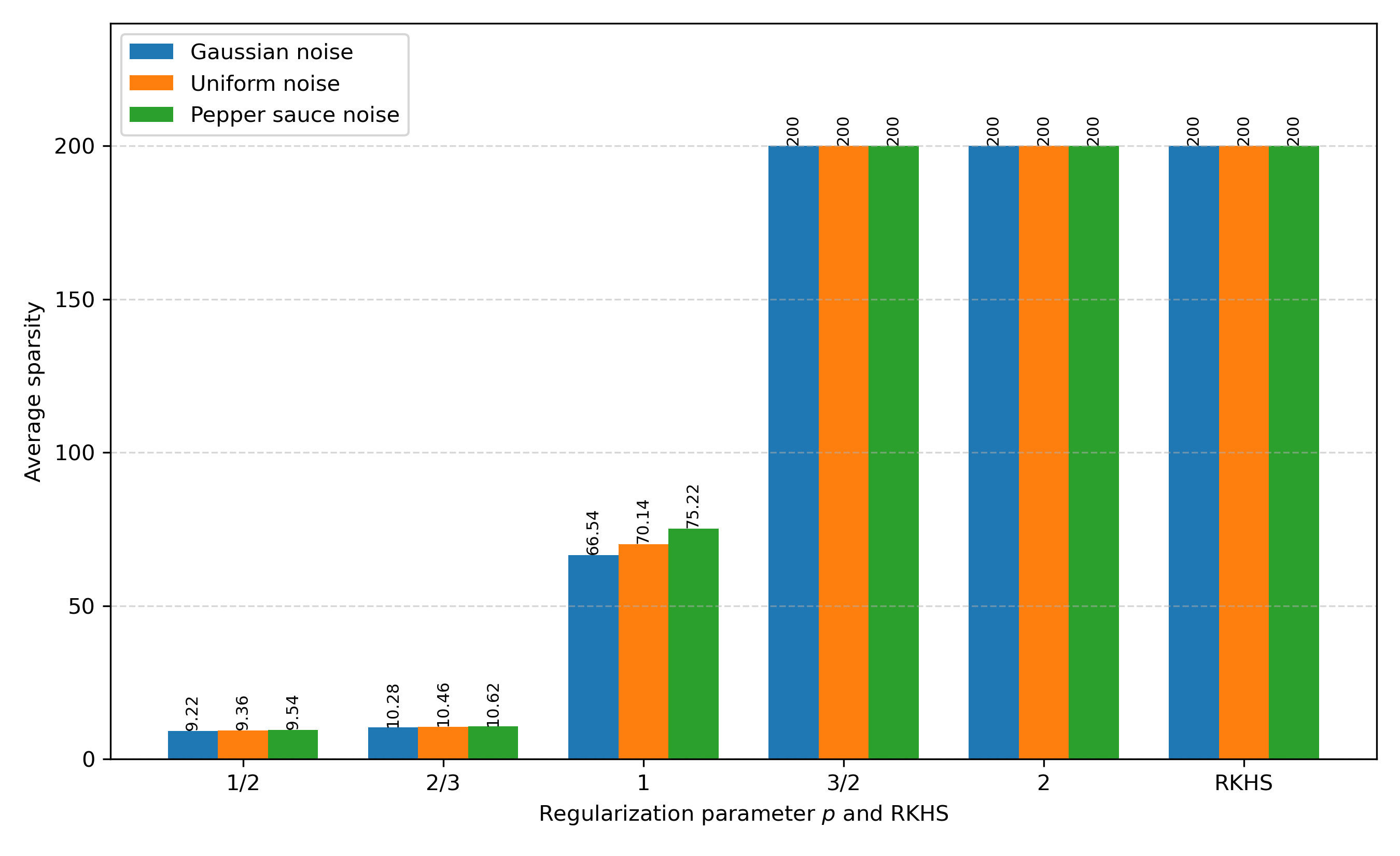}
        \caption{Average Sparsity of RKHS and \(\ell^p\)-Regularized Models under Different Noise Types Using the IMQ Kernel for the HRG function.}
        \label{fig:3-2}
    \end{subfigure}
\end{figure}

\begin{figure}[htbp]
    \ContinuedFloat
    \centering
    \begin{subfigure}[b]{0.48\textwidth}
        \centering
        \includegraphics[width=\textwidth]{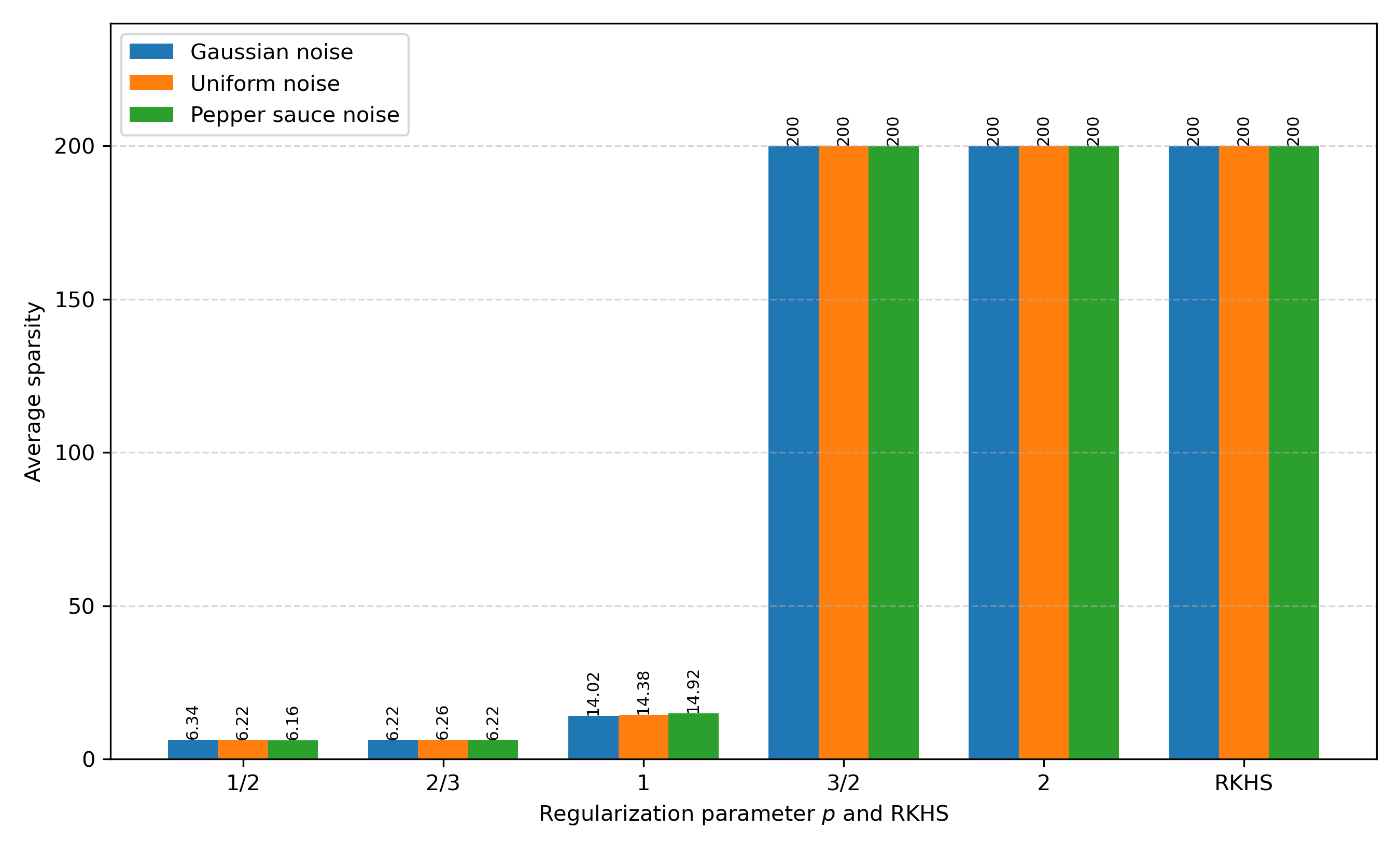}
        \caption{Average Sparsity of RKHS and \(\ell^p\)-Regularized Models under Different Noise Types Using the IMQ Kernel for the SPF function.}
        \label{fig:3-3}
    \end{subfigure}
    \hfill
    \begin{subfigure}[b]{0.48\textwidth}
        \centering
        \includegraphics[width=\textwidth]{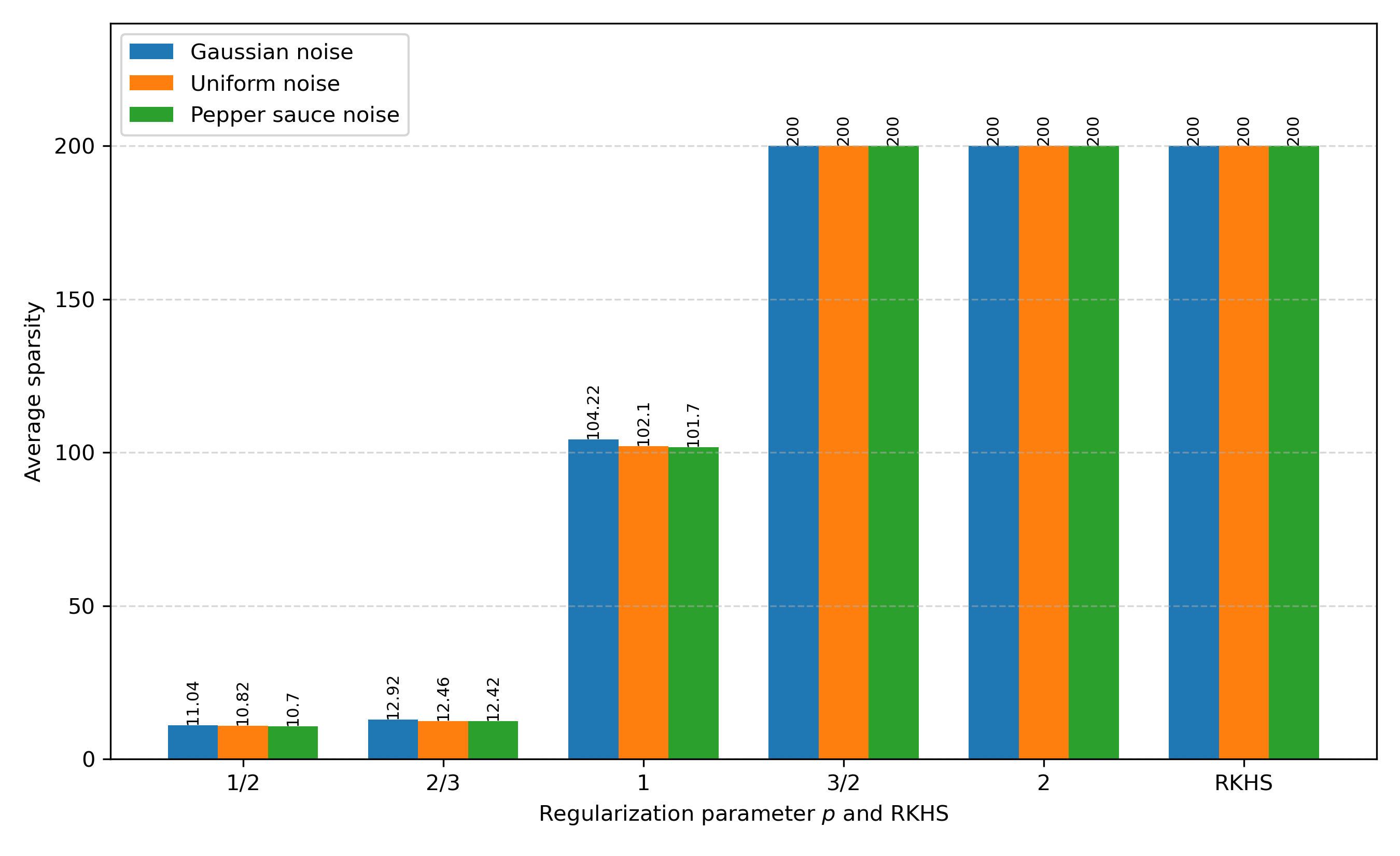}
        \caption{Average Sparsity of RKHS and \(\ell^p\)-Regularized Models under Different Noise Types Using the IMQ Kernel for the MTH function.}
        \label{fig:3-4}
    \end{subfigure}
\end{figure}

\section*{Declaration of AI use}
This work was carried out in the traditional manner, with minimal assistance from AI. The mathematical results and proofs were derived on the blackboard. The initial manuscript was developed by the authors without AI assistance. During subsequent revisions, DeepSeek and GPT-6-Astra were used for translation and language editing. The authors assume full responsibility for the accuracy, originality, and integrity of the final manuscript.

{\small
\bibliographystyle{amsplain-initials}
\bibliography{references}
}

\end{document}